\documentclass[sigconf]{acmart}
\renewcommand\footnotetextcopyrightpermission[1]{} % removes footnote with conference information in first column
\usepackage{fancyhdr}
\usepackage{algorithm}
\usepackage{algorithmic}
\usepackage{threeparttable}
\usepackage{makecell}
\usepackage{graphicx}
\usepackage{subcaption}

\AtBeginDocument{%
  \providecommand\BibTeX{{%
    \normalfont B\kern-0.5em{\scshape i\kern-0.25em b}\kern-0.8em\TeX}}}

\begin{document}

\title{Exploring Hidden Semantics in Neural Networks with Symbolic Regression}

\author{Yuanzhen Luo}
\affiliation{
    \institution{Beijing Key Laboratory of Petroleum Data Mining, China University of Petroleum-Beijing}
    \city{Beijing}
    \country{China}
}
\author{Qiang Lu}
\authornote{Corresponding author.}
\affiliation{
    \institution{Beijing Key Laboratory of Petroleum Data Mining, China University of Petroleum-Beijing}
    \city{Beijing}
    \country{China}
}
\email{luqiang@cup.edu.cn}
\author{Xilei Hu}
\affiliation{
    \institution{College of Artificial Intelligence , China University of Petroleum-Beijing}
    \city{Beijing}
    \country{China}
}
\author{Jake Luo}
\orcid{0000-0002-3900-643X}
\affiliation{
    \institution{Department of Health Sciences and Administration, University of Wisconsin Milwaukee, Milwaukee, WI, United States}
    \city{Beijing}
    \country{China}
}
\author{Zhiguang Wang}
\affiliation{
    \institution{Beijing Key Laboratory of Petroleum Data Mining, China University of Petroleum-Beijing}
    \city{Beijing}
    \country{China}
}

\begin{abstract}
Many recent studies focus on developing mechanisms to explain the black-box behaviors of neural networks (NNs). However, little work has been done to extract the potential hidden semantics (mathematical representation) of a neural network. A succinct and explicit mathematical representation of a NN model could improve the understanding and interpretation of its behaviors. To address this need, we propose a novel symbolic regression method for neural works (called SRNet) to discover the mathematical expressions of a NN. SRNet creates a Cartesian genetic programming (NNCGP) to represent the hidden semantics of a single layer in a NN. It then leverages a multi-chromosome NNCGP to represent hidden semantics of all layers of the NN. The method uses a (1+$\lambda$) evolutionary strategy (called MNNCGP-ES) to extract the final mathematical expressions of all layers in the NN. Experiments on 12 symbolic regression benchmarks and 5 classification benchmarks show that SRNet not only can reveal the complex relationships between each layer of a NN but also can extract the mathematical representation of the whole NN. Compared with LIME and MAPLE, SRNet has higher interpolation accuracy and trends to approximate the real model on the practical dataset.

\end{abstract}

\keywords{symbolic regression, neural network, Cartesian genetic programming}
\maketitle
\section{Introduction}
Neural networks (NNs) have been successfully applied in many problems, such as CNN for object recognition \cite{ren2015faster}, RNN for time series analysis \cite{goodfellow2016deep}, and Bert for natural language processing (NLP) \cite{devlin2018bert}. However, neural networks are often seen as black-boxes because their input-output (IO) relationships are difficult for a human to understand \cite{guidotti2018survey}. Sometimes, it is almost impossible to interpret the NN behaviours when models make unexpected predictions on some datasets, such as adversarial examples \cite{szegedy2013intriguing} and white noise images \cite{nguyen2015deep}. Some of the recent work has been centred on researching and explaining the black box behaviours, as summarized in several survey papers \cite{samek2019towards,bodria2021benchmarking}. 

%  Many researchers started to look into the black box. They proposed various methods to explain NN, mainly classified into example, attribution, hidden semantic, and rule according to the type of explanations \cite{samek2019towards,bodria2021benchmarking}. The example explanation method shows the NN interpretation by supporting or counter example \cite{zhang2020survey}. 

In this paper, we aim to develop a new symbolic regression-based method to explore hidden semantics in NNs. Here, a typical hidden semantics interpretation method refers to explaining a NN with an explicit math function. If a function $f(x^{(i)})$ explains a single hidden layer $h(x^{(i)})$ on a certain input-output $\left(x^{\left(i\right)},y^{\left(i\right)}\right)$ or a small range of inputs-outputs, it is called \textbf{local explanation} \cite{zhang2020survey}, such as LIME \cite{ribeiro2016should} and MAPLE \cite{plumb2018model}. If a function $f(x)$ is able to explain a hidden layer $h(x)$ on the whole dataset, it is called \textbf{global explanation}, such as Visualization method \cite{mahendran2015understanding, carter2019exploring} and Net2vec \cite{fong2018net2vec}. Although these methods show some degree of success in extracting hidden semantics from NN, they have the following three limitations. (1) Local explanation methods can give a mathematical expression, such as a linear model and a decision tree, for each $(x^{(i)},y^{(i)})$. However, they cannot obtain a general expression for the whole dataset. Although most global explanation methods can visualize NNs on the whole dataset, they cannot give a mathematical expression for explaining the dataset. (2) These local or global methods often leverage pre-defined interpretable models to explain the hidden semantics of NNs. For example, LIME can use linear models, decision trees, or falling rule lists as interpretable models. However, these pre-defined models may not capture hidden semantics in some situations because the real characteristics of a NN model are often unknown, and applying a predefined model to explain the networks may be inappropriate. (3) These methods cannot generate a mathematical expression that can represent the hidden semantics of all layers in a NN. 
%Each of them is a local explanation or global explanation, which cannot explain local and global hidden semantics simultaneously. So, they does not illustrate the relationship between hidden layers (local-local) or between a hidden layer and the output layer (local-global). 

% \textbf{and GPX \cite{ferreira2020applying}} \textbf{and possibly non-linear mathematical expression (GPX)} \textbf{and DTME only uses decision trees} \textbf{For example, DTME can generate a best selected decision tree to explain globally for a NN. However it cannot provide the hidden semantics for each hidden layer since the decision tree only presents the IO relationship of the whole NN model.}
% \textbf{and Decision Tree-based Model Extraction (DTME) \cite{evans2019s}}

To overcome these limitations, this paper leverages the \textbf{symbolic regression} (SR) method to explain a NN. In SR, for a given dataset $\lbrace x,y \rbrace$, the algorithm can find a symbolic function $f(x)=y'$ that minimizes the distance between $y$ and $y'$ in the mathematical expression space. SR has great flexibility in generating mathematical expressions; hence, it does not need a predefined model to capture the relationships in the dataset. However, classical SR methods, such as GP \cite{koza1992genetic,schmidt_distilling_2009}, GEP \cite{ferreira2001gene}, and linear GP \cite{brameier2007linear}, usually handle the symbolic function $f(x)$ with a single output $y'$, i.e., $y'$ is a number and not a vector. They cannot represent the relationship $g(W_ih_{i-1}+b)$ of each layer $i$ in a NN because each layer's output is a vector, matrix, or tensor. Therefore, when these GPs explain NN \cite{evans2019s,ferreira2020applying}, they can only give a mathematical expression to show the semantics of the whole NN with a single output value, not each layer in the NN.  
Although Cartesian Genetic Programming (CGP) \cite{miller2000cgp} supports multiple outputs, CGP cannot represent semantics in a NN. Because each CGP output corresponds to a hidden node, and it cannot provide a general model $f$ that represents hidden semantics in the layer. To obtain a general model, we assume that the relationship between input and output in a layer (or a NN) has the mathematical expression format $w_i^sf_i(h_{i-1}^s)+b_i$, where $w_i$ and $b_i$ may be a number, vector, matrix, or tensor, and $f_i$ is a mathematical function that represents the hidden semantics of the layer.        

\begin{figure}[htbp]
    \centering
    \includegraphics[width=0.47\textwidth]{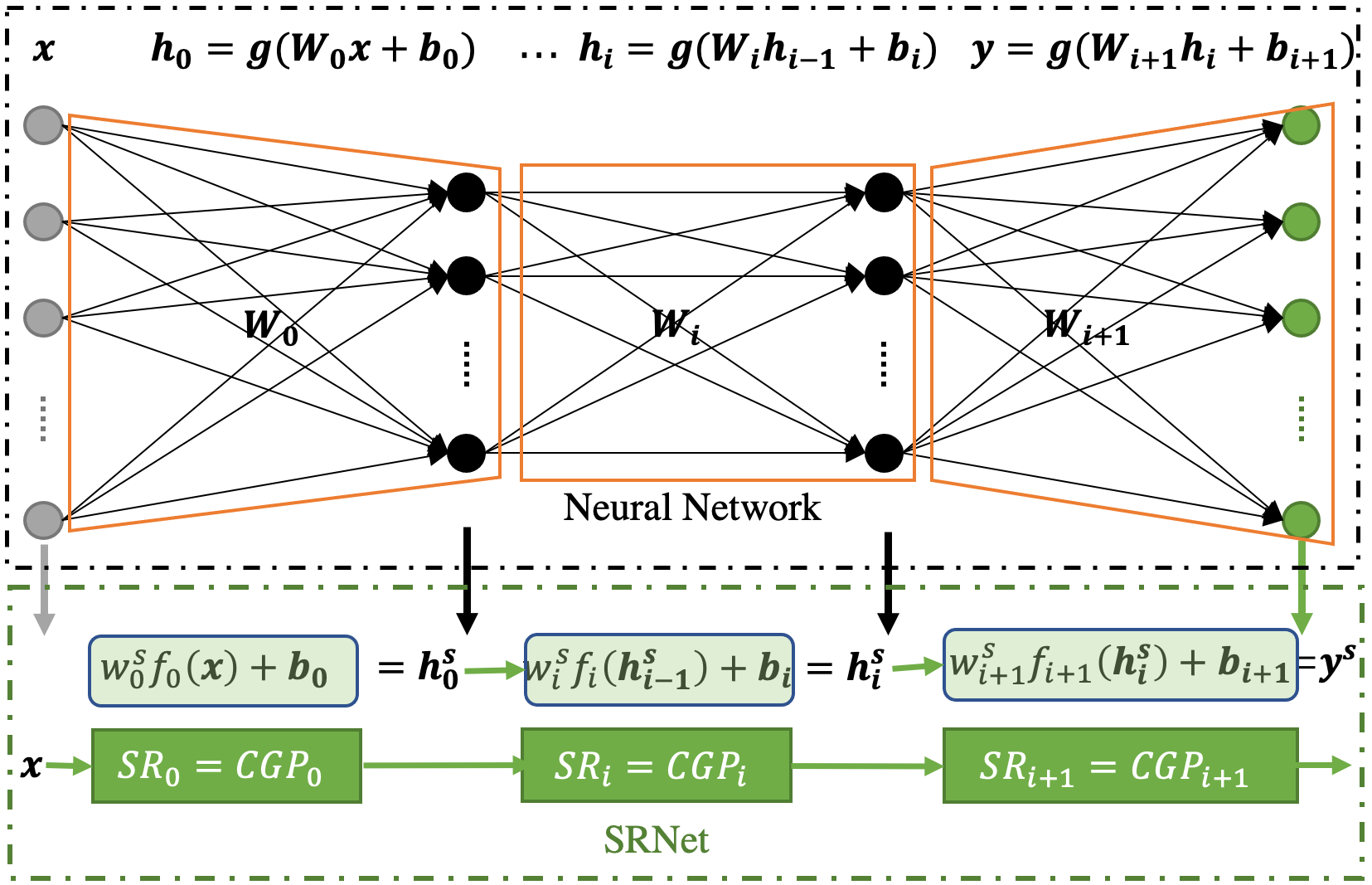}
    \caption{SRNet for exploring hidden semantics in NN.}
    \label{fig:SRNet}
\end{figure}

Based on the above assumption, this paper proposes a novel SR method (called \textbf{SRNet}) to mine hidden semantics of all layers in a NN simultaneously, as shown in Figure \ref{fig:SRNet}. SRNet is an evolutionary computing algorithm. In each evolution,  
SRNet first leverages the Cartesian Genetic Programming (CGP) \cite{miller2008cartesian, miller_cartesian_2019} to find each layer's mathematical function $f_i(h_{i-1}^s)$. 
It then uses the Newton-Raphson method \cite{ryaben2006theoretical} (or L-BFGS method \cite{liu1989limited}) for few (or many) variables to obtain $w_i^s$ and $b_i$ so that $h_i^s=w_i^sf_i(h_{i-1}^s)+ b_i$ approximates the output $h_i$ of the layer $i$ in a NN. At the end of the evolution, SRNet will capture hidden semantics of all layers in a NN when $h_i^s \approx h_i$ (including $y^s \approx y$).
The main contributions in the paper are summarized as follows:

\begin{itemize}
%	\item To the best of our knowledge, the paper is the first work to explain NN with SR. Compared with known NN explanation methods \cite{zhang2020survey}, SR is suitable for exploring all hidden semantics in NN because it does not need to pre-define a model to represent NN.
	\item The paper proposes a new method called SRNet to explain hidden semantics of all layers in a NN. SRNet generates a mathematical expression in the format of $w_i^s f_i(h_{i-1}^s)+b_i$ that can be used to explain a NN. 
	\item To speed up SRNet, we create a multi-chromosome CGP \cite{walker2006multi} evolutionary strategy embedded in the Newton-Raphson method.
	\item Experiments show that the proposed SRNet can capture hidden semantics in NN in 12 SR benchmarks and 5 classification benchmarks. Compared with LIME and MAPLE, SRNet has higher interpolation accuracy and trends to approximate the real model on the practical dataset.    
\end{itemize}

The remainder of this paper is organized as follows. In Section 2, we introduce the background knowledge about Cartesian Genetic Programming. Then, we propose SRNet to explore hidden semantics in NNs in Section 3. Section 4 and 5 report the experimental results. We conclude the paper in Section 6.

\section{Cartesian Genetic Programming}
\begin{figure}[htbp]
    \centering
    \includegraphics[width=0.46\textwidth]{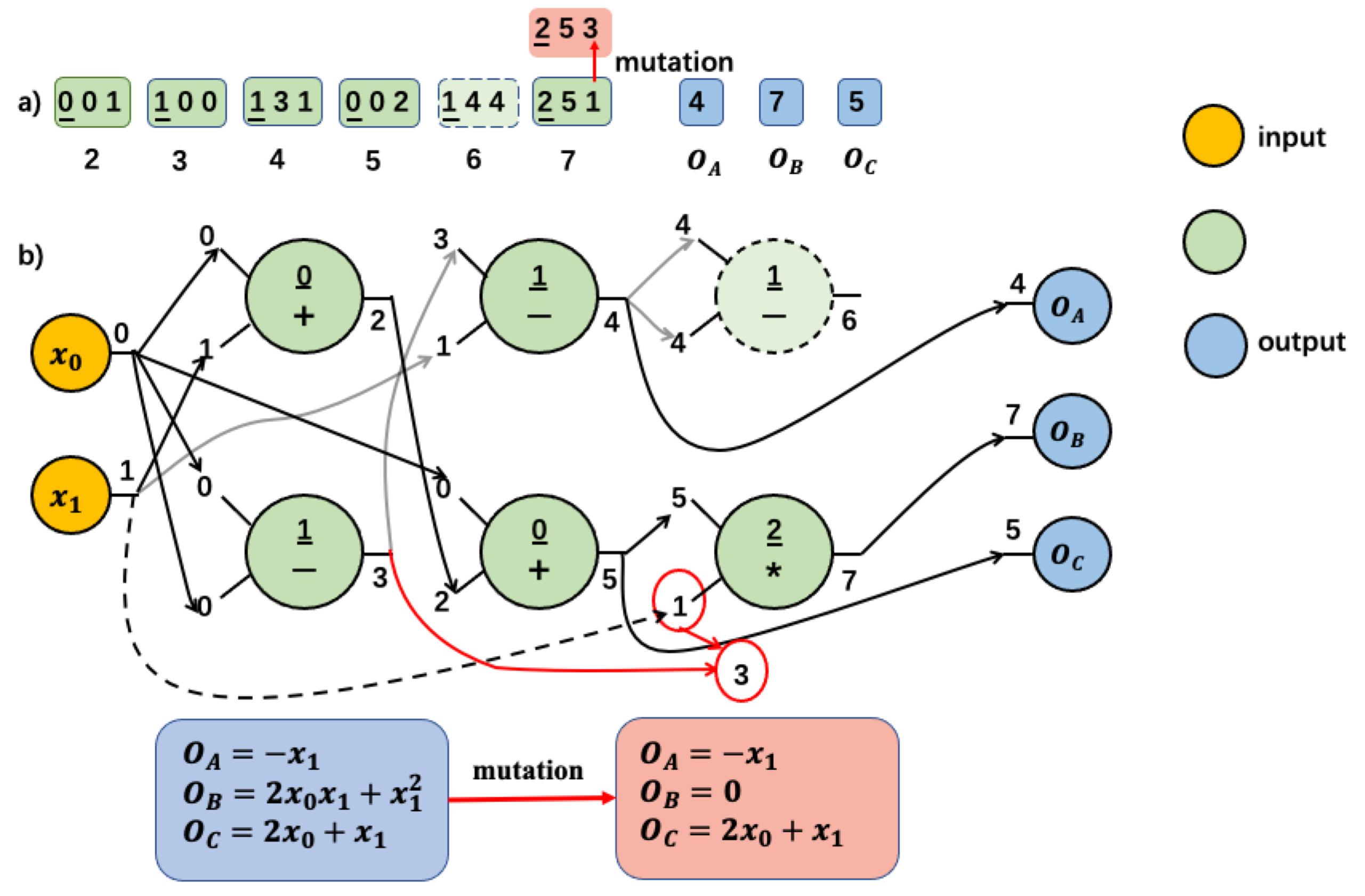}
    \caption{An example of CGP. a) genotype. b) phenotype.}
    \label{fig:CGPex}
\end{figure}

CGP is a directed acyclic graph-based genetic programming algorithm for addressing the SR problem \cite{miller2000cgp}. In CGP, the graph consists of a two-dimensional grid of computational nodes, as shown in Figure \ref{fig:CGPex}. These nodes are classified into three categories: input, output, and function. The input (or output) nodes represent the input (or output) values $x$ (or $o$), which is encoded into an integer, such as $x_0$ encoded by "$1$" and $O_A$ encoded by "$4$". The function nodes are computational expressions. Each function node has three parts, input, computational expression, and output, encoded by a series of integers. As each node has only one output, the output code is used to index the node. For example, a function node "$+$" is regarded as the code "$\langle \underline{0}012\rangle$". In this code, the first integer $\underline{0}$ is the code of the function "$+$". The two middle integers "$0$" and "$1$" represent two inputs of the function "$+$", and they are also the outputs of two previous nodes. The last integer "$2$" is the index of the node. 

In the example CGP diagram, there is no edge between any two nodes in the same column. Two nodes at different columns can be linked if one's input code equals the other's output code. A genotype is used to represent a CGP, as shown in Figure \ref{fig:CGPex}. The genotype contains two categories of nodes: the functional nodes and the output nodes. As the genotype has multiple output nodes, the genotype can describe multiple computational expressions. For example, the genotype in Figure \ref{fig:CGPex} generates three mathematical expressions, $O_A=-x_1$, $O_B=2x_0x_1+x_1^2$, and $2x_0+x_1$. 

CGP usually leverages the $(1+\lambda)$ evolutionary strategy \cite{rechenberg1978evolutionsstrategien,jansen2013analyzing} to find the best fitted mathematical expression. In each evolution, $(1+\lambda)$ EA utilizes mutation to generate $\lambda$ offsprings. For CGP, the mutation randomly chooses a gene location and changes the allele at the location to another valid random value. A valid value is from the function look-up table if a computational expression gene is chosen for mutation. If an input gene is chosen, a valid value is from the output set of its previous nodes. The mutation does not change the output gene. For example, in Figure \ref{fig:CGPex}, a mutation changes the input gene "$1$" of the node "$7$" to "$3$". Then, the output $O_B$ becomes "$0$".

The multi-chromosome Cartesian genetic programming (MCGP) \cite{walker2006multi} encodes multiple chromosomes into a single genotype. Each chromosome code is similar to the genotype of CGP. So, MCGP can provide a solution to a large problem by dividing it into many smaller sub-problems. For simultaneously exploring hidden semantics of all layers in a NN, MCGP encodes each layer semantics as a chromosome into a genotype. Thus, a chromosome represents the semantics of one layer, and the genotype represents all NN layers' semantics. After MCGP uses the ($1+\lambda$) multi-chromosome evolutionary strategy \cite{Walker2011} to acquire a best-fitted individual, it also obtains these semantics.

\section{SRNet}

This section proposes the SRNet, a method based on MCGP \cite{walker2006multi} to simultaneously explain the hidden semantics of layers in a NN. As shown in Figure \ref{fig:SRNet}, SRNet can find a group of $h_i^s=w_i^s f_i(h_{i-1}^s)+ b_i$ that approximates each NN layer output $h_i$, i.e., 
\begin{equation}
	\lbrace h_0^s,...,h_n^s \rbrace = \underset{h_i^s \in \mathcal{F}}{\mathit{arg} min} \sum_{i=0}^{n} \mathcal{L}(h_i,h_i^s).
\end{equation}
To find the group $\lbrace h_i^s \rbrace$ quickly, SRNet needs to address the following problems: 1) how to encode these $h_i^s$s, and 2) how to find functions to explain hidden semantics in a NN. Section \ref{sec:architecture} shows our solution to the first problem, and section \ref{sec:es} provides a multi-NNCGP evolutionary strategy to address the second problem. 
  
\subsection{SRNet Encoding} \label{sec:architecture}
\begin{figure}[htbp]
	\centering
	\includegraphics[width=0.47\textwidth]{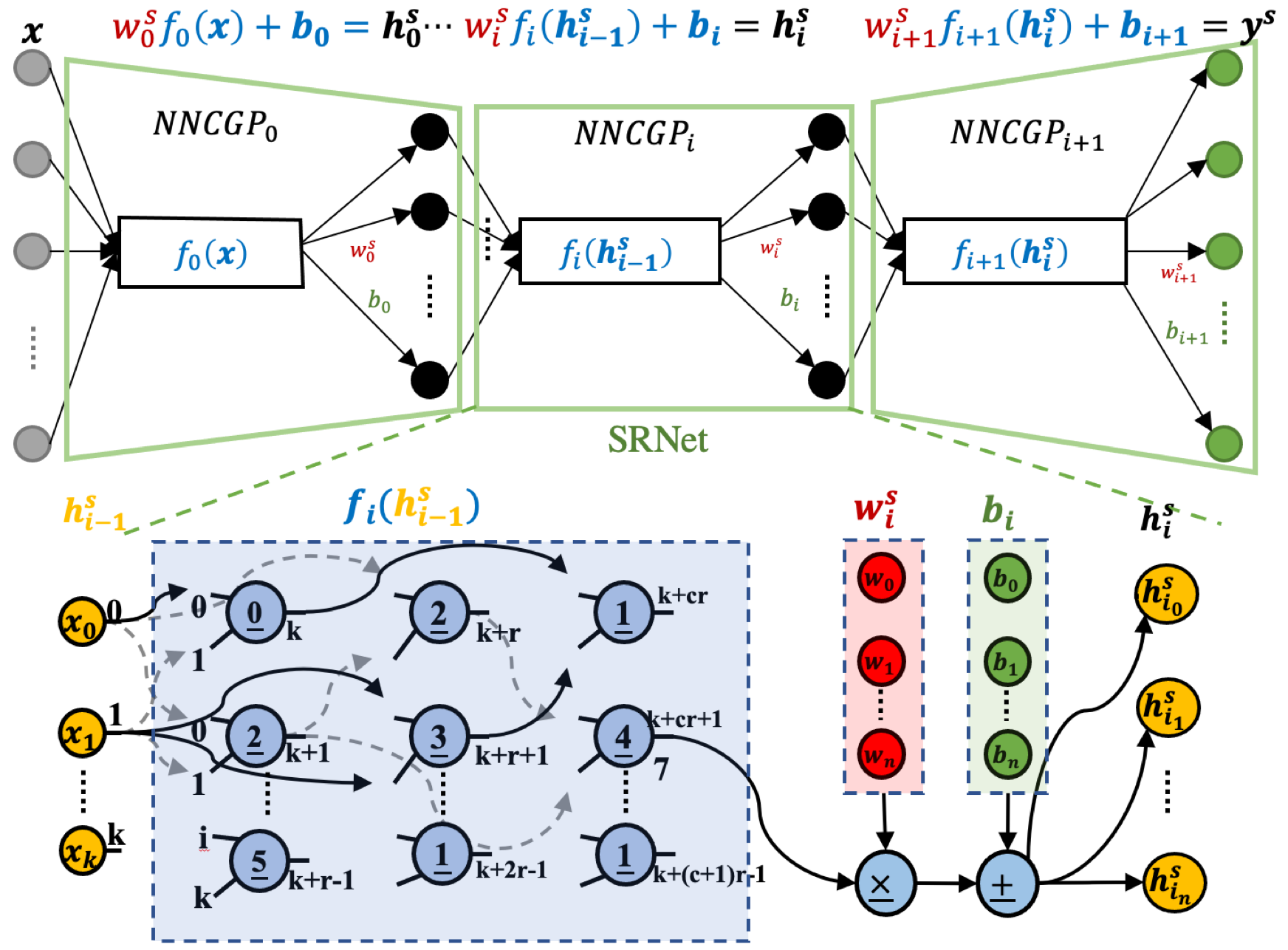}
	\caption{SRNet encoded by Multiple NNCGPs}
	\label{fig:cgpsrnet}
\end{figure}

The hidden semantics $h_i^s$  of each layer in a NN are represented as
\begin{equation} \label{eq:layermodel}
   w_i^s f_i(h_{i-1}^s) + b_i, 	
\end{equation}
where $f_i(h_{i-1}^s)$ is a mathematical expression that represents the general semantics in the layer $i$, and $w_i^s$ and $b_i$ are a weight vector, matrix, or a tensor, as shown in Figure \ref{fig:cgpsrnet}. 

To capture a NN layer's semantics $h_i^s$, we define that the neural network CGP (NNCGP) consists of three components, including general semantic model, constant, and operator. The general semantic model is used to generate $f_i(h_{i-1}^s)$. It has three parts, $k$ inputs, $c\times r$ functions and an output.  2) The constant component includes the weight vector $w_i^s$ and the bias vector $b_i$, where $|w_i^s|=|b_i|=|h_i^s|$. 3) The operator component consists of the two functions, "$\times$" and "$+$". 

SRNet uses multiple NNCGPs (called MNNCGP) to encode genotype. Since a NNCGP can represent one layer's semantic, multiple NNCGPs can be used to capture hidden semantics of all layers in a NN. These NNCGPs are regarded as chromosomes that constitute a genotype. In the genotype, each NNCGP$_i$'s output $h_i^s$ is the input of its next NNCGP$_{i+1}$.

\subsection{Evolution Strategy} \label{sec:es}

SRNet leverages a multi-NNCGP evolutionary strategy embedded by the Newton-Raphson method (called \textbf{MNNCGP-ES}) to find the best-fitted genotype that represents hidden semantics of all layers in a NN. MNNCGP-ES is similar to the ($1+\lambda$) multi-chromosome evolutionary strategy \cite{Walker2011}. MNNCGP-ES includes the following operations: mutation, fitness evaluation, and selection. Mutation, with a certain probability, change each allele in MMCGP to another valid random value \cite{miller2000cgp}.  

\subsubsection{Fitness Evaluation}
  To evaluate a genotype encoded by MNNCGP, the fitness function is defined as the following equation,
  \begin{equation} \label{eq:fitness}
  	\mathit{fitness} = \frac{1}{N} \sum_{i=0}^{N-1} \mathcal{L}(h_i,h_i^s) + \mathcal{L}_o(y,y^s)
  \end{equation}
  where $\mathcal{L}$ is the mean squared error (MSE) of each middle layer. $\mathcal{L}_o$ is an error function of the output layer. It is a cross-entropy loss in the classification task, while it is MSE in the regression task. $h_i$ ($h_i^s$) is the output of the $i$th layer in a NN (NNCGP$_i$). $y$ and $y^s$ are the outputs of the NN and SRNet, respectively.  

Equation \ref{eq:layermodel} indicates that obtaining $h_i^s$ needs two computations, as shown in Figure \ref{fig:cgpsrnet}. One is $f(h_{i-1}^s)$ that generates an output by the CGP code. The other is the parameter computation that obtains the constant vectors, $w_i^s$ and $b_i$, by the Newton-Raphson method that performs an update operation according to the following equation. 
\begin{equation} \label{eq:newton}
	p = p - H^{-1}(p)\nabla l(p),
\end{equation}  
where $p$ is $w_i^s$ or $b_i$,  $H(p)$ is Hessian, $\nabla l(p)$ is the gradient of the loss function $h_i - (w_i^sf(h_{i-1}^s)+b_i)$. The Hessian is difficult to obtain if it is a high-dimensional matrix (i.e., many neurons). Therefore, to solve this problem, the Limited-memory Broyden-Fletcher-Goldfarb-Shanno algorithm (L-BFGS) \cite{liu1989limited} is used for limited memory and time-saving.

\subsubsection{Selection} 

\begin{figure}[htpb]
  \centering
  \includegraphics[width=0.47\textwidth]{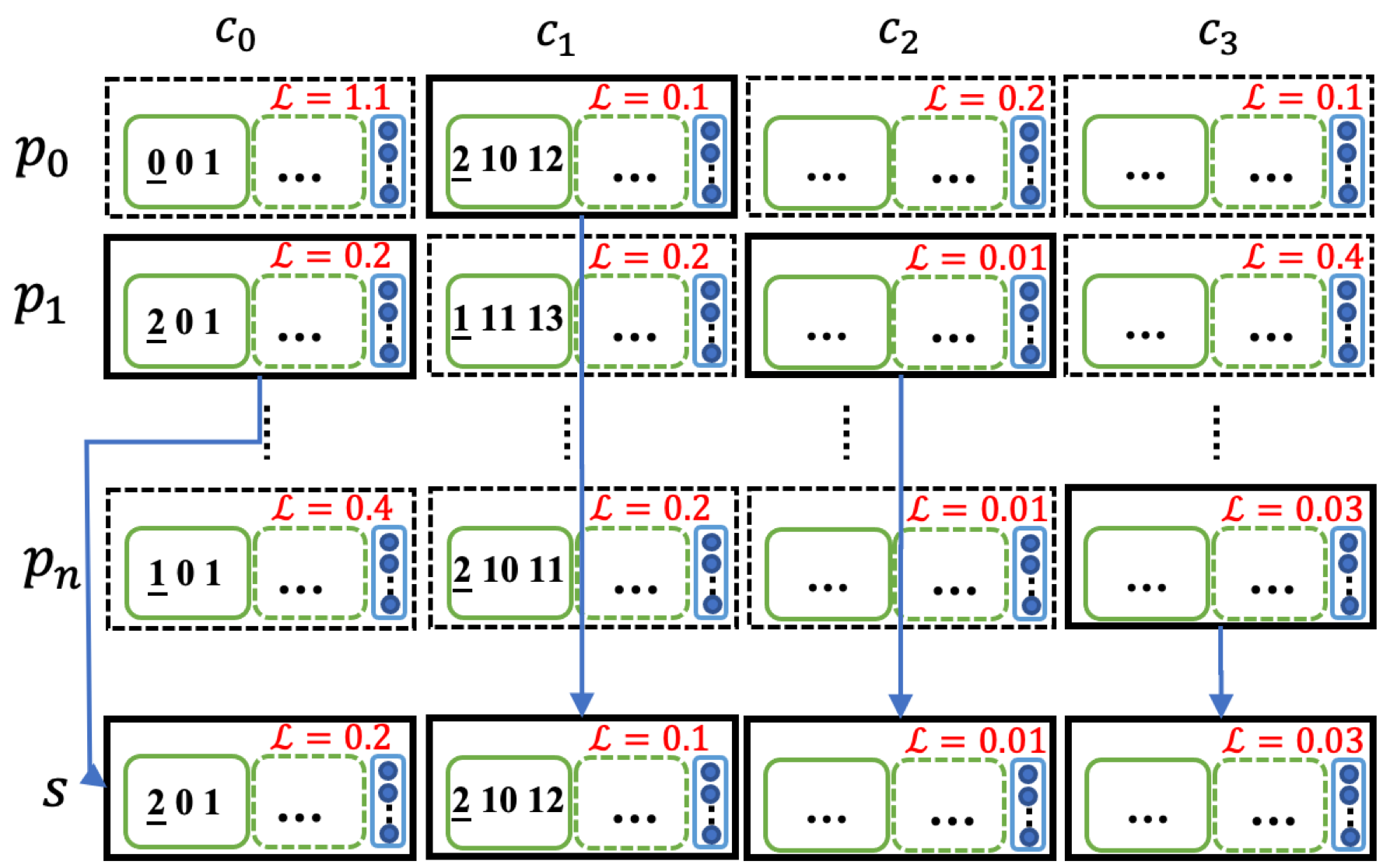}
  \caption{Selecting chromosomes at each points from all individuals as a "super" individual.}	
  \label{fig:selection}
\end{figure}

Selection aims at generating a 'super' individual from the population. The 'super' individual consists of a set of super chromosomes that have the best fitness at each position from all individuals, i.e., $min\ \mathcal{L}(h_i,h_i^s)$, as shown in Figure \ref{fig:selection}. Since these chromosomes constitute an input-output sequence where each chromosome output is the input of its next chromosome, they need to be selected in the order of their positions in the population. After evaluating the finesses of the chromosomes at a certain position ($c_0$) of the population, the selection picks up the chromosome that has the best fitness score as the super chromosome ($s_0$). Then, it evaluates chromosome at the next position($c_1$) with $s_0$ as input. Moreover, it obtains a chromosome with the best fitness as the next super chromosome ($s_{1}$). Repeating the above evaluation policies results in a group of selected chromosomes $\lbrace s_0,s_1,...\rbrace$. These selected chromosomes form a super individual.
% Two selection strategies are introduced to deal with low-dimensional and high-dimensional problems(i.e., each $h_i$ only a few or many outputs), respectively. 
% For low-dimensional problems, 

However, for high-dimensional problems, evaluating the finesses of the chromosomes is very time-consuming when using the Newton-Raphson method. So, L-BFGS is used to replace the Newton-Raphson method to compute the two weight vectors ($w_i^s$ and $b_i$) of all individuals. L-BFGS can speed up the fitness evaluation.  

% Before selecting the 'super' individual, L-BFGS compute 
% before we select the 'super' individual, we train all individuals by L-BFGS every 50 generations. Then the 'super' individual is selected using the minimal fitness score calculated by Equation \ref{eq:fitness}. This approach could significantly improve the running time.

\subsubsection{MNNCGP-ES}

The MNNCGP-ES pseudocode is listed in Algorithm \ref{alg:mnncgpes}. MNNCGP-ES combines the fitness evaluation and the selection method mentioned before, using the $(1+\lambda)$ evolution strategy to evolve generation-by-generation to obtain the optimal individual. 

\begin{algorithm}[ht]
	\caption{MNNCGP-ES}
	\label{alg:mnncgpes}
	\begin{algorithmic}[1]
	    \REQUIRE $\mathbb{D}s ( \boldsymbol{h_0}, \boldsymbol{h_1}, \dots, \boldsymbol{h_{n}})$, $\lambda$
	    \ENSURE a best-fitted individual
	    \STATE randomly initializes $\lambda$ individuals with MNNCGP\\
	    \WHILE {$fitness > 1e-4$ and max generation not reached}  
	       \STATE //obtain an new parent $s$ by the selection operation \\
    	   \STATE  $s \gets \mathit{NULL}$  \\
    	       % \FORALL { each $i$ in the range $[0,n)$}
    	   \STATE calculate each chromosome's output $h_{i}^s$ at the position $i$ according to Equations \ref{eq:layermodel} and \ref{eq:newton}; \\
    	   \STATE obtain the chromosome $c_{i}^k$ by Equation \ref{eq:fitness}\\
    	            
    	       % \ENDFOR
    	   % \ELSE 
	       %     \STATE obtains  $\lambda$ individuals by L-BFGS every 50 generations
	       %     \STATE select s with minimal fitness calculated by Equation \ref{eq:fitness}
	       % \ENDIF
	        \STATE $s \gets s\cup c_{i}^k$ \\
	        \STATE execute $\lambda$ mutations on $s$ to generate $\lambda$ offsprings.\\
	    \ENDWHILE
	    \STATE Return the best-fitted parent $s$ according to the fitness computed by Equation \ref{eq:fitness}.
	\end{algorithmic}
\end{algorithm}

\section{Experiments}

\begin{figure}[ht]
	\centering
	\scalebox{0.8}{
	\includegraphics[width=0.5\textwidth]{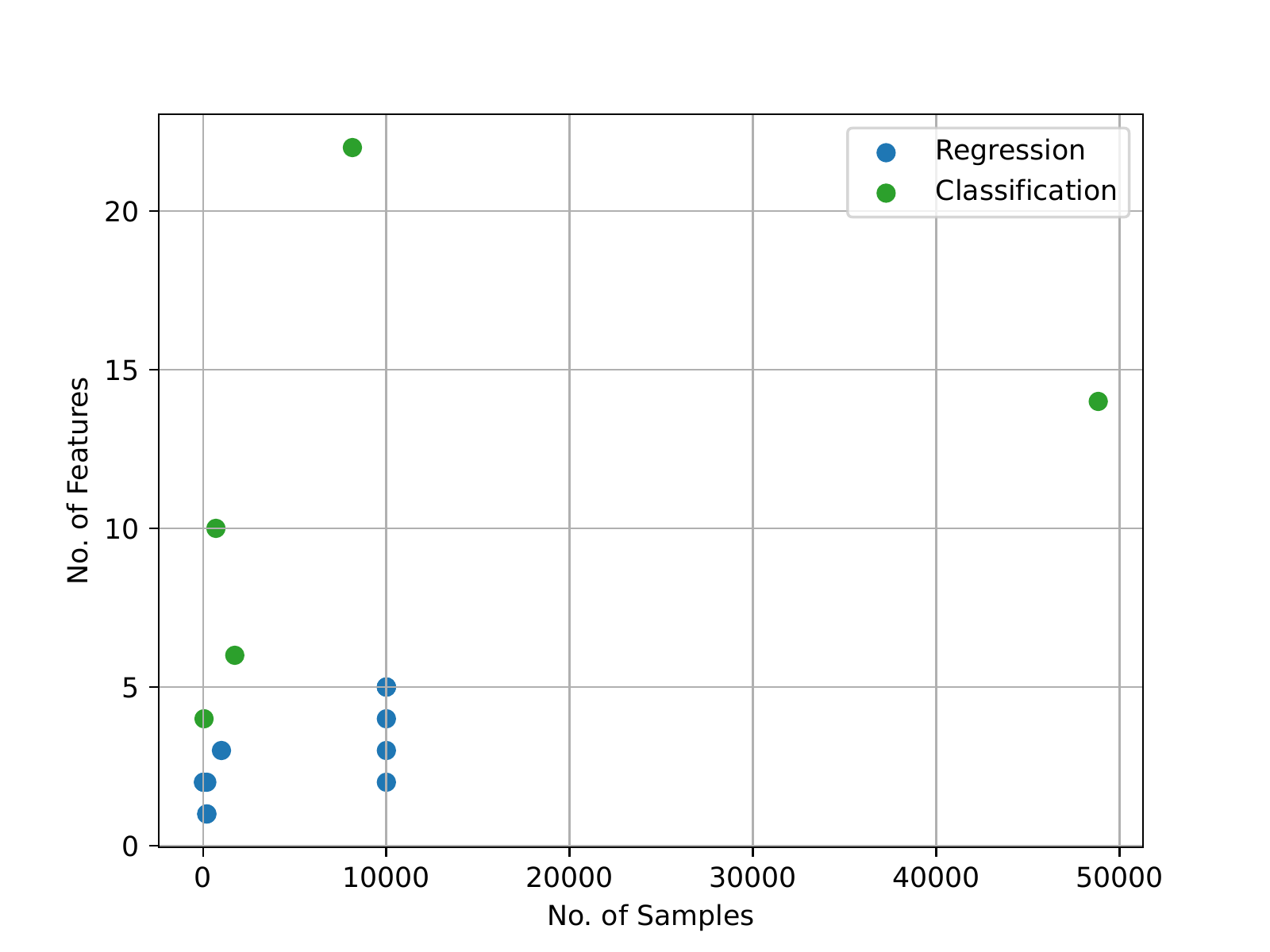}
	}
	\caption{The sizes of samples and features in 17 benchmarks }
	\label{fig:dataset}
\end{figure}

To validate the SRNet's ability to explain hidden semantics in the neural network (NN), we tested the SRNet
% \footnote{source code at https://github.com/KGAE-CUP/SRNet} 
on the built NNs of 12 symbolic regression benchmarks as well as 5 classification benchmarks, listed in Table \ref{tab:dataset}. Moreover, the sample sizes and feature sizes of the 17 benchmarks are illustrated in Figure \ref{fig:dataset}.

\begin{table}[h]
    \scalebox{0.9}{
	\begin{threeparttable}
      \begin{tabular}{cp{2.7cm}<{\centering}cp{2cm}<{\centering}}
        \hline
        Alias & Function/Name & Training Dataset & MLP\\
        \hline
        $K0$ & $sin(x)+sin(x+x^2)$ & $ [-1,1,200]$ & $[3,3]$ s0.01 \\
             
        $K1$ & $2sin(x)cos(y)$ & $ [-1,1,200]$ & $[3,3]$ s0.1 \\
        
        $K2$ & $3+2.13ln\left|x\right|$ & $[-50,50,200]$ & $[5,5]$ s0.03 \\
              
        $K3$ & $\frac{1}{1+x^{-4}}+\frac{1}{1+y^{4}}$ & $ [-5,5,10^4]$ & $[4, 4, 4]$ a0.03 \\

        $K4$ & $\frac{30xy}{(x-10)z^2}$ &\makecell[c]{$x,y:[-1,1,10^3]$ \\ $z:[1,2,10^3]$}                 & $[4, 4]$ a0.003 \\
             
               \\
        $K5$ & \makecell[c]{$xy + sin((x-1)$\\$(y-1))$} & $[-3,3,20]$ & $[5, 5]$ a0.003 \\

        $F0$ & $\frac{m_0}{\sqrt{1-\frac{v^2}{c^2}}}$ & \makecell[c]{$m_0: [1,5,10^4]$\\$ v: [1,2,10^4]$\\$c: [3,10,10^4]$ }  & $[3, 3]$ a0.01 \\

        $F1$ & $q_1q_2\frac{r}{4\pi\epsilon r^3}$ & $[1,5,10^4]$ & $[3, 3]$ a0.01 \\

        $F2$ & $Gm_1m_2(\frac{1}{r_2}-\frac{1}{r_1})$ & $[1,5,10^4]$ &$[3, 3]$ a0.01\\
        
        $F3$ & $ \frac{1}{2}kx^2 $ & $[1,5,10^4]$ & $[3, 3]$ a0.01 \\
        
        $F4$ & \makecell[c]{$-6.4\frac{G^{4}}{c^{5}} \frac{1}{r^{5}} \left(m_{1} m_{2}\right)^{2}$\\$\left(m_{1}+m_{2}\right)$} & \makecell[c]{$m_1, m_2: [1,5,10^4]$\\$G,c,r: [1,2,10^4]$} & $[5, 5]$ a0.03 \\
        
        $F5$ & \makecell[c]{$ \frac{q}{4 \pi \epsilon y^{2}}[4 \pi \epsilon V_{e} d- $ \\ $\frac{q d y^{3}}{\left(y^{2}-d^{2}\right)^{2}}] $} & \makecell[c]{$q, V_{e}, \epsilon: [1,5,10^4]$ \\ $ d:[4,6,10^4]$ \\ $y:[1,3,10^4]$} & $[3, 3]$ a0.03 \\
        
        $P0$ & adult & 48842 & $[100, 100]$ s0.01 \\
        $P1$ & analcatdata\_aids & 50 & $[200, 100, 100]$ s0.01 \\
        $P2$ & agaricus\_lepiota & 8145 & $[100, 100]$ s0.01 \\
        $P3$ & breast & 699 & $[100, 100]$ s0.03 \\
        $P4$ & car & 1728 & $[100, 100, 100]$ s0.01 \\
        
        \hline
     \end{tabular}
    \end{threeparttable}
    }
    \caption{ The dataset of training 17 MLPs. 
%    In the column 'Training Dataset', '[a,b,c]' refers to
%    , according to the function in the Function column, 
%    randomly sampling c values from [a, b] for a variable. 
    In each cell of the column 'MLP', the integer list is the number of neurons in each hidden layer, 'a' or 's' is the Adam or SGD optimization method, respectively, float number being the learning rate. 
    }
    \label{tab:dataset}
\end{table}

\begin{table}[htpb]
    \centering
    \scalebox{0.8}{
    \begin{tabular}{ccc}
        \hline
            \textbf{Name} & \textbf{Parameter} & \textbf{Value} \\
        \hline
        LIME &\makecell[c]{Number of Features\\Number of Samples\\Distance Metric\\Regressor Model}& \makecell[c]{10\\5000\\Euclidean\\Ridge}\\
        \hline
        
        MAPLE &\makecell[c]{Number of Estimators\\Max Features\\Min Samples Leafs\\Regularization\\Ensemble Model\\Regressor Model\\Classifier Model} &\makecell[c]{200\\0.5\\10\\0.001\\Random Forest\\Ridge\\Logistic Regressor}\\
        \hline
        
        SRNet &\makecell[c]{Number of Rows\\Number of Cols\\Function Set\\Number of Constants\\Population Size\\Max Generations\\Mutation Probability} & \makecell[c]{10\\10\\$+, -, \times, \div, sqrt, square, \sin, \cos, ln, \tan, exp$\\1\\200\\5000\\0.4}\\
        \hline
    \end{tabular}
    }
    \caption{Algorithm parameters}
    \label{tab:algo_params}
\end{table}

Table \ref{tab:algo_params} lists the parameters of the three algorithms, LIME, MAPLE, and SRNet. All parameters of the three algorithms are fixed on all benchmarks. 
% All parameters of MNNCGP are fixed for the SRNet datasets. The number of function node was set to $10\times 10$, and the function set used for building explainable models is  $[+, -, \times, \div, \sqrt{(\cdot)}, (\cdot)^2,$ $ln(\cdot), sin(\cdot), cos(\cdot), tan(\cdot), exp(\cdot)]$. The point mutation probability was 0.4. The population size and evolution generation were set to 200 and 5000 respectively. 

\subsection{Regression Task}
To validate the SRNet's ability to explore hidden semantics of NN in the regression task, we chose 12 symbolic regression benchmarks $K0-K5$ and $F0-F5$. The benchmarks $K0-K5$ were chosen from the commonly used SR Benchmarks \cite{mcdermott2012genetic}, while $F0-F5$ are from physical laws \cite{udrescu2020ai}. We generated 12 datasets (called \textbf{true datasets}) according to these benchmarks. Each of these datasets has different sample sizes (see 'Training Dataset' in Table \ref{tab:dataset}). For example, for the $K1$ problem in Table \ref{tab:dataset}, we randomly sampled 200 $x$ and $y$ values in the range of $[-1,1]$, respectively. Moreover, combining these values can generate $200$ samples for the $K1$ dataset. For each of the 12 datasets, we randomly 
took $80\%$ samples from it as the \textbf{training dataset}, and the other as the \textbf{test dataset}. Then, we built 12 Multi-Layer Perception neural networks (\textbf{MLP}) with a sigmoid activation function using the training datasets. The training parameters are listed in Table \ref{tab:dataset}. For example, for $K1$, we created an MLP with two hidden layers where each layer had three hidden nodes. The MLP was trained by the SGD optimization method with a learning rate of $0.01$.

After each MLP was trained, 
% we standardized the input data to the Gaussian distribution $N(0, 1)$ distribution. 
we collected each NN layer's input and output data of the 12 MLPs as the NN explanation datasets. We then ran the MNNCGP-ES, LIME, and MAPLE 30 times on each NN explanation dataset. 

% MNNCGP-ES randomly initializes $w_i$ and $b_i$ on each layer according to the Gaussian distribution $N(0,1)$. 

\subsection{Classification Task} \label{sec:USDB}
To validate the SRNet's ability to explore hidden semantics of NN in the classification task, we chose 5 classification benchmarks named $P0-P4$ from the PMLB \cite{orzechowski2018we} with the different number of samples and features (see Figure \ref{fig:dataset}). The column 'Training Dataset' on the rows "P0-P4" indicates the number of samples, as shown in Table \ref{tab:dataset}. Training 5 MLPs is similar to the regression task except for the function "softmax" that replaces their output functions.

\begin{figure}[htpb]
  \centering
  \includegraphics[height=0.6\linewidth, width=\linewidth]{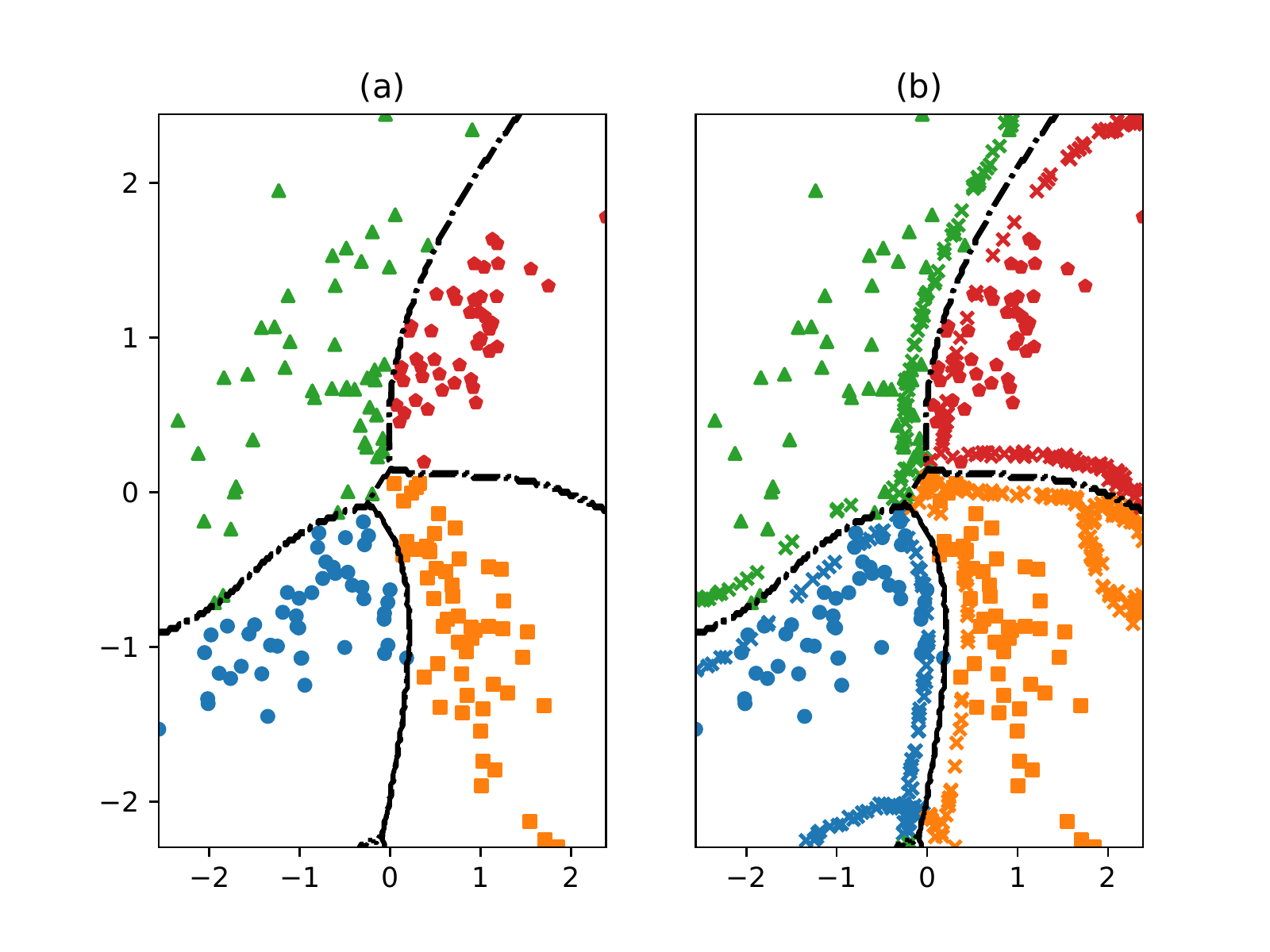}
  \caption{(a) The decision boundary of a classification model with 4 classes. (b) sampling around decision boundaries. The marker 'X' is represented as the new sample around the decision boundaries.}	
  \label{fig:SamplingDB}
\end{figure}

After training these classification MLPs, we need to compare SRNet with their decision boundaries, not their outputs. Because the MLP's outputs on the training datasets are sparse and do not fully represent the NN's classification ability, as shown in Figure \ref{fig:SamplingDB}(a). Using these outputs to train SRNet may result in wrong results. To obtain the decision boundaries of the trained MLP, we leverage a uniform sample method around its decision boundary (called \textbf{USDB}). USDB first evaluates the range of the training dataset. It then randomly samples $n$ points in the range. It finally selects $s$ points with the shortest distance to the decision boundary according to Equation \ref{eq:distance} \cite{fawzi2018empirical, li2018decision}.

\begin{equation} \label{eq:distance}
    d(x_i, B) = \sum^C_k{|p_k(x_i)-\frac{1}{C}|}
\end{equation}  
, where $x_i$ is a sample, $B$ is the decision boundary of a NN, and $C$ is the number of sample classifications. $p_k$ is the probability that the $x_i$ belongs to the $k$th classification, which is the NN output owing to its activation function "softmax". 

After each classification MLP was trained, we utilized USDB to generate samples around the decision boundary of the MLP. We then fed these samples into the MLP and collected each NN layer's input and output as the MLP explanation dataset. We finally ran the MNNCGP-ES, LIME, and MAPLE 30 times on the explanation dataset.

% we collected each middle NN layer's input and output data of the 5 MLPs, and use USDB to generate samples around the decision boundary of the output layer in each MLP. We combine the middle layer data with samples We utilizes USDB to generate samples around the decision boundaries of the five trained NN. We then run SRNet, LIME, and MAPLE 30 times on these samples. We use the input an trained  These samples lets them converge to the "real" decision boundary of the MLP. 

% Figure \ref{fig:SamplingDB}(a) shows the decision boundary of a blackbox classifier in a dataset with 4 classes. The color and the shape of each instance represent the different classes. The decision boundaries are shown as black-dashdot lines. After we applied the features standardization and the sampling method according to Equation \ref{eq:distance}, we could enlarge the original dataset with lots of new instances around DB, as shown in Figure \ref{fig:SamplingDB}(b). Since the new instances of the enlarged dataset are all located near the DB, they could help the SRNet learns the classification strategy of NNs, so that SRNet can converge faster. Note that it must be admitted that USDB is only efficient in the dataset with low and medium dimensional features since the number of demands for $n$ and $s$ will increase sharply with the number of input features.

% After we enlarged the classification SRNet dataset, we ran MNNCGP with parameters similar to the settings of the regression task, except that we added a softmax layer after the last NNCGP to achieve classification for SRNet. 

\section{Result and Analysis}

\subsection{Regression Task}
In the following regression tasks we only show results on the six benchmarks, $K0-K3$, $F0$, and $F1$. The regression results on all benchmarks are in the appendix.
\subsubsection{Fitness Convergence}

\begin{figure}[htbp]
    \centering
        \includegraphics[width=0.9\linewidth]{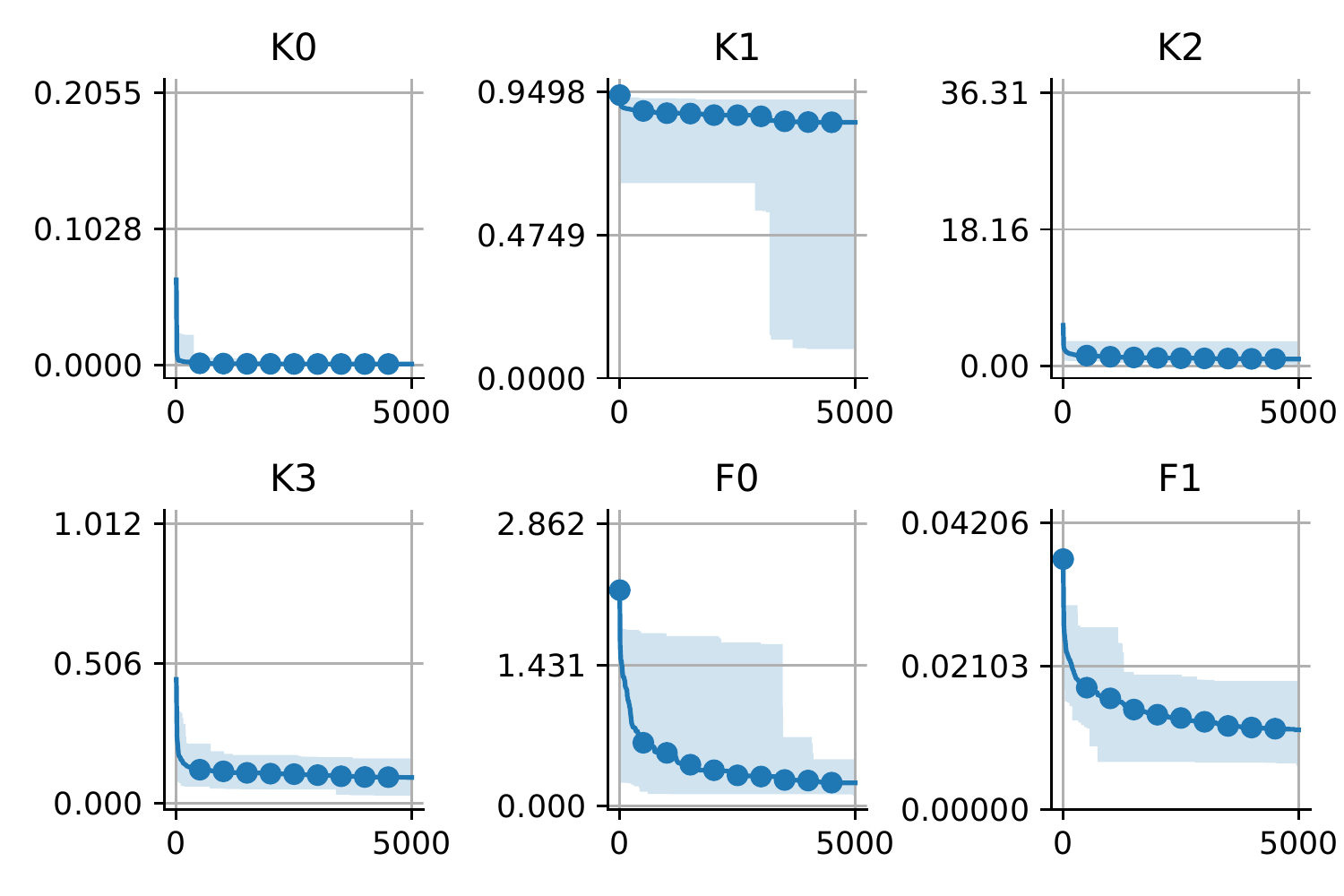}
    \caption{MNNCGP-ES convergence curve. 
    % that run 30 times on each MLP for $K0-K3$, $F0$, and $F1$. 
    % The blue line represents the average fitness. The light blue range is the fitness range (maximum and minimum values) of the 30 results. 
    % Convergence results on all benchmarks in the appendix
    }
    % \vspace{-0.5cm}
    \label{fig:trend_result}
\end{figure}

Figure \ref{fig:trend_result} illustrates the convergence curve of the fitness scores of MNNCGP-ES on each MLP for the regression tasks. The blue line is the average fitness score in 30 experiments. It gradually decreases at the beginning and trends to a flat curve. It means that, statistically, MNNCGP-ES could find the mathematical expressions that approximate the hidden semantics of each layer in NNs according to Equation \ref{eq:fitness}, as shown in Table \ref{tab:inter_expressions}. It also indicates the feasibility to use the combination of these mathematical expressions to represent the whole semantics of each MLP, as shown in Table \ref{tab:overall_expressions}. 
% Note that Table \ref{tab:inter_expressions} and Table \ref{tab:overall_expressions} only list expressions of some of NNs, see \ref{appendix_sec:expressions} for more details.
\begin{table}[htpb]
    \centering
    \scalebox{0.6}{
    \begin{tabular}{c|c|c|c|c}
        \hline
            \textbf{Dataset} & $\boldsymbol{h_0^s(x)}$ & $\boldsymbol{h_1^s(h_0^s)}$ & $\boldsymbol{h_2^s(h_1^s)}$ & $\boldsymbol{y^s}$  \\
        \hline
        $K0$ & \makecell[c]{$sin(0.26x +$\\$sin(sin(x))-$\\$0.068)$\\\textbf{(6.74e-04)}} & \makecell[c]{$0.88 - \cos{(h^s_0)_0}$\\\textbf{(7.54e-05)}} & $-$ & \makecell[c]{$-sin(sin((h^s_0)_0$\\$-0.36))$\\\textbf{(2.37e-04)}} \\
             
        $K1$ & \makecell[c]{5.15e-05$x_1 -$\\$ 0.0072sin(x_0) -$\\5.15e-05\\\textbf{(5.34e-02)}} & \makecell[c]{$-(h^s_0)_0 + (h^s_0)_2$\\\textbf{(1.39e-03)}} & $-$ & \makecell[c]{$-(h^s_1)_0 + $\\$(h^s_1)_2 + $\\$0.0011$\\\textbf{(6.88e-02)}} \\
        
        $K2$ & \makecell[c]{$\frac{\sqrt{x^{2}} \log{\left(x^{2} \right)}}{x + 1.20}$\\\textbf{(3.26e-02)}} & \makecell[c]{$- tan(((h^s_0)_{2} +$\\$(h^s_0)_{3}^{2} -$\\$(h^s_0)_{3}))$\\\textbf{(1.04e-02)}} & $-$ & \makecell[c]{$\sin{\left(\frac{(h^s_1)_{0}}{(h^s_1)_{2}^{2}} \right)}$\\\textbf{(2.00e-01)}} \\
        
        $K3$ & \makecell[c]{$- sin((0.24 x_{0} +$\\$ \sin{\left(0.23 x_{1} \right)}))$\\\textbf{(3.86e-02)}} & \makecell[c]{$(h^s_0)_{1} (h^s_0)_{2}^{2}$\\$ \sin{\left((h^s_0)_{1} \right)}$\\\textbf{(3.64e-03)}} & \makecell[c]{$(h^s_1)_{0} (- (h^s_1)_{2} +$\\$ (h^s_1)_{3})$\\\textbf{(2.87e-04)}} & \makecell[c]{$-0.99 + \frac{(h^s_2)_{3}}{(h^s_2)_{0}}$\\\textbf{(1.32e-02)}} \\
        
        % $K4$ & \makecell[c]{$- \sin{(x_{1})} + $\\ $\sin{(\frac{\sin{(x_{0})}}{x_{2}})}$\\\textbf{(1.17e-02)}} 
        % &\makecell[c]{$((h^s_0)_{0} + (h^s_0)_{3}$\\$ - \tan{\left((h^s_0)_{1} \right)})^{2}$\\\textbf{(1.31e-02)}} & $-$ & \makecell[c]{$- \frac{\sin{\left((h^s_1)_{1} - (h^s_1)_{2} \right)}}{(h^s_1)_{0}^{2}}$\\\textbf{(6.14e-01)}} \\
        
        % $K5$ & \makecell[c]{$x_{1} - 0.097$\\\textbf{(9.01e-02)}} & \makecell[c]{$\frac{\sin{\left((h^s_0)_{3} - 0.48 \right)}}{(h^s_0)_{2}}$\\\textbf{(2.44e-02)}} & $-$ & \makecell[c]{$\frac{\tan{\left((h^s_1)_{0} + 0.51 \right)}}{(h^s_1)_{2} - 0.26}$\\\textbf{(4.56)}} \\
        
        $F0$ & \makecell[c]{$cos((\frac{0.67 x_{1}}{x_{2}} + $\\$\log{(x_{0})}))$\\\textbf{(6.35e-03)}} & \makecell[c]{$- (h^s_0)_{1} + $\\$\cos{\left((h^s_0)_{2} \right)}$\\\textbf{(1.04e-02)}} & $-$ & \makecell[c]{$tan((tan(((h^s_1)_{0} $\\$- 0.33)) + 0.11 ))$\\\textbf{(9.63e-02)}} \\ 
             
        $F1$ & \makecell[c]{$\log{\left(\frac{x_{0} x_{1}}{x_{2} x_{3}^{\frac{3}{2}}} \right)}$\\\textbf{(9.59e-03)}} & \makecell[c]{$(h^s_0)_{1}^{4}$\\\textbf{(2.46e-04)}} & $-$ & \makecell[c]{$(h^s_1)_{0} + \sin{\left((h^s_1)_{0} \right)}$\\\textbf{(1.45e-03)}} \\
        
        % $F2$ & \makecell[c]{$- \frac{x_{4}}{x_{3}} + \frac{x_{4}}{x_{2}}$\\\textbf{(1.42e-02)}} & \makecell[c]{$(h^s_0)_{2} \left((h^s_0)_{1} + 1\right)$\\\textbf{(1.36e-02)}} & $-$ & \makecell[c]{$\frac{0.082 - (h^s_1)_{2}}{(h^s_1)_{0}}$\\\textbf{(1.15e02)}} \\
        
        % $F3$ & \makecell[c]{$sin((\sqrt[4]{2} \sqrt[4]{x_{0}} +$\\$ \sqrt{x_{1}}))$\\\textbf{(1.22e-02)}} & \makecell[c]{$\tan{\left((h^s_0)_{1} \right)} $\\$- 0.30$\\\textbf{(1.64e-02)}} & $-$ & \makecell[c]{$(h^s_1)_{1} - $\\$\cos{\left(4 (h^s_1)_{2}^{2} \right)}$\\\textbf{(5.71)}} \\
        
        % $F4$ & \makecell[c]{$x_{1} x_{4} - \sqrt{x_{2} + x_{3}}$\\\textbf{(7.49e-02)}} & \makecell[c]{$(h^s_0)_{4} \cos^{2}{\left((h^s_0)_{1} \right)}$\\\textbf{(3.61e-02)}} & $-$ & \makecell[c]{$(h^s_1)_{0}^{2} (h^s_1)_{4}^{2}$\\\textbf{(1.21e08)}} \\
        
        % $F5$ & \makecell[c]{$\sqrt{x_{1}} + \log{\left(\frac{x_{1}}{x_{0} x_{2}} \right)}$\\\textbf{(3.84e-03)}} & \makecell[c]{$\sin^{2}{\left(0.51 (h^s_0)_{0} \right)}$\\\textbf{(1.20e-03)}} & \makecell[c]{$\frac{\left(2 (h^s_1)_{1} - \tan{\left((h^s_1)_{1} \right)}\right)^{4}}{(h^s_1)_{3}^{2}}$\\\textbf{(5.12e-03)}} & \makecell[c]{$\tan{\left(\tan{\left((h^s_2)_{0} \right)} \right)}$\\\textbf{(5.75e01)}} \\
    
        \hline 
    \end{tabular}  
    }
    \caption{ The mathematical expressions of each layer in NNs. 
    % $y^s$ is the output layer in the NN. The number below each $f_i$ is its fitness score.
    }
    \label{tab:inter_expressions}
\end{table}

\begin{table}[t]
    \centering
    \scalebox{0.7}{
    \begin{tabular}{c|c}
        \hline
            \textbf{Dataset} & $\boldsymbol{O^s(x)}$\\
        \hline
        $K0$ & \makecell[c]{$0.29 - 4.01 sin((sin((2.36 cos((0.41 sin((0.26 x +$\\$\sin{\left(\sin{\left(x \right)} \right)} - 0.068)) + 0.49)) - 2.03))))$\\\textbf{(6.11e-04)}}\\
        
        $K1$ & \makecell[c]{$- 0.01 x_{1} + 1.69 \sin{\left(x_{0} \right)} + 0.0021$\\\textbf{(9.62e-02)}}\\
        
        $K2$ & \makecell[c]{$4.49 \sin{\left(\frac{6.70 \left(0.51 \tan{\left(0.061 \left(1 - \frac{0.19 \sqrt{x^{2}} \log{\left(x^{2} \right)}}{x + 1.20}\right)^{2} - 0.044 + \frac{0.084 \sqrt{x^{2}} \log{\left(x^{2} \right)}}{x + 1.20} \right)} + 0.27\right)}{\left(\tan{\left(0.061 \left(1 - \frac{0.19 \sqrt{x^{2}} \log{\left(x^{2} \right)}}{x + 1.20}\right)^{2} - 0.044 + \frac{0.084 \sqrt{x^{2}} \log{\left(x^{2} \right)}}{x + 1.20} \right)} + 0.63\right)^{2}} \right)}$\\$ + 7.90$\\\textbf{(2.22e-01)}}\\
        
         %$K3$ & \makecell[c]{$17.08 (- 0.81 ((- 1.28 (0.57 - 0.55 \sin{(0.24 x_{0} + \sin{(0.23 x_{1} )} )}) $\\$(0.93 \sin{(0.24 x_{0} + \sin{(0.23 x_{1} )} )} + 1)^{2}$\\$ \sin{(0.55 \sin{(0.24 x_{0} + \sin{(0.23 x_{1} )} )} - 0.57 )}$\\$ + 0.39))((- 0.88 (0.57 - 0.55 \sin{(0.24 x_{0} + \sin{(0.23 x_{1} )} )}) $\\$(0.93 \sin{(0.24 x_{0} + \sin{(0.23 x_{1} )} )} + 1)^{2}$\\$\sin{(0.55 \sin{(0.24 x_{0} + \sin{(0.23 x_{1} )}))} - 0.57 )}$\\$ - 0.12) + 0.43)\div (- 0.51 (- 1.28 (0.57 - 0.55 \sin{(0.24 x_{0} + \sin{(0.23 x_{1} )} )}) $\\$(0.93 \sin{(0.24 x_{0} + \sin{(0.23 x_{1} )} )} + 1)^{2} $\\$\sin{(0.55 \sin{(0.24x_{0} + \sin{(0.23 x_{1} )} )} - 0.57 )}$\\$ + 0.39) (- 0.88 (0.57 - 0.55 \sin{(0.24 x_{0} + \sin{(0.23 x_{1} )} )}) $\\$(0.93 \sin{(0.24 x_{0} + \sin{(0.23 x_{1} )} )} + 1)^{2} $\\$\sin{(0.55 \sin{(0.24 x_{0} + \sin{(0.23 x_{1} )} )} - 0.57 )} - 0.12) + 0.44) - 15.36$\\\textbf{(2.73e-02)} }\\
         
         $K3$ & \makecell[c]{Too long (See appendix for details)\\\textbf{(2.73e-02)} }\\

        %  $K4$ & \makecell[c]{$-0.0080 + 1.78 \sin((2.43 (0.19 \sin{\left(x_{1} \right)} - 0.19 \sin{\left(\frac{\sin{\left(x_{0} \right)}}{x_{2}} \right)} -$\\$ \tan{\left(0.020 \sin{\left(x_{1} \right)} - 0.020 \sin{\left(\frac{\sin{\left(x_{0} \right)}}{x_{2}} \right)} + 0.73 \right)} + 0.90)^{2} - 0.026))$\\$\div (0.35 - (0.19 \sin{\left(x_{1} \right)} - 0.19 \sin{\left(\frac{\sin{\left(x_{0} \right)}}{x_{2}} \right)} $\\$- \tan{\left(0.20 \sin{\left(x_{1} \right)} - 0.020 \sin{\left(\frac{\sin{\left(x_{0} \right)}}{x_{2}} \right)} + 0.73 \right)} + 0.90)^{2})^{2}$\\\textbf{(6.27e-01)}}\\

        % $K5$ & \makecell[c]{$-0.14 + \frac{0.011 \tan{\left(1.42 - \frac{0.051 \sin{\left(0.25 x_{1} - 0.14 \right)}}{0.18 - 0.020 x_{1}} \right)}}{0.16 - \frac{0.0414421632885933 \sin{\left(0.25 x_{1} - 0.14 \right)}}{0.18 - 0.020 x_{1}}}$\\\textbf{(4.62)}}\\
        
        $F0$ & \makecell[c]{$3.05 - 7.00 tan((tan((0.58 \cos{\left(\log{\left(m_{0} \right)} + \frac{0.67 v}{c} \right)} - $\\$0.76 \cos{\left(0.22 \cos{\left(\log{\left(m_{0} \right)} + \frac{0.67 v}{c} \right)} - 1.01 \right)} + 0.35)) - 0.11))$\\\textbf{(1.05e-01)}} \\

        $F1$ & \makecell[c]{$0.022 \left(0.46 \log{\left(\frac{q_{1} q_{2}}{e r^{\frac{3}{2}}} \right)} + 1\right)^{4} + 3.22 \sin{\left(0.0068 \left(0.46 \log{\left(\frac{q_{1} q_{2}}{e r^{\frac{3}{2}}} \right)} + 1\right)^{4} - 0.00072 \right)}$\\$ + 0.0059$\\\textbf{(6.37e-03)}}\\

        \hline 
    \end{tabular}    
    }    
    \caption{The mathematical expression of each whole NN.}
    \label{tab:overall_expressions}
\end{table}

Not all ranges of fitness scores (light blue areas) become smaller as the MNNCGP-ES runs, such as $k1$. However, the low bounds of these lines always become smaller and trend to be zero at the later stage. It means that the more episodes the MNNCGP-ES runs, the more likely MNNCGP-ES is able to find the mathematical expressions that can be used to explain the hidden semantics of a NN. The slow decrease of fitness curves also indicates the need to run MNNCGP-ES with sufficient times to obtain the best-fitted mathematical expressions.

\subsubsection{Semantics Evaluation}

% \begin{figure}[htbp]
%     \centering
%     \includegraphics[width=\linewidth]{hidden_output9.pdf}
%     \caption{ 
% Each group of 9 heat maps represents the comparison of outputs of the SRNet layer vs the NN layer with 9 random input values.
%     }
%     \label{fig:hidden_heat_map}
% \end{figure}

\begin{figure}[ht]
	\centering
% 	\begin{subfigure}{0.24\linewidth}
% 		\includegraphics[width=\linewidth]{kkk0_0.pdf}
% 	\end{subfigure}
	\begin{subfigure}{0.49\linewidth}
	    \caption{K0-h1}
		\includegraphics[width=\linewidth]{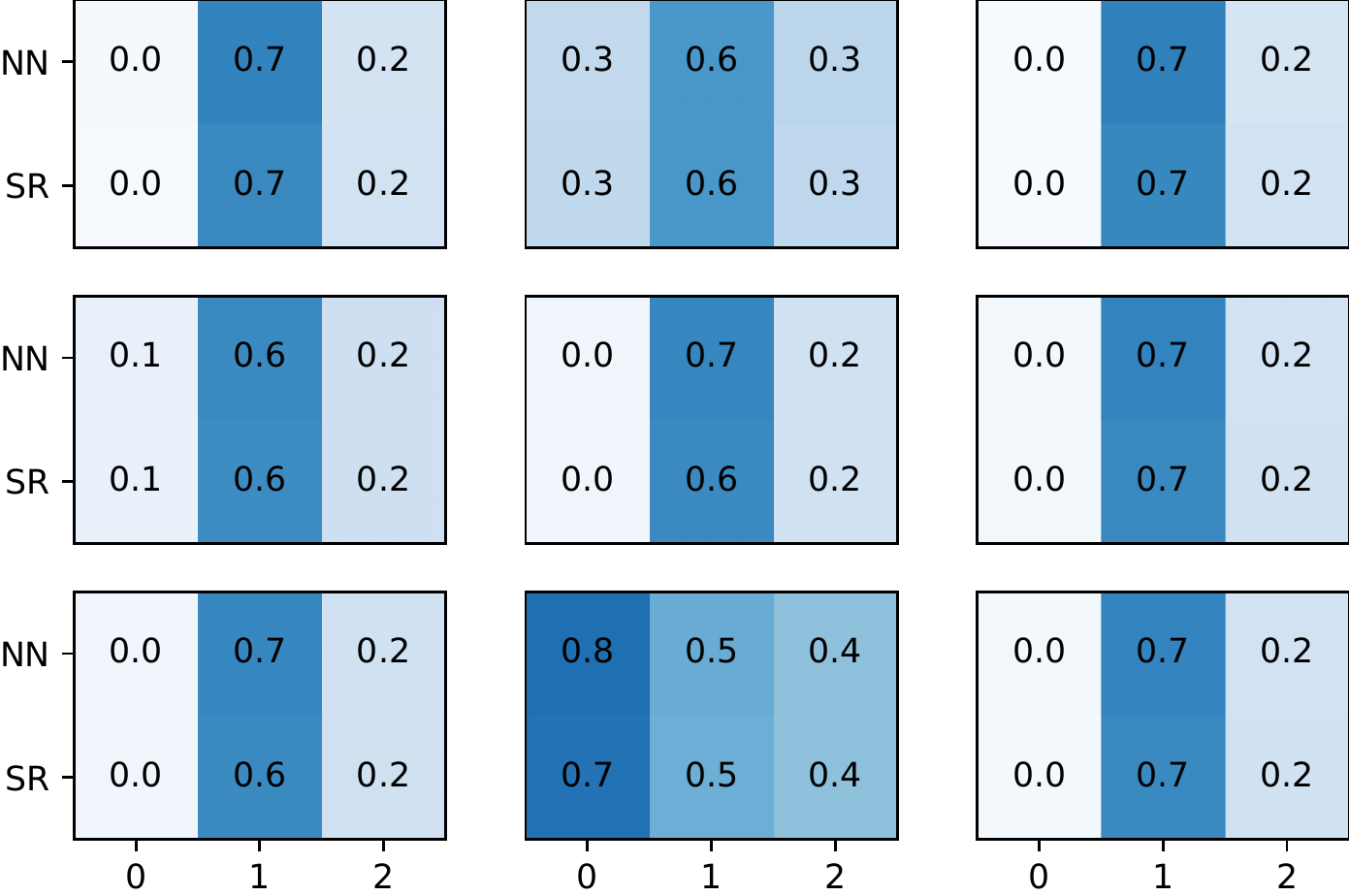}
	\end{subfigure}
% 	\begin{subfigure}{0.24\linewidth}
% 		\includegraphics[width=\linewidth]{KKK1_0.pdf}
% 	\end{subfigure}
	\begin{subfigure}{0.49\linewidth}
	    \caption{K1-h1}
		\includegraphics[width=\linewidth]{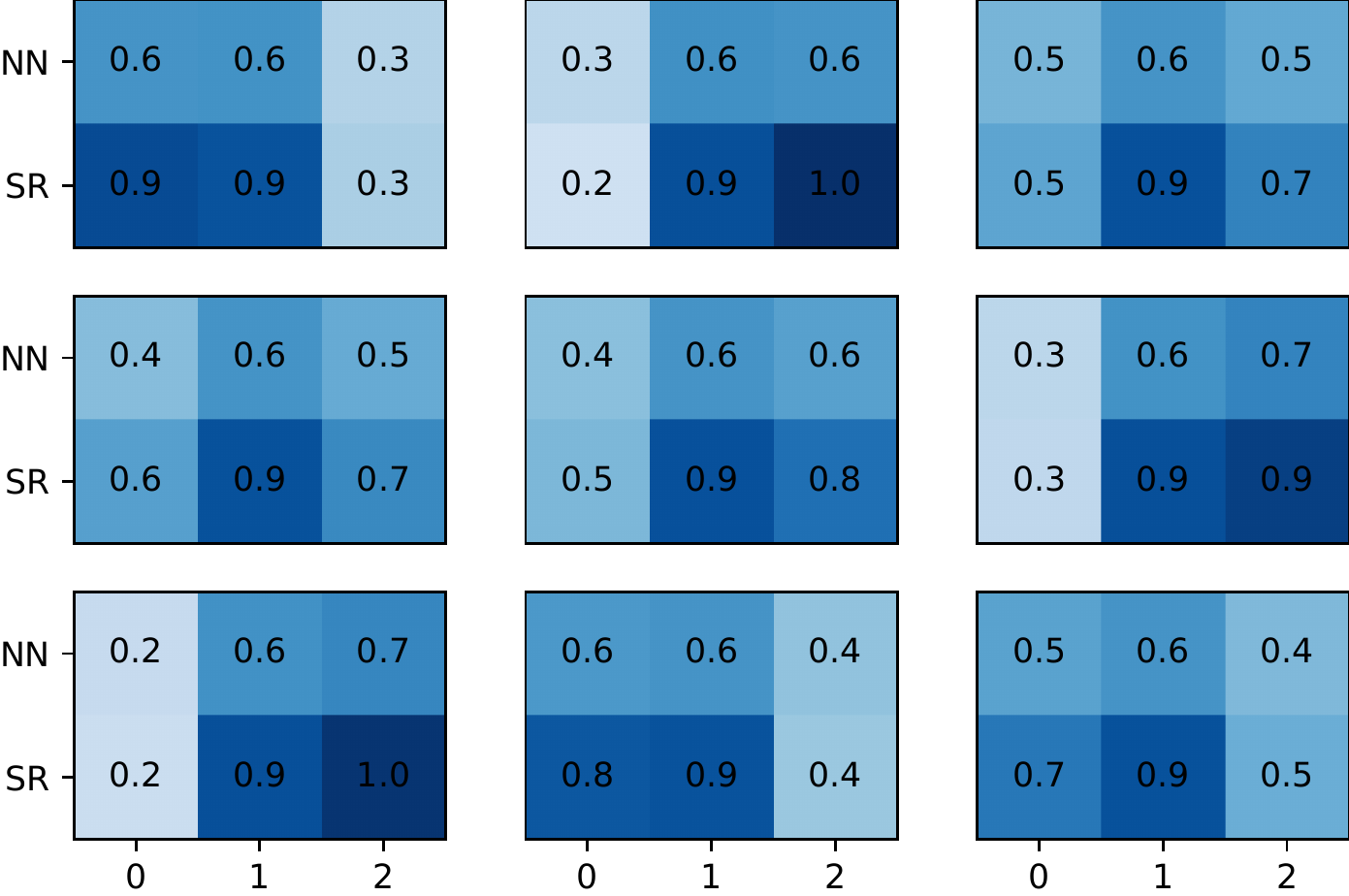}
	\end{subfigure}
 \\
% 	\begin{subfigure}{0.24\linewidth}
% 		\includegraphics[width=\linewidth]{KKK2_0.pdf}
% 	\end{subfigure}
	\begin{subfigure}{0.49\linewidth}
	    \caption{K2-h1}
		\includegraphics[width=\linewidth]{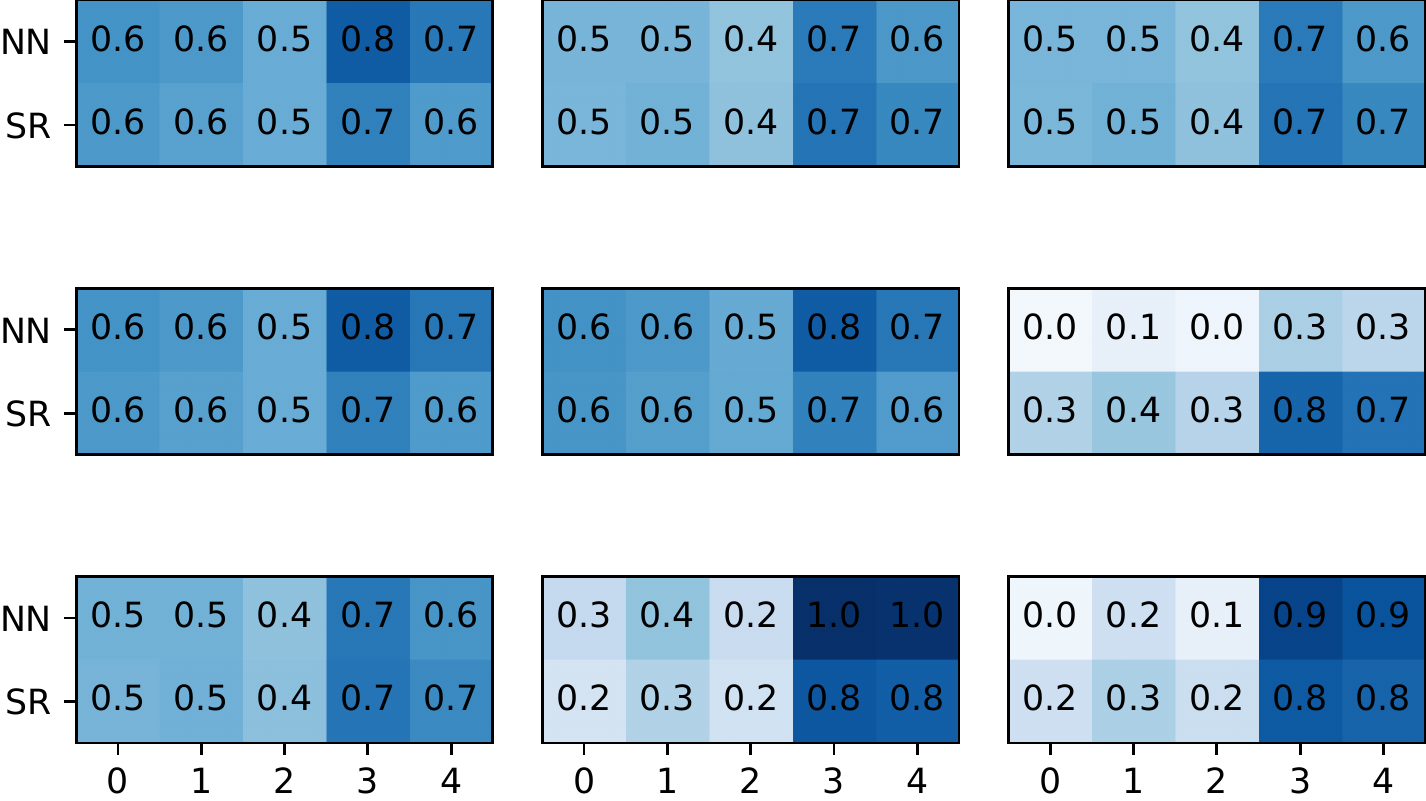}
	\end{subfigure}
% 	\begin{subfigure}{0.16\linewidth}
% 		\includegraphics[width=\linewidth]{KKK3_0.pdf}
% 	\end{subfigure}
	\begin{subfigure}{0.49\linewidth}
	    \caption{K3-h1}
		\includegraphics[width=\linewidth]{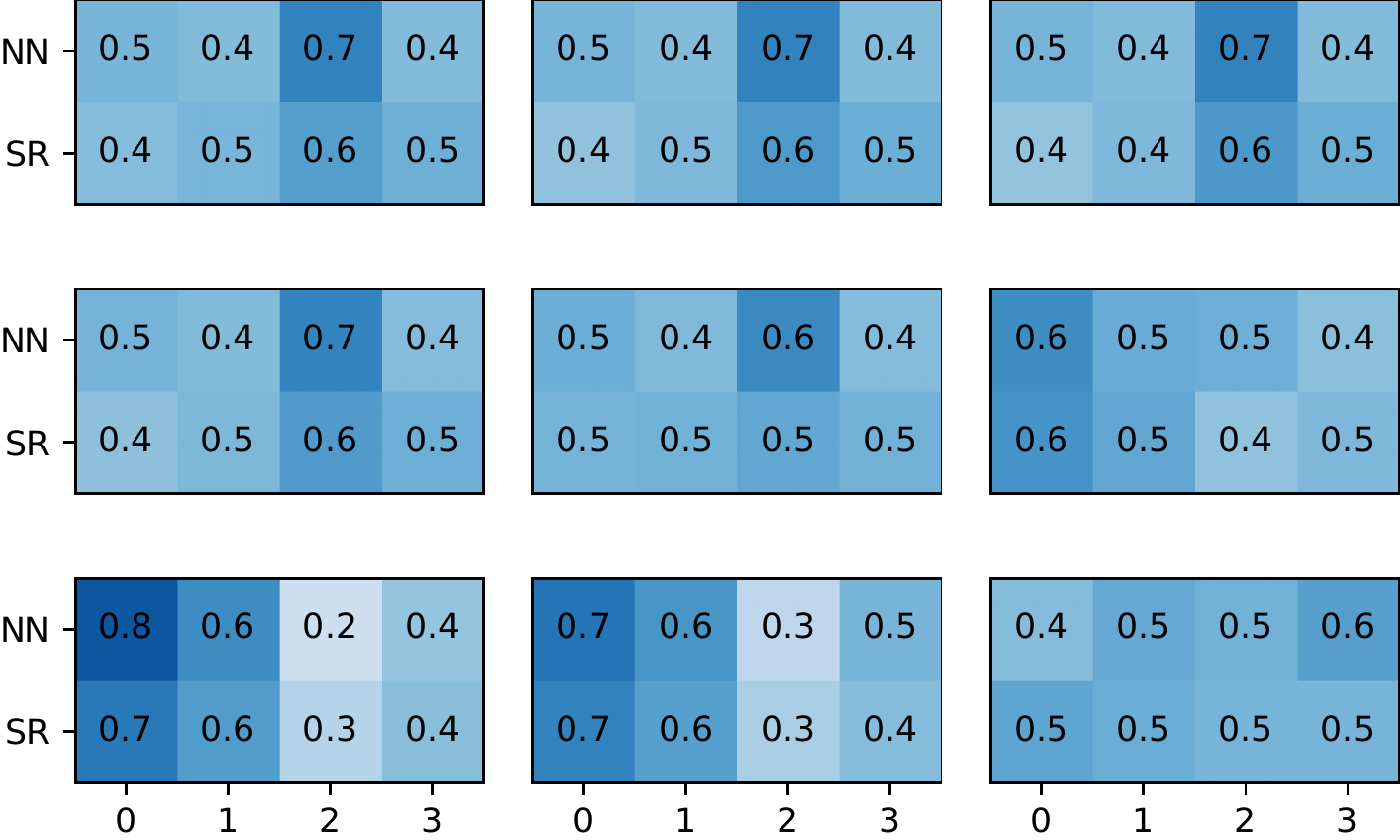}
	\end{subfigure}
% 	\begin{subfigure}{0.16\linewidth}
% 		\includegraphics[width=\linewidth]{KKK3_2.pdf}
% 	\end{subfigure}
\\
% 	\begin{subfigure}{0.24\linewidth}
% 		\includegraphics[width=\linewidth]{kkk4_0.pdf}
% 	\end{subfigure}
% 	\begin{subfigure}{0.24\linewidth}
% 		\includegraphics[width=\linewidth]{kkk4_1.pdf}
% 	\end{subfigure}
% 	\begin{subfigure}{0.24\linewidth}
% 		\includegraphics[width=\linewidth]{KKK5_0.pdf}
% 	\end{subfigure}
% 	\begin{subfigure}{0.24\linewidth}
% 		\includegraphics[width=\linewidth]{KKK5_1.pdf}
% 	\end{subfigure}
% \\
% 	\begin{subfigure}{0.24\linewidth}
% 		\includegraphics[width=\linewidth]{feynman0_0.pdf}
% 	\end{subfigure}
	\begin{subfigure}{0.49\linewidth}
	    \caption{F0-h1}
		\includegraphics[width=\linewidth]{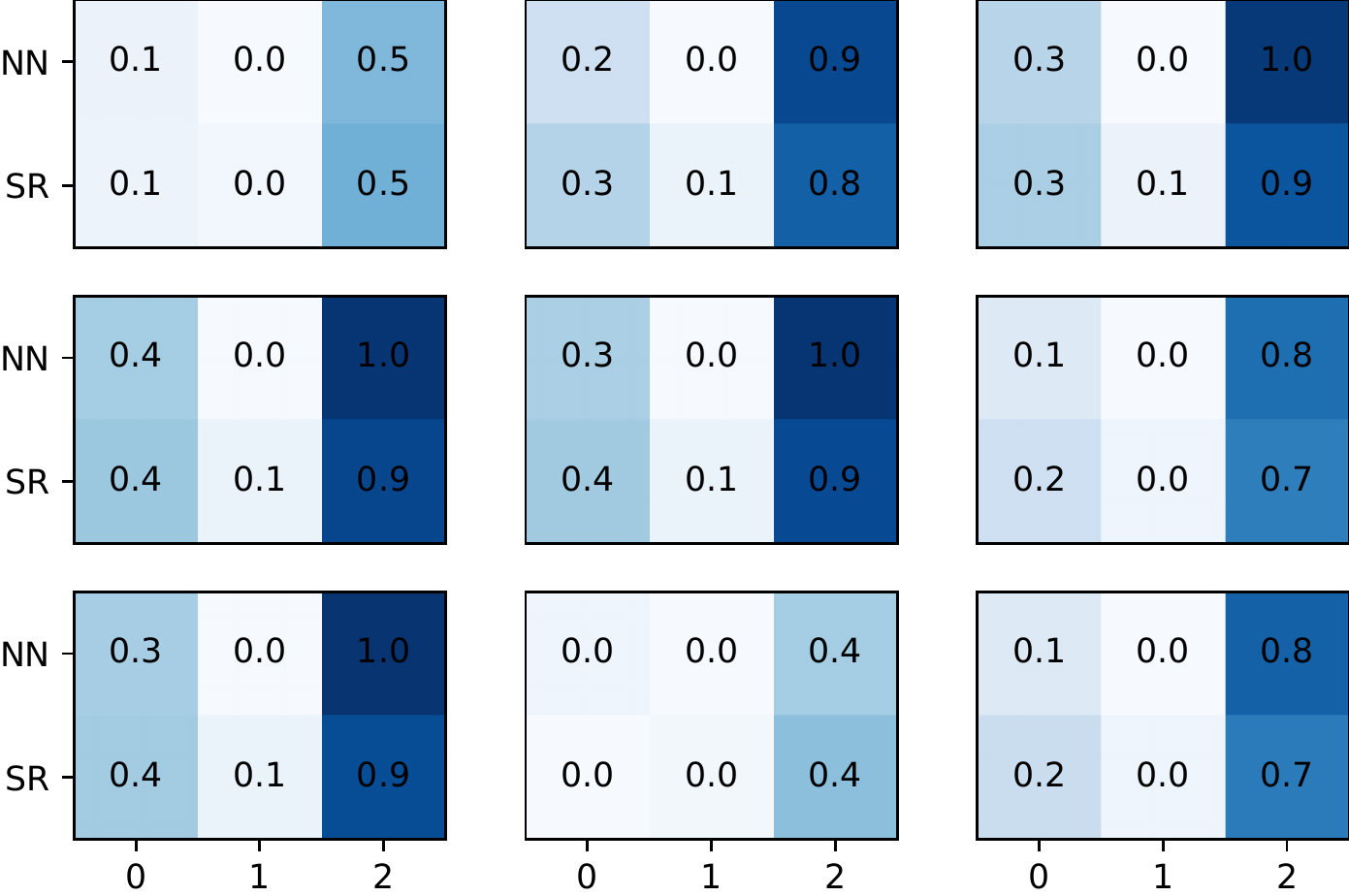}
	\end{subfigure}
% 	\begin{subfigure}{0.24\linewidth}
% 		\includegraphics[width=\linewidth]{feynman1_0.pdf}
% 	\end{subfigure}
	\begin{subfigure}{0.49\linewidth}
	    \caption{F1-h1}
		\includegraphics[width=\linewidth]{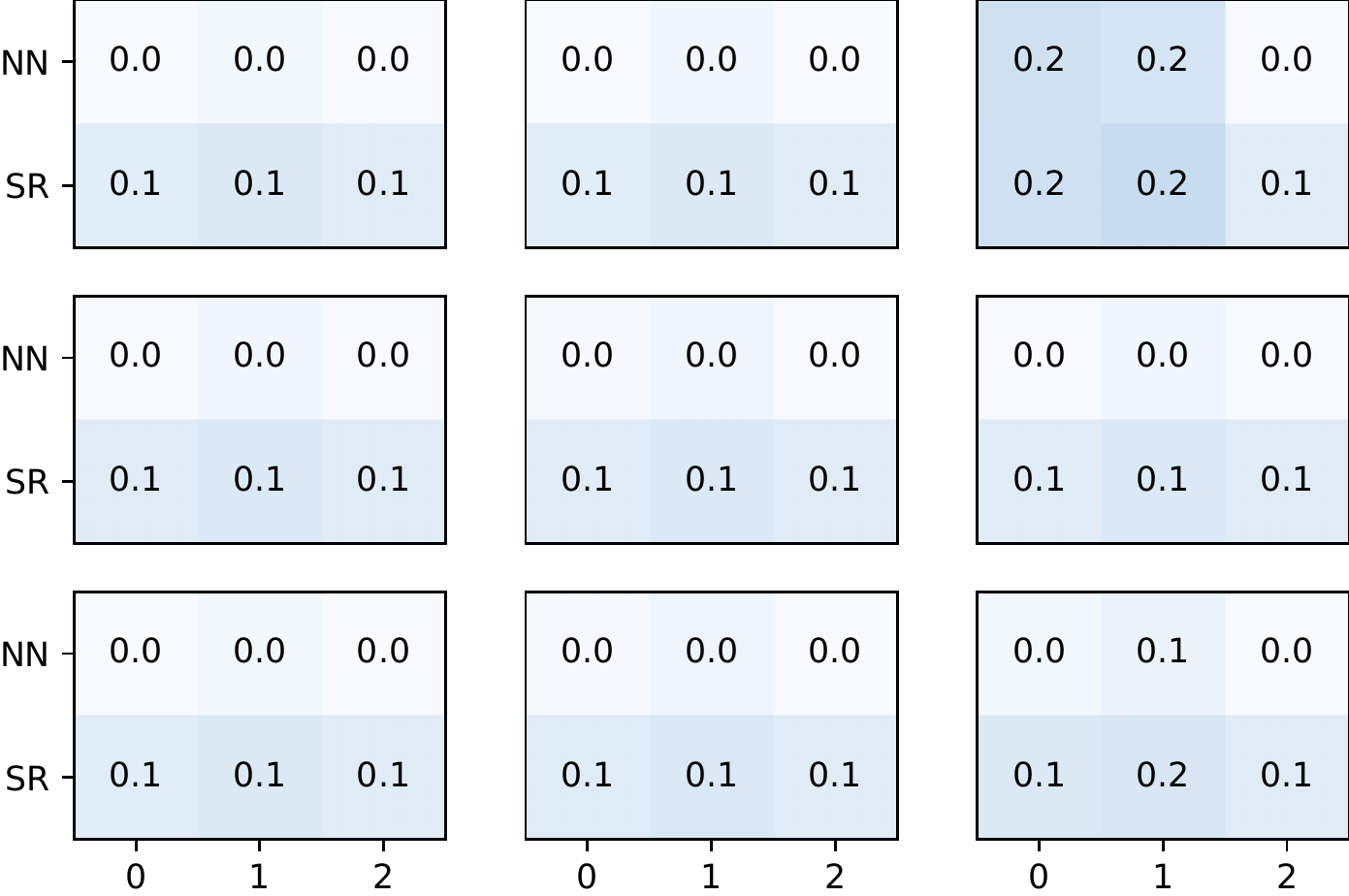}
	\end{subfigure}
\\
% 	\begin{subfigure}{0.24\linewidth}
% 		\includegraphics[width=\linewidth]{feynman2_0.pdf}
% 	\end{subfigure}
% 	\begin{subfigure}{0.24\linewidth}
% 		\includegraphics[width=\linewidth]{feynman2_1.pdf}
% 	\end{subfigure}
% 	\begin{subfigure}{0.24\linewidth}
% 		\includegraphics[width=\linewidth]{feynman3_0.pdf}
% 	\end{subfigure}
% 	\begin{subfigure}{0.24\linewidth}
% 		\includegraphics[width=\linewidth]{feynman3_1.pdf}
% 	\end{subfigure}
% \\
% 	\begin{subfigure}{0.24\linewidth}
% 		\includegraphics[width=\linewidth]{feynman4_0.pdf}
% 	\end{subfigure}
% 	\begin{subfigure}{0.24\linewidth}
% 		\includegraphics[width=\linewidth]{feynman4_1.pdf}
% 	\end{subfigure}
% 	\begin{subfigure}{0.16\linewidth}
% 		\includegraphics[width=\linewidth]{feynman5_0.pdf}
% 	\end{subfigure}
% 	\begin{subfigure}{0.16\linewidth}
% 		\includegraphics[width=\linewidth]{feynman5_1.pdf}
% 	\end{subfigure}
% 	\begin{subfigure}{0.16\linewidth}
% 		\includegraphics[width=\linewidth]{feynman5_2.pdf}
% 	\end{subfigure}
% 	\caption{Each group of 9 heat maps represents the comparison of outputs of the SRNet layer vs the NN layer with 9 random input values.}
    	\caption{The outputs of the SRNet layer vs the NN layer with 9 random input values.}
	\label{fig:hidden_heat_map}
\end{figure}

The results show that the proposed SRNet method can acquire a fitted mathematical expression to explain hidden semantics of each layer, as shown in Table \ref{tab:inter_expressions}. Each of these mathematical expressions represents the general semantics $f_i$ in the expression $w_i^sf_i(h_{i-1}^s)+b_i$ (Equation \ref{eq:layermodel}). The number below each $f_i$ is its fitness. For example, for $K_1$, MNNCGP-ES finds the general semantics $-(h_0^s)_0+(h_0^s)_2$ at the second hidden layer $h_1$. The fitness of $w_i^s\times(-(h_0^s)_0+(h_0^s)_2)+b_i$ is $1.39e-03$. All fitness values in the hidden layers are less than 0.1. It means that $f_i$ captured by MNNCGP-ES can represent (approximate) the semantics of each hidden layer in a NN.

Figure \ref{fig:hidden_heat_map} shows the details of mathematical expressions that MNNCGP-ES finds. The expressions approximate the output of each layer in a NN. To evaluate each layer, we input 9 random values into the mathematical expression and the hidden nodes in the NN, respectively. We then obtained 9 heat maps to show the difference between the output of the mathematical expression and that of the NN hidden nodes. For example, the sub-figure "$K1-h1$" represents the $K1$ outputs of SRNet vs NN in the layer $h1$ with 9 random input values. In the first heat map in "$K1-h1$", $[0.6,0.6,0.3]$ are the outputs of three hidden nodes in the NN layer $h1$ with the first input value, while $[0.9,0.9,0.3]$ are the output of the mathematical expression in the SRNet layer $h1$. So, Figure \ref{fig:hidden_heat_map} explains why a mathematical expression has low fitness, and another has high fitness. Comparing $K1-h1$ with $K0-h1$ in Figure \ref{fig:hidden_heat_map}, the outputs between NN and SRNet in $K0-h1$ are closer than the output between them in $K1-h1$. Thus, the fitness "$7.54e-05$" of "$0.88-cos(h_0^s)_0$" is less than "$1.39e-03$" of " $-(h_0^s)_0+(h_0^s)_2$" in Table \ref{tab:inter_expressions}.       

For the output layers in the NNs, the last column $y^s$ in Table \ref{tab:inter_expressions} lists the mathematical expressions to present the outputs. Their fitness scores of the output layer are better than the scores of the previous layers. There are two formulas of $K5$ and $F5$ whose fitness scores are greater than 1. The reason causing the higher scores at the output layers is that, in a NN, the network structure of the outer layer is different from the hidden layers. The computation function on the output layer is $y=W_{i+1}h_i+b_{i+1}$, while that on the hidden layer is $h_i=W_{i}h_{i-1}+b_{i}$. Since $y$ is a number and $h_i$ is a vector, $y=W_{i+1}h_i+b_{i+1}$ is a multivariate linear equation. However, $y^s = w_{i+1}^s f_{i+1}(h_i^s)+b_{i+1}$ on the output layer in SRNet is one-variable linear equation because $ w_{i+1}^s$ and $b_{i+1}$ are two numbers, not vectors. Although MNNCGP-ES can find a fitted mathematical function $f_{i+1}$ to represent the NN layers, the one-variable linear equation is still hard to approximate the multi-variable linear equation in the layers.

Table \ref{tab:overall_expressions} lists the mathematical expressions that represent the whole NN semantics for the regression tasks. Each of final expressions is obtained by combining the mathematical expressions in different layers shown in Table \ref{tab:inter_expressions}. The complexity and length of the mathematical expressions could increase substantially, such as the mathematical expression in $K3$, due to the combination of expressions on every layer. If there are more layers in a NN, the length of the mathematical expression could be longer. Although the math representation generated by SRNet could be lengthy, it provides a straightforward expression to show all layers' hidden semantics of the whole NN. 

\subsubsection{Performance Comparison}
To evaluate the SRNet performance on regression tasks, we ran and compared SRNet, LIME \cite{ribeiro2016should}, and MAPLE \cite{plumb2018model} on an interpolation dataset and an extrapolation dataset. The interpolation dataset consists of the NN input-output values. In contrast, the extrapolation dataset consists of the data sampled directly from the original symbolic expression in the column "Function" in Table \ref{tab:dataset}. The interpolation domain is the same as the range of the training dataset shown in Table \ref{tab:dataset}. The size of the extrapolation domain is five times that of the interpolation domain. 

\begin{figure}[ht]
	\centering
	\begin{subfigure}{0.32\linewidth}
		\includegraphics[width=\linewidth]{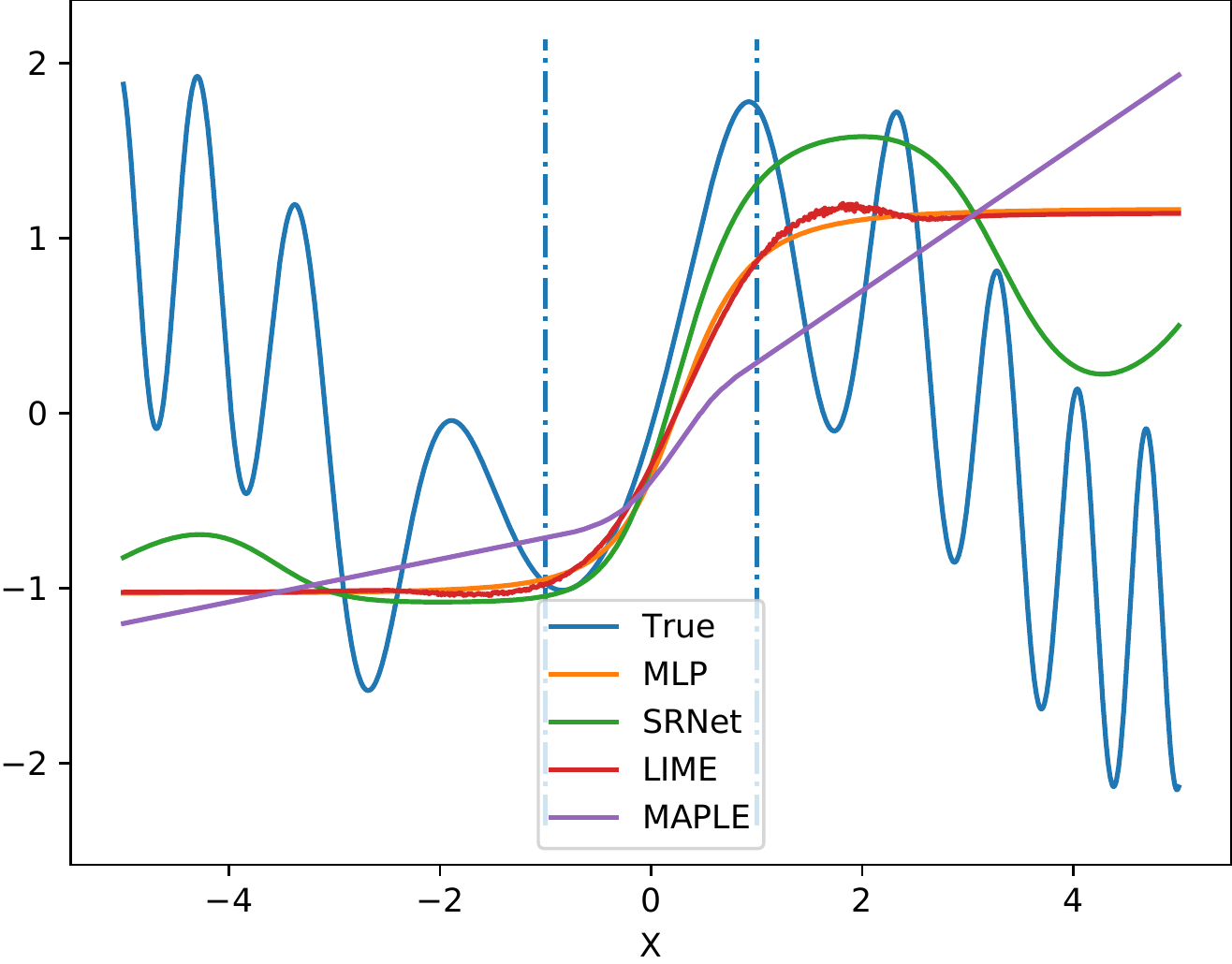}
		\caption{K0}
	\end{subfigure}
	\begin{subfigure}{0.32\linewidth}
		\includegraphics[width=\linewidth]{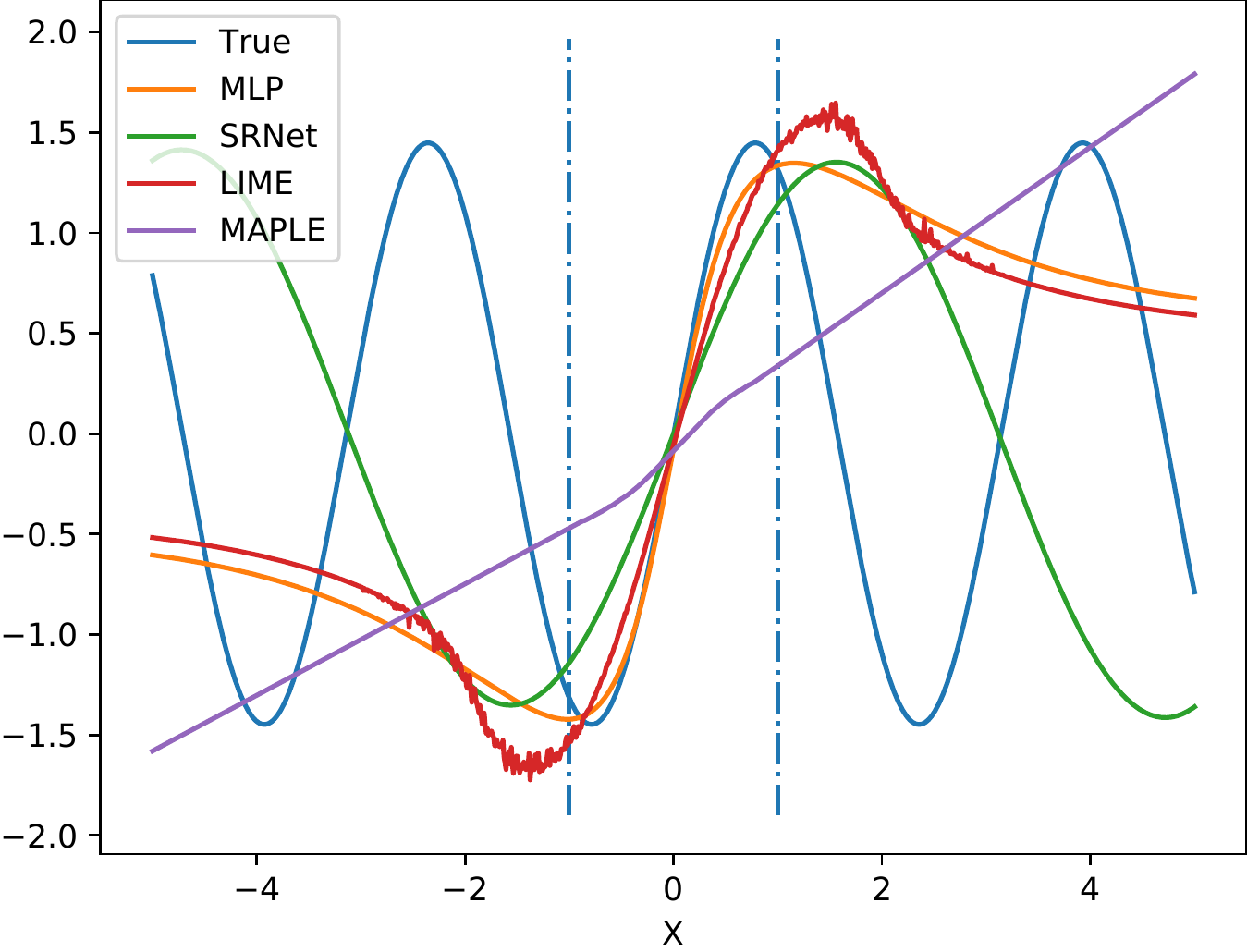}
		\caption{K1}
	\end{subfigure}
	\begin{subfigure}{0.32\linewidth}
		\includegraphics[width=\linewidth]{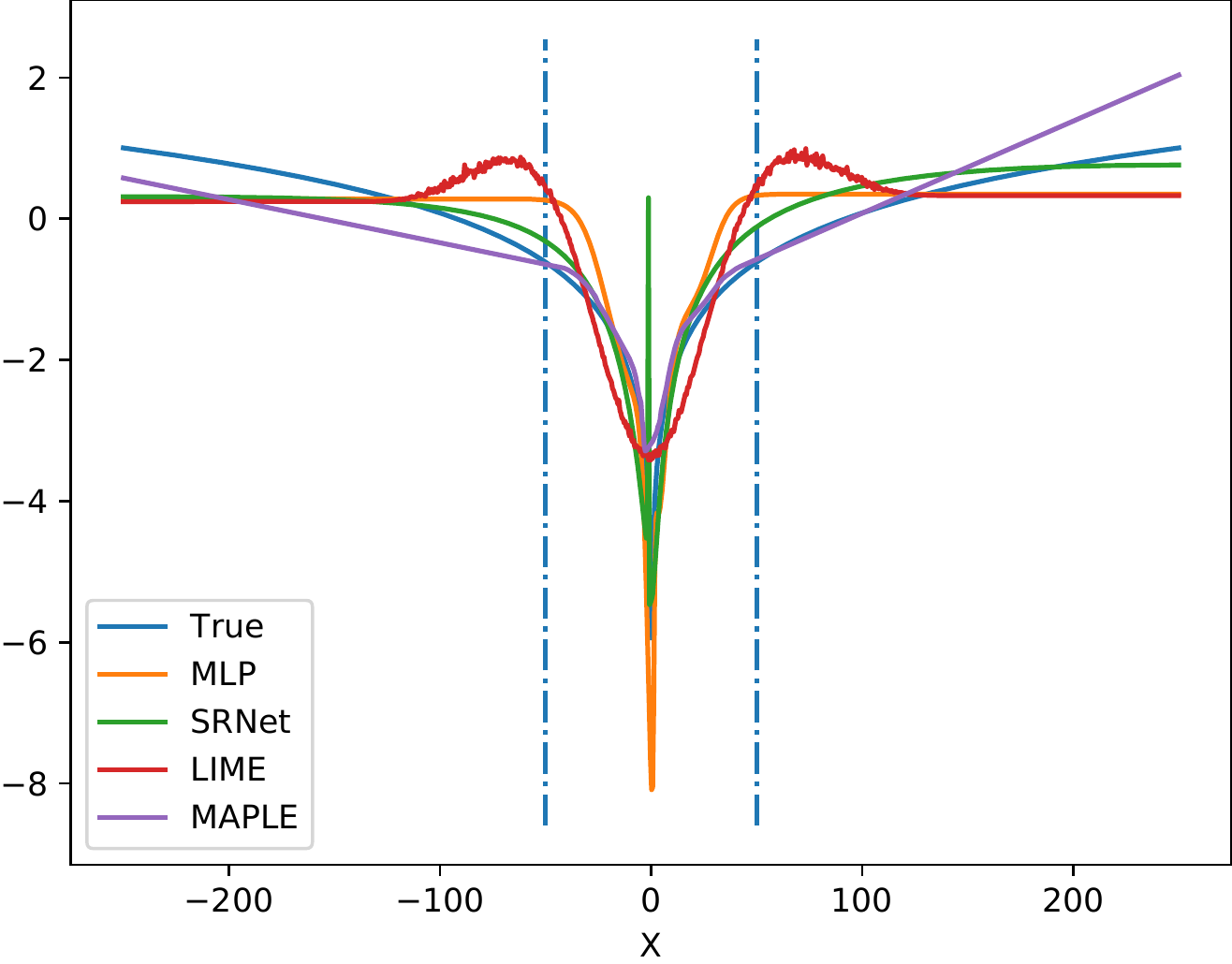}
		\caption{K2}
	\end{subfigure}
\\
	\begin{subfigure}{0.32\linewidth}
		\includegraphics[width=\linewidth]{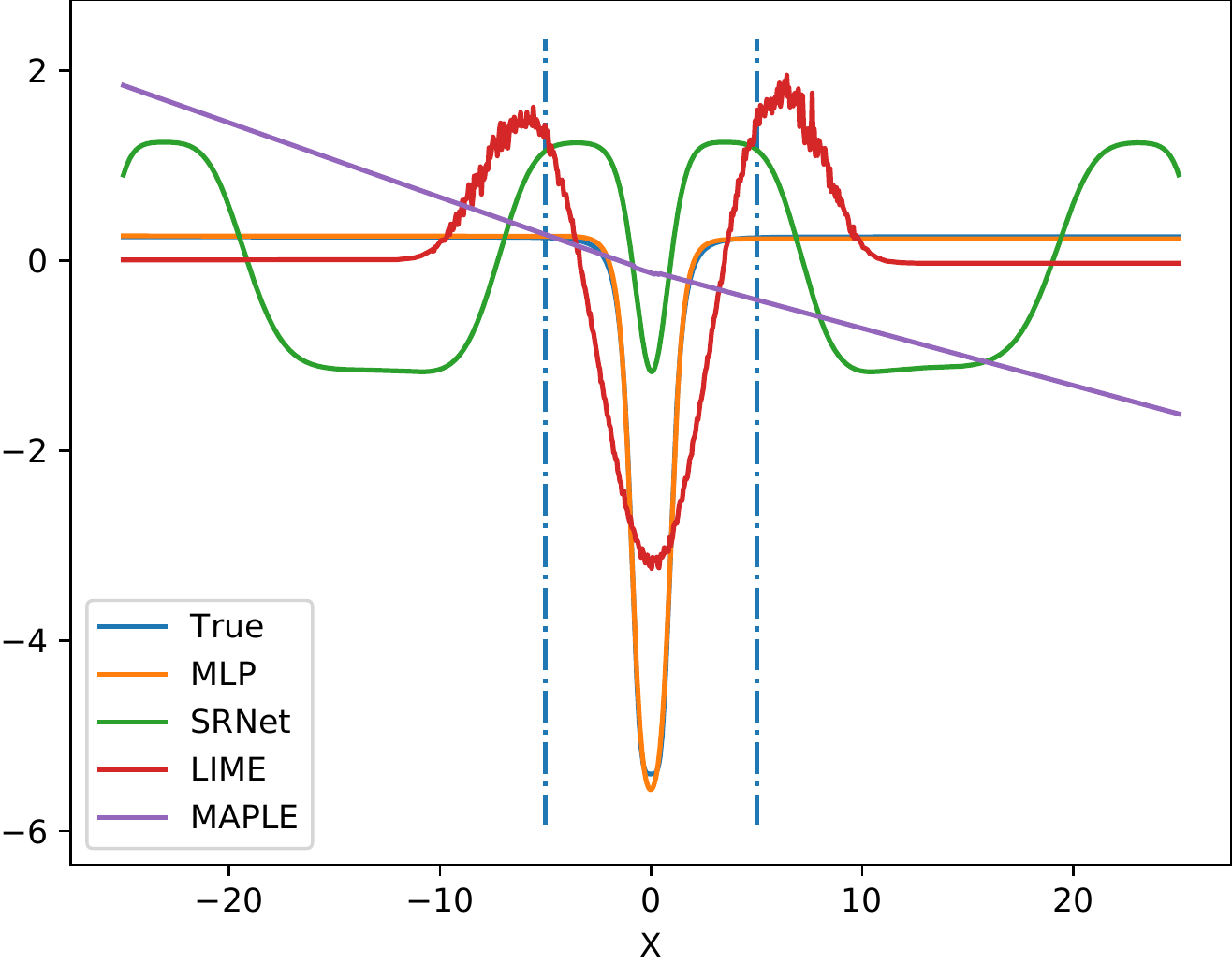}
		\caption{K3}
	\end{subfigure}
% 	\begin{subfigure}{0.33\textwidth}
% 		\includegraphics[width=\textwidth]{KKK4.pdf}
% 		\caption{K4}
% 	\end{subfigure}
% 	\begin{subfigure}{0.4\linewidth}
% 		\includegraphics[width=\textwidth]{KKK5.pdf}
% 		\caption{K5}
% 	\end{subfigure}
	\begin{subfigure}{0.32\linewidth}
		\includegraphics[width=\linewidth]{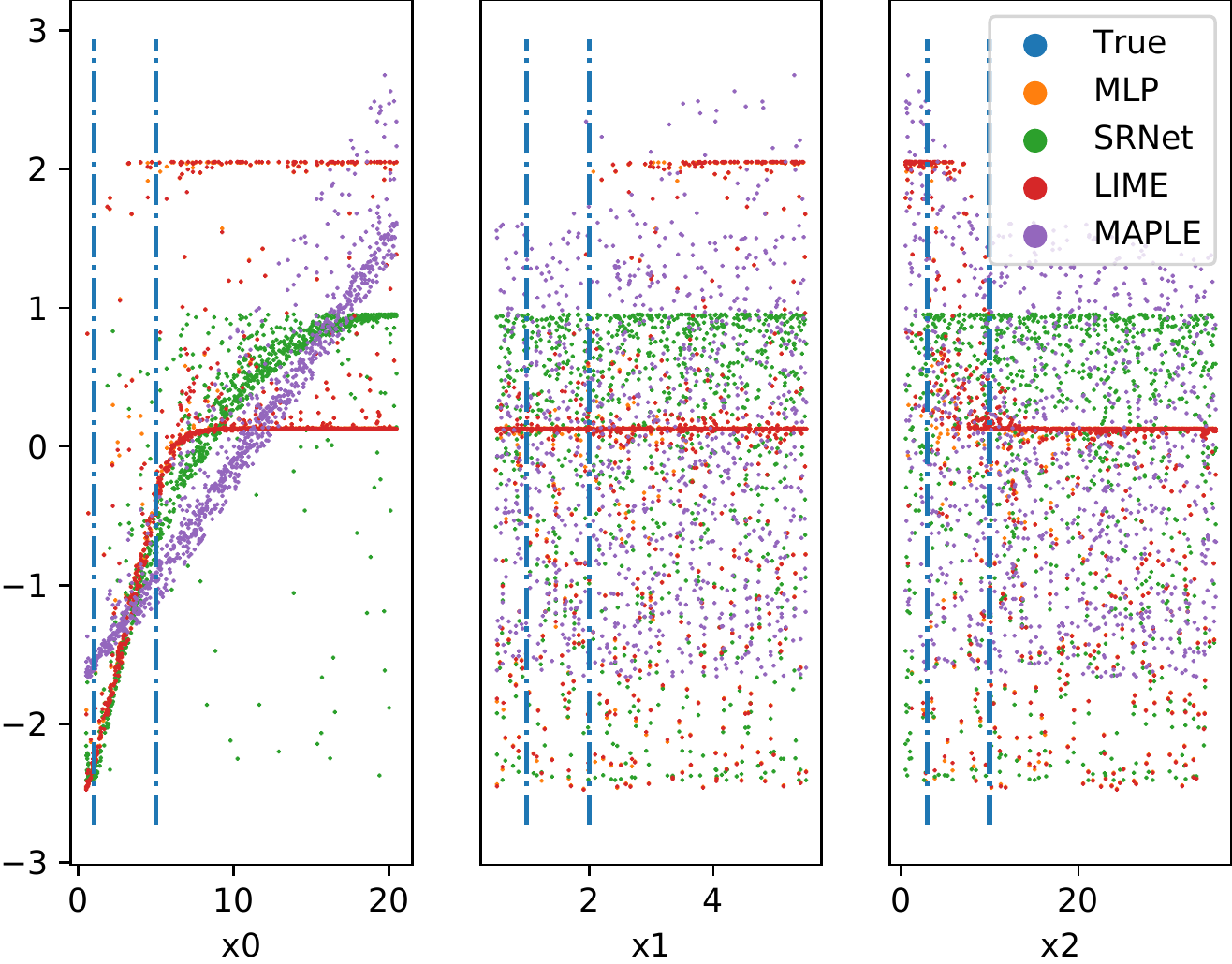}
		\caption{F0}
	\end{subfigure}
	\begin{subfigure}{0.32\linewidth}
		\includegraphics[width=\linewidth]{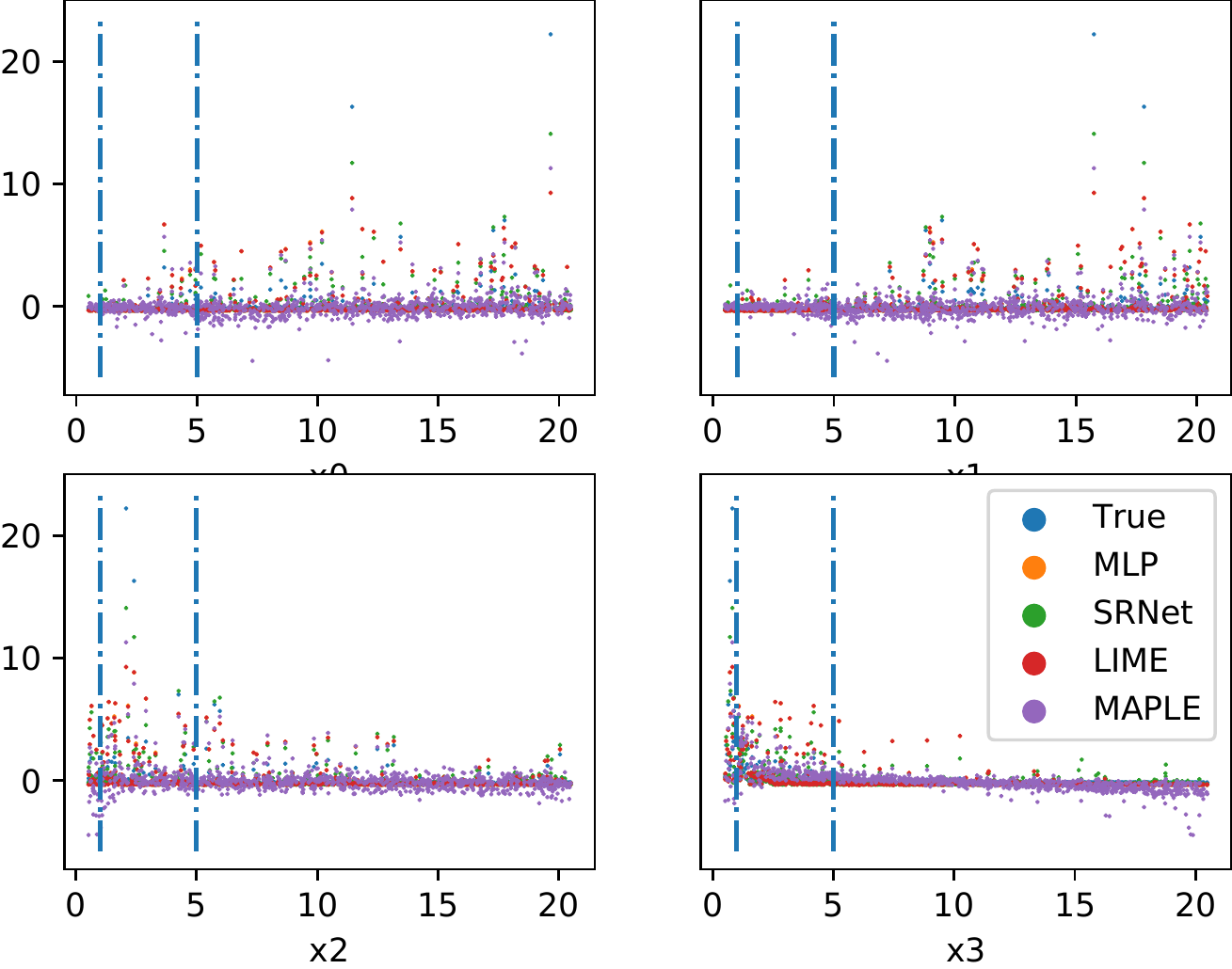}
		\caption{F1}
	\end{subfigure}
\\
% 	\begin{subfigure}{0.24\textwidth}
% 		\includegraphics[width=\textwidth]{feynman2.pdf}
% 		\caption{F2}
% 	\end{subfigure}
% 	\begin{subfigure}{0.24\textwidth}
% 		\includegraphics[width=\textwidth]{feynman3.pdf}
% 		\caption{F3}
% 	\end{subfigure}
% 	\begin{subfigure}{0.24\textwidth}
% 		\includegraphics[width=\textwidth]{feynman4.pdf}
% 		\caption{F4}
% 	\end{subfigure}
% 	\begin{subfigure}{0.24\textwidth}
% 		\includegraphics[width=\textwidth]{feynman5.pdf}
% 		\caption{F5}
% 	\end{subfigure}
	\caption{SRNet vs LIME vs MAPLE on the interpolation and extrapolation domain. The area between two blue vertical lines is the interpolation domain. The other area is the extrapolation domain}
	\label{fig:extrapolation}
\end{figure}

Figure \ref{fig:extrapolation} illustrates the curves (or distribution points) of the true dataset, as well as the results of NN(MLP)s, SRNet, LIME, and MAPLE, on different symbolic regression benchmarks. For the high dimension datasets, it is not easy to visualize the curves. So, the curves are projected into samples on multiple planes. The curves between two vertical blue lines represent interpolated results, while those outside the two lines are the extrapolated results.

% \begin{figure*}[ht]
%     \centering
%     \includegraphics[width=0.9\linewidth]{extrapolation(2).pdf}
%     \caption{SRNet vs LIME vs MAPLE on the interpolation and extrapolation domain. The area between two blue vertical lines is the interpolation domain. The other area is the extrapolation domain}
%     \label{fig:extrapolation}
% \end{figure*}

% As the output of a MLP is a smooth curve, LIME based on a local interpolation method only approximates MLP with local granularity cannot represent MLP with global granularity. While SAMPLE can obtain smooth curve, the curve does not approximate the MLP output.  

% especially on the $K2$ and $K3$ benchmarks. 
% Nevertheless, on the $K5$ benchmark, SRNet has two discontinuity points, leading to its poor results. The two points are caused by $tan$ in $K5$ mathematical expression in Table \ref{tab:inter_expressions}. If some additional methods let MNNCPS-ES avoid selecting the $tan$ function to fit the dataset continuity, SRNet may find more accurate mathematical expressions.
The interpolated results show that SRNet can find the mathematical expressions close to MLP on most of these benchmarks. SRNet can find smoother results than LIME, while it can  
find results closer to MLP compared with MAPLE. 
The extrapolated results show that LIME can find the model closest to MLP because LIME is the local explanation method that generates a model for local (several) samples. For a extrapolate dataset, LIME divides the datasets into many groups of local samples and generates a model for each group of samples. Therefore, it needs many local models to explain the extrapolated dataset and cannot provide a general model to describe the whole dataset. Unlike LIME, SRNet finds the mathematical expression that represents the whole dataset. In addition, it can trend towards the true dataset generated by the symbolic regression benchmark. Therefore, SRNet has better extrapolation results in our evaluation \cite{castillo2013symbolic}. The ability of SRNet extrapolation could help SRNet judge why a NN fails to predict some test data. Under the assumption that the mathematical expression found by SRNet can represent the real model on the practical dataset. For example, on the benchmark $K1$, if a NN is unable to predict certain output, the prediction result could be compared with the output of $- 0.01 x_{1} + 1.69 \sin{\left(x_{0} \right)} + 0.0021$ found by SRNet. The difference between the NN output and SRNet can be used to analyze the gap between the NN and the real model.   

\subsection{Classification Task}

In the following classification tasks we only show the results on the two benchmarks, $P0$, and $P1$. The classification results on all classification benchmarks are included in the appendix.

\subsubsection{Fitness Convergence}
\begin{figure}[ht]
    \centering
    \includegraphics[width=0.8\linewidth]{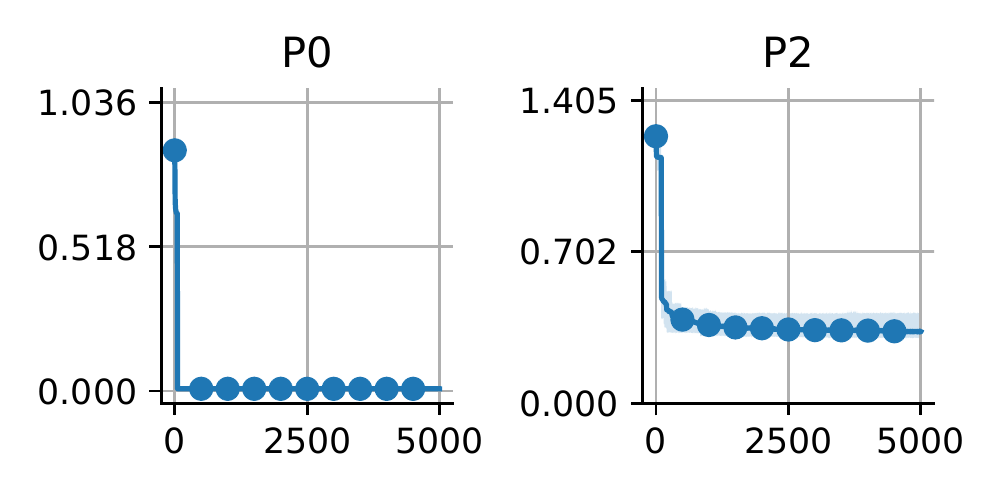}
    \caption{MNNCGP-ES convergence curve. 
    % $P0$, and $P1$ .
    }
    \label{fig:clas_convergence}
\end{figure}

Figure \ref{fig:clas_convergence} shows the fitness convergences curves of the SRNet for the classification tasks on the two benchmarks, $P0$ and $ P2$. SRNet converges rapidly, especially within about 100th generations. As MNNCGP-ES runs L-BFGS to obtain weight vectors ($w_i^s$ and $b_i$) of all individuals ($w_i^sf_i(h_{i-1}^s)+b_i$) every 50 generations, MNNCGP-ES only runs L-BFGS two times, and it can converge to an accurate result. The reason is that SRNet does not need a whole dataset to be trained, but a thousand samples around the decision boundary of a NN. For example, on the benchmark $P_0$, although it has $48842$ points, these points are only used to train a classification NN. After training the NN, USDB ( Section \ref{sec:USDB}) randomly samples 1000 points around the classification NN. Then, the 1000 points are used to train SRNet. So, the process of training SRNet in the classification task is fast.   

\subsubsection{Semantics Evaluation}
Table \ref{tab:clas_inter_expressions} lists a mathematical expression of each NN layer obtained by SRNet for the two classification tasks, $P0$ (adult) and $P2$ (agaricus\_lepiota). 
Moreover, Table \ref{tab:clas_overall_expressions} shows the final mathematical expressions obtained by SRNet. For the adult dataset, the two mathematical expressions, $Pr0$ and $Pr1$, represent the prediction probability of SRNet for class 0 (adult makes over \$50K a year) and class 1 (adult makes below \$50k a year), respectively. Moreover, $Pr0$ and $Pr1$ indicate that the classification results only depends on the three features, $x_6$ (relationship), $x_7$ (race), and $x_9$ (capital-gain). In addition, they provide a mathematical explanation of how the trained NN classifies data. 
For the agaricus\_lepiota dataset, SRNet gives a simple linear model as the explanation of the classification NN, as shown in $P2$ in Table \ref{tab:clas_overall_expressions}. According to $Pr0$ and $Pr1$, the classification NN mainly focused on the four features,  $x_{10}$ (stalk-root), $x_{20}$ (population), $x_8$ (gill-color) and $x_9$ (stalk-shape). In this case, SRNet degenerates into a simple linear model. So, when SRNet explains a NN on the classification task, it not only shows how the NN computes on the dataset, but also represents which features (variables) the NN focuses on.    
%Similarly for breast dataset, its prediction of class probability is the square root of the weighted sum of features $x_2$ (uniformity of cell size), $x_4$ (marginal adhesion), $x_6$ (bare nuclei) and $x_9$ (mitoses) and then plus a bias value.  
 
\begin{table}[htpb]
    \centering
    \scalebox{0.9}{
    \begin{tabular}{c|c|c|c|c}
        \hline
            \textbf{Dataset} & $\boldsymbol{h_0^s(x)}$ & $\boldsymbol{h_1^s(h_0^s)}$ & $\boldsymbol{h_2^s(h_1^s)}$ & $\boldsymbol{y^s}$  \\
        \hline
        $P0$ & \makecell[c]{$-x_6+x_7+x_9$\\\textbf{(1.58e-02)}} & \makecell[c]{$(h^s_0)_{85}$\\\textbf{(3.09e-04)}} & - & \makecell[c]{$\frac{(h^s_1)_{41}}{(h^s_1)_{31}}$\\\textbf{(0.0)}}  \\
        
        % $P1$ & \makecell[c]{$x_3$\\\textbf{(3.76e-03)}} & \makecell[c]{$(h^s_0)_{14}-(h^s_0)_{67}$\\\textbf{(1.10e-04)}} & \makecell[c]{$(h^s_1)_2$\\\textbf{(3.53e-05)}} & \makecell[c]{$\sin{\sqrt{(h^s_2)_0}+(h^s_2)_{19}}$\\\textbf{(0.0)}}  \\
        
        $P2$ & \makecell[c]{$-x_{10}-x_{20}+x_8+x_9$\\\textbf{(2.08e-02)}} & \makecell[c]{$(h^s_0)_{63}$\\\textbf{(5.58e-03)}} & - & \makecell[c]{$(h^s_1)_{33}$\\\textbf{(2.91e-01)}}  \\
        
        % $P3$ & \makecell[c]{$x_2+x_4+x_6+x_9$\\\textbf{(1.40e-02)}} & \makecell[c]{$(h^s_0)_{17}$\\\textbf{(2.76e-04)}} & - & \makecell[c]{$\sqrt{(h^s_1)_{19}}$\\\textbf{(0.0)}}  \\
        
        % $P4$ & \makecell[c]{$x_0-x_1+x_3+x_4$\\\textbf{(1.09e-02)}} & \makecell[c]{$(h^s_0)_{72}$\\\textbf{(2.73e-04)}} & \makecell[c]{$(h^s_1)_{34}$\\\textbf{(7.14e-06)}} & \makecell[c]{$(h^s_2)_{40}$\\\textbf{(0.0)}}\\
        
        \hline 
    \end{tabular}  
    }
    \caption{ The mathematical expression of each layer in NNs. 
    % for the two classification tasks, $P0$ and $P2$. 
    % $y^s$ is the output layer. The number below each $f_i$ is its fitness.
    }
    \label{tab:clas_inter_expressions}
\end{table}

\begin{table}[t]
    \centering
    \scalebox{0.8}{
    \begin{tabular}{c|c}
        \hline
            \textbf{Dataset} & $\boldsymbol{O^s(x)}$\\
        \hline
        $P0$ & \makecell[c]{$Pr0 = -14.16 - \frac{0.00043x_6 - 0.00043x_7 - 0.00043x_9 + 0.42}{-0.0025x_6 + 0.0025x_7 + 0.0025x_9 + 0.489}$\\$Pr1 = 14.57 + \frac{0.00047x_6 - 0.00047x_7 - 0.00047x_9 + 0.46}{-0.0025x_6 + 0.0025x_7 + 0.0025x_9 + 0.49}$\\\textbf{(8.05e-03)}}\\
        % $P1$ & \makecell[c]{$Pr0 = -5.10sin(0.00033x_3 + 0.67\sqrt{1 - 0.00044x_3} + 0.45) - 6.51$\\$Pr1 = 5.42sin(0.00033x_3 + 0.67\sqrt{1 - 0.00044x_3} + 0.45) + 6.97$\\\textbf{(1.30e-03)}}
        $P2$ & \makecell[c]{$Pr0 = 0.66x_{10} + 0.66x_{20} - 0.66x_8 - 0.66x_9 - 1.57$\\$Pr1 = -0.53x_{10} - 0.53x_{20} + 0.53x_8 + 0.53x_9 + 1.17$\\\textbf{(3.04e-01)}} \\
        % $P3$ & \makecell[c]{$Pr0 = \sqrt{-0.15x_2 - 0.15x_4 - 0.15x_6 - 0.15x_9 + 19.27} + 14.87 $ \\ $Pr1 = \sqrt{-0.18x_2 - 0.18x_4 - 0.18x_6 - 0.18x_9 + 22.28} - 15.34$ \\ \textbf{(7.15e-03)}}\\
        \hline 
    \end{tabular}
    }
    \caption{The mathematical expressions of the whole NN.}
    \label{tab:clas_overall_expressions}
\end{table}

\subsubsection{performance comparison}  
Table \ref{tab:cls_accuracy} lists their average (+ std) prediction accuracy on the train and test dataset. Interestingly, SRNet is better than LIME and MAPLE on most classification tasks and only fails on the 'agaricus\_lepiota' ($P2$) task. The fail reason is that the dataset 'agaricus\_lepiota' has 22 input features, making 1000 points sampled by USDB very sparse in the high-dimensional space. However, SRNet still shows better or competitive accuracy in the other four classification tasks than LIME and MAPLE.  

\begin{table}[t]
    \centering
    \scalebox{0.8}{
    \begin{tabular}{c|c|c|c}
        \hline
            \textbf{Dataset} & \textbf{Method} & \textbf{Train} & \textbf{Test}\\
        \hline
        $P0$(adult) & \makecell[c]{LIME\\MAPLE\\SRNet} & \makecell[c]{$84.47\pm2.3\%$ \\ $98.47\pm1.51\%$ \\ \boldmath $100\pm 0\%$} & \makecell[c]{$71.33\pm1.78\%$ \\ $92.11\pm0.80\%$ \\ \boldmath $100\pm 0\%$}\\
        \hline
        $P1$(analcatdata\_aids) & \makecell[c]{LIME\\MAPLE\\SRNet} & \makecell[c]{$85.18\pm 0.11\%$ \\ \boldmath$93.78\pm0.23\%$ \\ $91.89\pm8.11\%$} & \makecell[c]{$90.89\pm 0.62\%$ \\ \boldmath$100\pm0\%$ \\ $100\pm0\%$}\\
        \hline
        $P2$(agaricus\_lepiota) & \makecell[c]{LIME\\MAPLE\\SRNet} & \makecell[c]{$88.28\pm0.16\%$ \\ \boldmath $99.22\pm0.04\%$ \\ $75.82\pm0\%$} & \makecell[c]{$93.81\pm0.30\%$ \\ \boldmath $98.16\pm0.12\%$ \\ $75.60\pm0\%$} \\
        \hline
        $P3$(breast) & \makecell[c]{LIME\\MAPLE\\SRNet} & \makecell[c]{$100\pm0\%$ \\ $100\pm0\%$ \\ $100\pm0\%$} & \makecell[c]{$100\pm0\%$ \\ $100\pm0\%$ \\ $100\pm0\%$}\\
        \hline
        $P4$(car) & \makecell[c]{LIME\\MAPLE\\SRNet} & \makecell[c]{$100\pm0\%$ \\ $100\pm0\%$ \\ $100\pm0\%$} & \makecell[c]{$100\pm0\%$ \\ $100\pm0\%$ \\ $100\pm0\%$}\\
        \hline
    \end{tabular}   
    }
    \caption{LIME vs MAPLE vs SRNet.}
    \label{tab:cls_accuracy}
\end{table}

\begin{figure}
    \centering
    \includegraphics[width=0.8\linewidth]{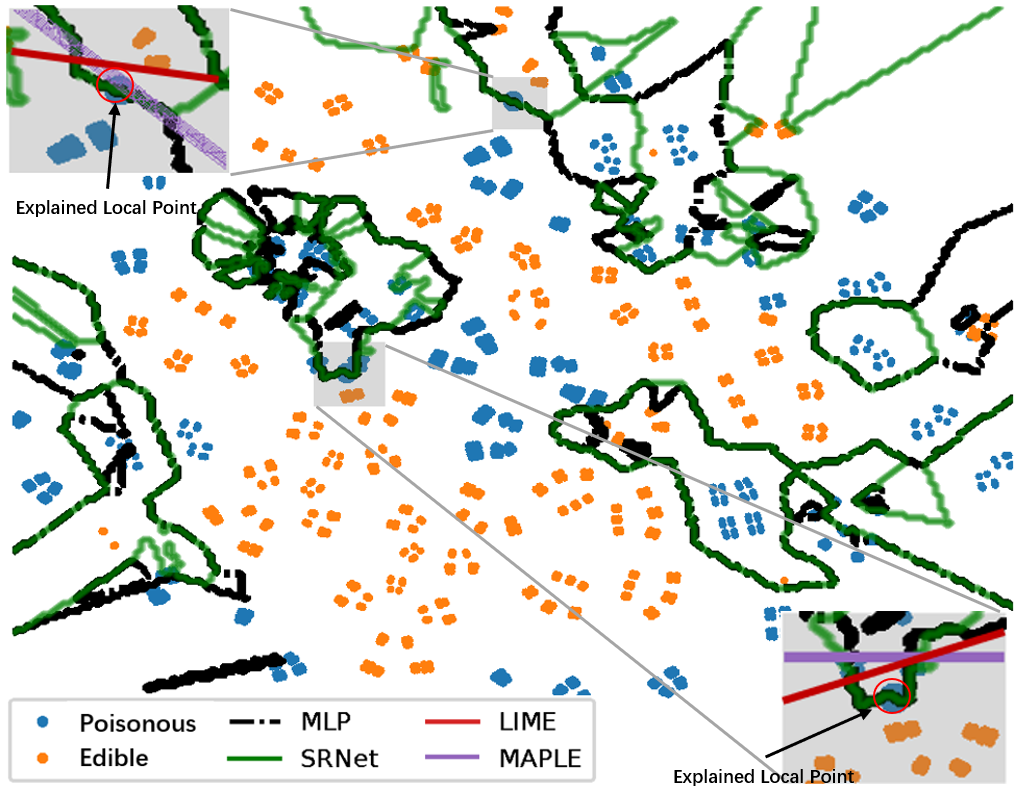}
    \caption{LIME vs MAPLE vs SRNet on $P2$ decision boundary.}
    \label{fig:local_db}
\end{figure}

Although LIME and MAPLE have good performance in explaining local classification samples, their decision boundaries cannot approximate the MLP's decision boundary. The reason is that they only leverage a linear model to explain several local samples, as shown in red lines and purple lines in Figure \ref{fig:local_db}. So, once the MLP's decision boundary is a complex curve at some local samples, the line generated by LIME or MAPLE cannot approximate it. In contrast, SRNet can approximate the complex decision boundary of MLP on all samples. Therefore, SRNet is more suitable for explaining NN on the classification task than LIME and MAPLE.

\section{Conclusion}

This paper proposes a new evolutionary algorithm called SRNet to address the NN's black box problem. SRNet leverages MNNCGP-ES to find the mathematical expressions that can represent each NN layer's hidden semantics. The combination of every layer's expression represents the whole NN. Compared with the models found by LIME and MAPLE, the mathematical expression provided by SRNet is closer to the NNs in the interpolated domain. The experiment also shows the SRNet models trend to approximate the real data model that used trains NN. The close alignment of SRNet with the real model and its explicit mathematical expression can be used to facilitate the explanation of NN prediction behaviours, such as regression and classification. 
% Although SRNet demonstrated great potential to explain NN, it needs future work to overcome the two challenges: 1) expression complexity with more layers in NNs, and 2) invalidation to high dimensions.   

\bibliographystyle{ACM-Reference-Format}
\bibliography{references}
\end{document}

\title{Supplementary material}
\appendix
\section{Appendix}
\subsection{Convergence Curve of Fitness}
\label{appendix_sec:convergence_fitness}
% Figure \ref{append_fig:regression_convergence} illustrates the convergence curve of fitness of MNNCGP-ES that run 30 times on SRNet regression datasets $K0-K5, F0-F5$. 
\begin{figure}[htbp]
    \centering
    \scalebox{0.6}{
        \includegraphics[width=\linewidth]{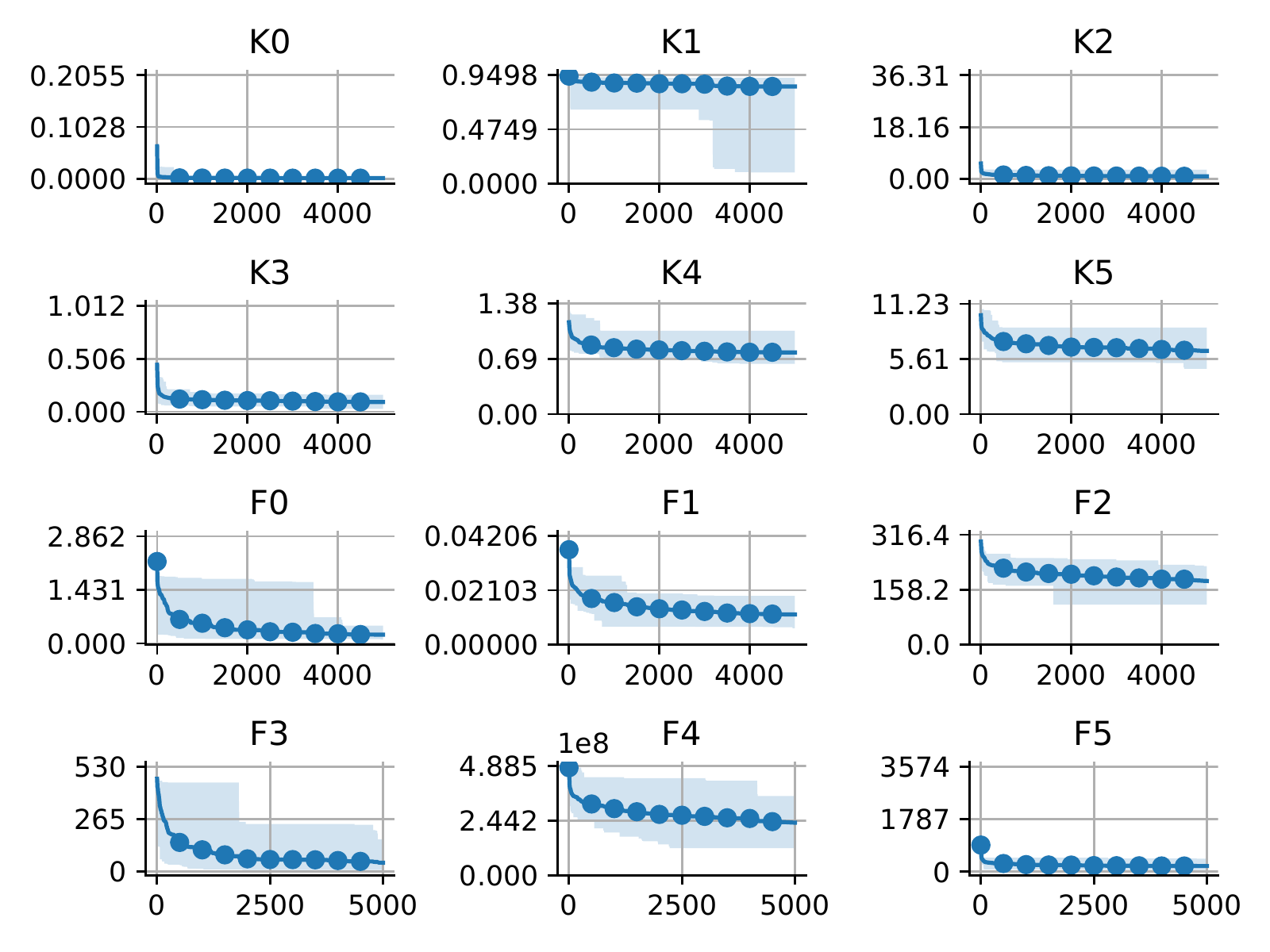}
    }
    \caption{The convergence curve of fitness of MNNCGP-ES that run 30 times on each MLP of $K0-K5, F0-F5$. The blue line represents the average fitness. The light blue range is the fitness range (maximum and minimum values) of the 30 results.}
    \label{append_fig:regression_convergence}
\end{figure}

% Figure \ref{append_fig:clas_convergence} illustrates the convergence curve of fitness of MNNCGP-ES that run 30 times on SRNet classification datasets $P0-P5$. 
\begin{figure}[ht]
    \centering
    \scalebox{0.6}{
    \includegraphics[width=0.8\linewidth]{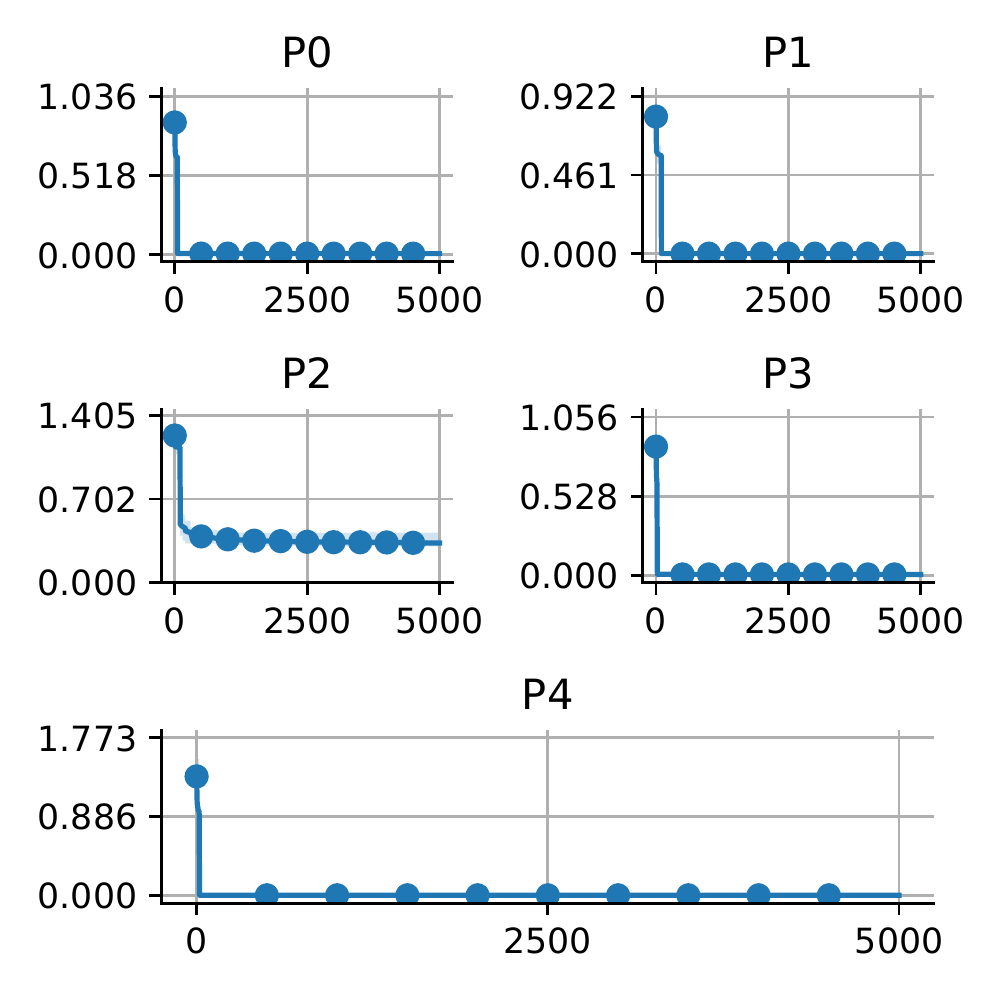}
    }
    \caption{The convergence curve of fitness of MNNCGP-ES that run 30 times on each MLP for the classification tasks $P0-P5$.}
    \label{append_fig:clas_convergence}
\end{figure}

\newpage
\subsection{Mathematical Expressions of Neural Networks}
\label{appendix_sec:expressions}
% Table \ref{append_tab:inter_expressions} shows all the best fitted mathematical expression ($f_i$) for explaining the hidden semantics of each layer $h_i^s$ in NN. 

\begin{table*}[h]
    \centering
    \begin{tabular}{c|c|c|c|c}
        \hline
            \textbf{Dataset} & $\boldsymbol{h_0^s(x)}$ & $\boldsymbol{h_1^s(h_0^s)}$ & $\boldsymbol{h_2^s(h_1^s)}$ & $\boldsymbol{y^s}$  \\
        \hline
        $K0$ & \makecell[c]{$sin(0.26x +sin(sin(x))-0.068)$\\\textbf{(6.74e-04)}} & \makecell[c]{$0.88 - \cos{(h^s_0)_0}$\\\textbf{(7.54e-05)}} & $-$ & \makecell[c]{$-sin(sin((h^s_0)_0-0.36))$\\\textbf{(2.37e-04)}} \\
             
        $K1$ & \makecell[c]{5.15e-05$x_1-0.0072sin(x_0) -$5.15e-05\\\textbf{(5.34e-02)}} & \makecell[c]{$-(h^s_0)_0 + (h^s_0)_2$\\\textbf{(1.39e-03)}} & $-$ & \makecell[c]{$-(h^s_1)_0 + (h^s_1)_2 + 0.0011$\\\textbf{(6.88e-02)}} \\
        
        $K2$ & \makecell[c]{$\frac{\sqrt{x^{2}} \log{\left(x^{2} \right)}}{x + 1.20}$\\\textbf{(3.26e-02)}} & \makecell[c]{$- tan(((h^s_0)_{2} +(h^s_0)_{3}^{2} -(h^s_0)_{3}))$\\\textbf{(1.04e-02)}} & $-$ & \makecell[c]{$\sin{\left(\frac{(h^s_1)_{0}}{(h^s_1)_{2}^{2}} \right)}$\\\textbf{(2.00e-01)}} \\
        
        $K3$ & \makecell[c]{$- sin((0.24 x_{0} + \sin{\left(0.23 x_{1} \right)}))$\\\textbf{(3.86e-02)}} & \makecell[c]{$(h^s_0)_{1} (h^s_0)_{2}^{2} \sin{\left((h^s_0)_{1} \right)}$\\\textbf{(3.64e-03)}} & \makecell[c]{$(h^s_1)_{0} (- (h^s_1)_{2} + (h^s_1)_{3})$\\\textbf{(2.87e-04)}} & \makecell[c]{$-0.99 + \frac{(h^s_2)_{3}}{(h^s_2)_{0}}$\\\textbf{(1.32e-02)}} \\
        
        $K4$ & \makecell[c]{$- \sin{(x_{1})} + $\\ $\sin{(\frac{\sin{(x_{0})}}{x_{2}})}$\\\textbf{(1.17e-02)}} 
        &\makecell[c]{$((h^s_0)_{0} + (h^s_0)_{3} - \tan{\left((h^s_0)_{1} \right)})^{2}$\\\textbf{(1.31e-02)}} & $-$ & \makecell[c]{$- \frac{\sin{\left((h^s_1)_{1} - (h^s_1)_{2} \right)}}{(h^s_1)_{0}^{2}}$\\\textbf{(6.14e-01)}} \\
        
        $K5$ & \makecell[c]{$x_{1} - 0.097$\\\textbf{(9.01e-02)}} & \makecell[c]{$\frac{\sin{\left((h^s_0)_{3} - 0.48 \right)}}{(h^s_0)_{2}}$\\\textbf{(2.44e-02)}} & $-$ & \makecell[c]{$\frac{\tan{\left((h^s_1)_{0} + 0.51 \right)}}{(h^s_1)_{2} - 0.26}$\\\textbf{(4.56)}} \\

        $F0$ & \makecell[c]{$cos((\frac{0.67 x_{1}}{x_{2}} + \log{(x_{0})}))$\\\textbf{(6.35e-03)}} & \makecell[c]{$- (h^s_0)_{1} + \cos{\left((h^s_0)_{2} \right)}$\\\textbf{(1.04e-02)}} & $-$ & \makecell[c]{$tan((tan(((h^s_1)_{0} - 0.33)) + 0.11 ))$\\\textbf{(9.63e-02)}} \\ 
             
        $F1$ & \makecell[c]{$\log{\left(\frac{x_{0} x_{1}}{x_{2} x_{3}^{\frac{3}{2}}} \right)}$\\\textbf{(9.59e-03)}} & \makecell[c]{$(h^s_0)_{1}^{4}$\\\textbf{(2.46e-04)}} & $-$ & \makecell[c]{$(h^s_1)_{0} + \sin{\left((h^s_1)_{0} \right)}$\\\textbf{(1.45e-03)}} \\
        
        $F2$ & \makecell[c]{$- \frac{x_{4}}{x_{3}} + \frac{x_{4}}{x_{2}}$\\\textbf{(1.42e-02)}} & \makecell[c]{$(h^s_0)_{2} \left((h^s_0)_{1} + 1\right)$\\\textbf{(1.36e-02)}} & $-$ & \makecell[c]{$\frac{0.082 - (h^s_1)_{2}}{(h^s_1)_{0}}$\\\textbf{(1.15e02)}} \\
        
        $F3$ & \makecell[c]{$sin((\sqrt[4]{2} \sqrt[4]{x_{0}} + \sqrt{x_{1}}))$\\\textbf{(1.22e-02)}} & \makecell[c]{$\tan{\left((h^s_0)_{1} \right)} - 0.30$\\\textbf{(1.64e-02)}} & $-$ & \makecell[c]{$(h^s_1)_{1} - \cos{\left(4 (h^s_1)_{2}^{2} \right)}$\\\textbf{(5.71)}} \\
        
        $F4$ & \makecell[c]{$x_{1} x_{4} - \sqrt{x_{2} + x_{3}}$\\\textbf{(7.49e-02)}} & \makecell[c]{$(h^s_0)_{4} \cos^{2}{\left((h^s_0)_{1} \right)}$\\\textbf{(3.61e-02)}} & $-$ & \makecell[c]{$(h^s_1)_{0}^{2} (h^s_1)_{4}^{2}$\\\textbf{(1.21e08)}} \\
        
        $F5$ & \makecell[c]{$\sqrt{x_{1}} + \log{\left(\frac{x_{1}}{x_{0} x_{2}} \right)}$\\\textbf{(3.84e-03)}} & \makecell[c]{$\sin^{2}{\left(0.51 (h^s_0)_{0} \right)}$\\\textbf{(1.20e-03)}} & \makecell[c]{$\frac{\left(2 (h^s_1)_{1} - \tan{\left((h^s_1)_{1} \right)}\right)^{4}}{(h^s_1)_{3}^{2}}$\\\textbf{(5.12e-03)}} & \makecell[c]{$\tan{\left(\tan{\left((h^s_2)_{0} \right)} \right)}$\\\textbf{(5.75e01)}} \\
        
        $P0$ & \makecell[c]{$-x_6+x_7+x_9$\\\textbf{(1.58e-02)}} & \makecell[c]{$(h^s_0)_{85}$\\\textbf{(3.09e-04)}} & - & \makecell[c]{$\frac{(h^s_1)_{41}}{(h^s_1)_{31}}$\\\textbf{(0.0)}}  \\
        
        $P1$ & \makecell[c]{$x_3$\\\textbf{(3.76e-03)}} & \makecell[c]{$(h^s_0)_{14}-(h^s_0)_{67}$\\\textbf{(1.10e-04)}} & \makecell[c]{$(h^s_1)_2$\\\textbf{(3.53e-05)}} & \makecell[c]{$\sin{\sqrt{(h^s_2)_0}+(h^s_2)_{19}}$\\\textbf{(0.0)}}  \\
        
        $P2$ & \makecell[c]{$-x_{10}-x_{20}+x_8+x_9$\\\textbf{(2.08e-02)}} & \makecell[c]{$(h^s_0)_{63}$\\\textbf{(5.58e-03)}} & - & \makecell[c]{$(h^s_1)_{33}$\\\textbf{(2.91e-01)}}  \\
        
        $P3$ & \makecell[c]{$x_2+x_4+x_6+x_9$\\\textbf{(1.40e-02)}} & \makecell[c]{$(h^s_0)_{17}$\\\textbf{(2.76e-04)}} & - & \makecell[c]{$\sqrt{(h^s_1)_{19}}$\\\textbf{(0.0)}}  \\
        
        $P4$ & \makecell[c]{$x_0-x_1+x_3+x_4$\\\textbf{(1.09e-02)}} & \makecell[c]{$(h^s_0)_{72}$\\\textbf{(2.73e-04)}} & \makecell[c]{$(h^s_1)_{34}$\\\textbf{(7.14e-06)}} & \makecell[c]{$(h^s_2)_{40}$\\\textbf{(0.0)}}\\
        
        \hline 
    \end{tabular}
    \caption{ The best fitted mathematical expression ($f_i$) for explaining the hidden semantics of each layer $h_i^s$ in NN. $y^s$ is the output layer in NN. The number below each $f_i$ is its fitness.}
    \label{append_tab:inter_expressions}
\end{table*}

\newpage
% All the mathematical expressions of explaining whole NN are shown in Table \ref{append_tab:overall_expressions}.

\begin{table*}[h]
    \centering
    \scalebox{0.8}{
    \begin{tabular}{c|c}
        \hline
            \textbf{Dataset} & $\boldsymbol{O^s(x)}$\\
        \hline
        $K0$ & \makecell[c]{$0.29 - 4.01 sin((sin((2.36 cos((0.41 sin((0.26 x +\sin{\left(\sin{\left(x \right)} \right)} - 0.068)) + 0.49)) - 2.03))))$\\\textbf{(6.11e-04)}}\\
        
        $K1$ & \makecell[c]{$- 0.01 x_{1} + 1.69 \sin{\left(x_{0} \right)} + 0.0021$\\\textbf{(9.62e-02)}}\\
        
        $K2$ & \makecell[c]{$4.49 \sin{\left(\frac{6.70 \left(0.51 \tan{\left(0.061 \left(1 - \frac{0.19 \sqrt{x^{2}} \log{\left(x^{2} \right)}}{x + 1.20}\right)^{2} - 0.044 + \frac{0.084 \sqrt{x^{2}} \log{\left(x^{2} \right)}}{x + 1.20} \right)} + 0.27\right)}{\left(\tan{\left(0.061 \left(1 - \frac{0.19 \sqrt{x^{2}} \log{\left(x^{2} \right)}}{x + 1.20}\right)^{2} - 0.044 + \frac{0.084 \sqrt{x^{2}} \log{\left(x^{2} \right)}}{x + 1.20} \right)} + 0.63\right)^{2}} \right)}$\\$ + 7.90$\\\textbf{(2.22e-01)}}\\
        
         $K3$ & \makecell[c]{$17.08 (- 0.81 ((- 1.28 (0.57 - 0.55 \sin{(0.24 x_{0} + \sin{(0.23 x_{1} )} )})(0.93 \sin{(0.24 x_{0} + \sin{(0.23 x_{1} )} )} + 1)^{2} \sin{(0.55 \sin{(0.24 x_{0} + \sin{(0.23 x_{1} )} )} - 0.57 )} + 0.39))$\\$((- 0.88 (0.57 - 0.55 \sin{(0.24 x_{0} + \sin{(0.23 x_{1} )} )})(0.93 \sin{(0.24 x_{0} + \sin{(0.23 x_{1} )} )} + 1)^{2}\sin{(0.55 \sin{(0.24 x_{0} + \sin{(0.23 x_{1} )}))} - 0.57 )} - 0.12) + 0.43)$\\$\div (- 0.51 (- 1.28 (0.57 - 0.55 \sin{(0.24 x_{0} + \sin{(0.23 x_{1} )} )})(0.93 \sin{(0.24 x_{0} + \sin{(0.23 x_{1} )} )} + 1)^{2} \sin(0.55 \sin{(0.24x_{0} + \sin{(0.23 x_{1} )} )} $\\$- 0.57 ) + 0.39) (- 0.88 (0.57 - 0.55 \sin{(0.24 x_{0} + \sin{(0.23 x_{1} )} )}) (0.93 \sin{(0.24 x_{0} + \sin{(0.23 x_{1} )} )} + 1)^{2} \sin{(0.55 \sin{(0.24 x_{0} + \sin{(0.23 x_{1} )} )} - 0.57 )} - 0.12) + 0.44) - 15.36$\\\textbf{(2.73e-02)} }\\

         $K4$ & \makecell[c]{$-0.0080 + 1.78 \sin((2.43 (0.19 \sin{\left(x_{1} \right)} - 0.19 \sin{\left(\frac{\sin{\left(x_{0} \right)}}{x_{2}} \right)} - \tan{\left(0.020 \sin{\left(x_{1} \right)} - 0.020 \sin{\left(\frac{\sin{\left(x_{0} \right)}}{x_{2}} \right)} + 0.73 \right)} + 0.90)^{2} - 0.026))$\\$\div (0.35 - (0.19 \sin{\left(x_{1} \right)} - 0.19 \sin{\left(\frac{\sin{\left(x_{0} \right)}}{x_{2}} \right)} - \tan{\left(0.20 \sin{\left(x_{1} \right)} - 0.020 \sin{\left(\frac{\sin{\left(x_{0} \right)}}{x_{2}} \right)} + 0.73 \right)} + 0.90)^{2})^{2}$\\\textbf{(6.27e-01)}}\\

        $K5$ & \makecell[c]{$-0.14 + \frac{0.011 \tan{\left(1.42 - \frac{0.051 \sin{\left(0.25 x_{1} - 0.14 \right)}}{0.18 - 0.020 x_{1}} \right)}}{0.16 - \frac{0.0414421632885933 \sin{\left(0.25 x_{1} - 0.14 \right)}}{0.18 - 0.020 x_{1}}}$\\\textbf{(4.62)}}\\
        
        $F0$ & \makecell[c]{$3.05 - 7.00 tan((tan((0.58 \cos{\left(\log{\left(m_{0} \right)} + \frac{0.67 v}{c} \right)} - 0.76 \cos{\left(0.22 \cos{\left(\log{\left(m_{0} \right)} + \frac{0.67 v}{c} \right)} - 1.01 \right)} + 0.35)) - 0.11))$\\\textbf{(1.05e-01)}} \\

        $F1$ & \makecell[c]{$0.022 \left(0.46 \log{\left(\frac{q_{1} q_{2}}{e r^{\frac{3}{2}}} \right)} + 1\right)^{4} + 3.22 \sin{\left(0.0068 \left(0.46 \log{\left(\frac{q_{1} q_{2}}{e r^{\frac{3}{2}}} \right)} + 1\right)^{4} - 0.00072 \right)} + 0.0059$\\\textbf{(6.37e-03)}}\\
        
        $F2$ & \makecell[c]{$0.66 + \frac{173.88 \left(1.74 \left(0.0077 + \frac{0.014 r_{2}}{r_{1}} - \frac{0.014 r_{2}}{m_{2}}\right) \left(0.93 - \frac{0.054 r_{2}}{r_{1}} + \frac{0.054 r_{2}}{m_{2}}\right) - 0.023 + \frac{0.025 r_{2}}{r_{1}} - \frac{0.025 r_{2}}{m_{2}}\right)}{- 0.72 \left(0.0077 + \frac{0.014 r_{2}}{r_{1}} - \frac{0.014 r_{2}}{m_{2}}\right) \left(0.93 - \frac{0.054 r_{2}}{r_{1}} + \frac{0.054 r_{2}}{m_{2}}\right) + 0.97 - \frac{0.010 r_{2}}{r_{1}} + \frac{0.010 r_{2}}{m_{2}}}$\\\textbf{(1.15e02)}}\\

        $F3$ & \makecell[c]{$- 31.43 \cos{\left(4.40 \left(- \tan{\left(0.46 \sin{\left(\sqrt[4]{2} \sqrt[4]{k} + \sqrt{x} \right)} - 0.27 \right)} - 0.091\right)^{2} \right)} - 31.98 \tan{\left(0.46 \sin{\left(\sqrt[4]{2} \sqrt[4]{k} + \sqrt{x} \right)} - 0.27 \right)} + 31.62$\\\textbf{(5.72)}}\\

        $F4$ & \makecell[c]{$- 1.15e14 (\left(- 0.0026 c r + 0.0026 \sqrt{m_{1} + m_{2}} + 0.00035\right) \cos^{2}{\left(0.37 c r - 0.37 \sqrt{m_{1} + m_{2}} + 0.58 \right)} - 0.00061)^{2} $\\$(\left(- 0.0026 c r + 0.0026 \sqrt{m_{1} + m_{2}} + 0.00035\right) \cos^{2}{\left(0.37 c r - 0.37 \sqrt{m_{1} + m_{2}} + 0.58 \right)} - 0.00050)^{2} - 93.80$\\\textbf{(1.21e08)}} \\
        
        $F5$ & \makecell[c]{$12.10 - 735.10 tan((tan(0.016 - (6.39e06 (\sin^{2}{\left(0.0072 \sqrt{y} + 0.0072 \log{\left(\frac{y}{V q} \right)} - 0.0083 \right)}-0.0010 \tan((489.69 \sin^{2}{\left(0.0072 \sqrt{y} + 0.0072 \log{\left(\frac{y}{V q} \right)} - 0.0084 \right)} $\\$+ 0.10)) + 0.00021)^{4})\div \left(\sin^{2}{\left(0.0072 \sqrt{y} + 0.0072 \log{\left(\frac{y}{V q} \right)} - 0.0084 \right)} + 0.00033\right)^{2})))$\\\textbf{(57.47)}}\\
        
        $P0$ & \makecell[c]{$Pr0 = -14.16 - \frac{0.00043x_6 - 0.00043x_7 - 0.00043x_9 + 0.42}{-0.0025x_6 + 0.0025x_7 + 0.0025x_9 + 0.489}$\\$Pr1 = 14.57 + \frac{0.00047x_6 - 0.00047x_7 - 0.00047x_9 + 0.46}{-0.0025x_6 + 0.0025x_7 + 0.0025x_9 + 0.49}$\\\textbf{(8.05e-03)}}\\
        
        $P1$ & \makecell[c]{$Pr0 = -5.10sin(0.00033x_3 + 0.67\sqrt{1 - 0.00044x_3} + 0.45) - 6.51$\\$Pr1 = 5.42\sin(0.00033x_3 + 0.67\sqrt{1 - 0.00044x_3} + 0.45) + 6.97$\\ \textbf{(8.05e-03)}}\\
        
        $P2$ & \makecell[c]{$Pr0 = 0.66x_{10} + 0.66x_{20} - 0.66x_8 - 0.66x_9 - 1.57$\\$Pr1 = -0.53x_{10} - 0.53x_{20} + 0.53x_8 + 0.53x_9 + 1.17$\\ \textbf{(3.04e-01)}}\\
        
        $P3$ & \makecell[c]{$Pr0 = \sqrt{-0.15x_2 - 0.15x_4 - 0.15x_6 - 0.15x_9 + 19.27} + 14.87 $ \\ $Pr1 = \sqrt{-0.18x_2 - 0.18x_4 - 0.18x_6 - 0.18x_9 + 22.28} - 15.34$ \\ \textbf{(7.15e-03)}}\\
        
        $P4$ & \makecell[c]{$Pr0 = -2.06e-5x_0 + 2.06e-5x_1 - 2.06e-5x_3 - 2.06e-5x_4 - 5.00$\\$Pr1 = -1.64e-5x_0 + 1.64e-5x_1 - 1.64e-5x_3 - 1.64e-5x_4 - 5.18$\\$Pr2 = 3.84e-5x_0 - 3.84e-5x_1 + 3.84e-5x_3 + 3.84e-5x_4 + 12.80$\\$Pr3 = -1.35e-5x_0 + 1.35e-5x_1 - 1.35e-5x_3 - 1.35e-5x_4 - 4.57$ \\ \textbf{(3.73e-03)} }\\
        \hline 
    \end{tabular}    
    }
    \caption{The mathematical expressions of explaining whole NN.}
    \label{append_tab:overall_expressions}
\end{table*}
\newpage
\subsection{Comparison}
\label{appendix_sec:comparison}
% For regression tasks, we compare LIME, MAPLE and SRNet on the interpolation domain and extrapolation domain. Figure \ref{append_fig:extrapolation} illustrates such comparison for $K0-K5, F0-F5$.
\begin{figure*}[ht]
	\centering
	\begin{subfigure}{0.24\textwidth}
		\includegraphics[width=\textwidth]{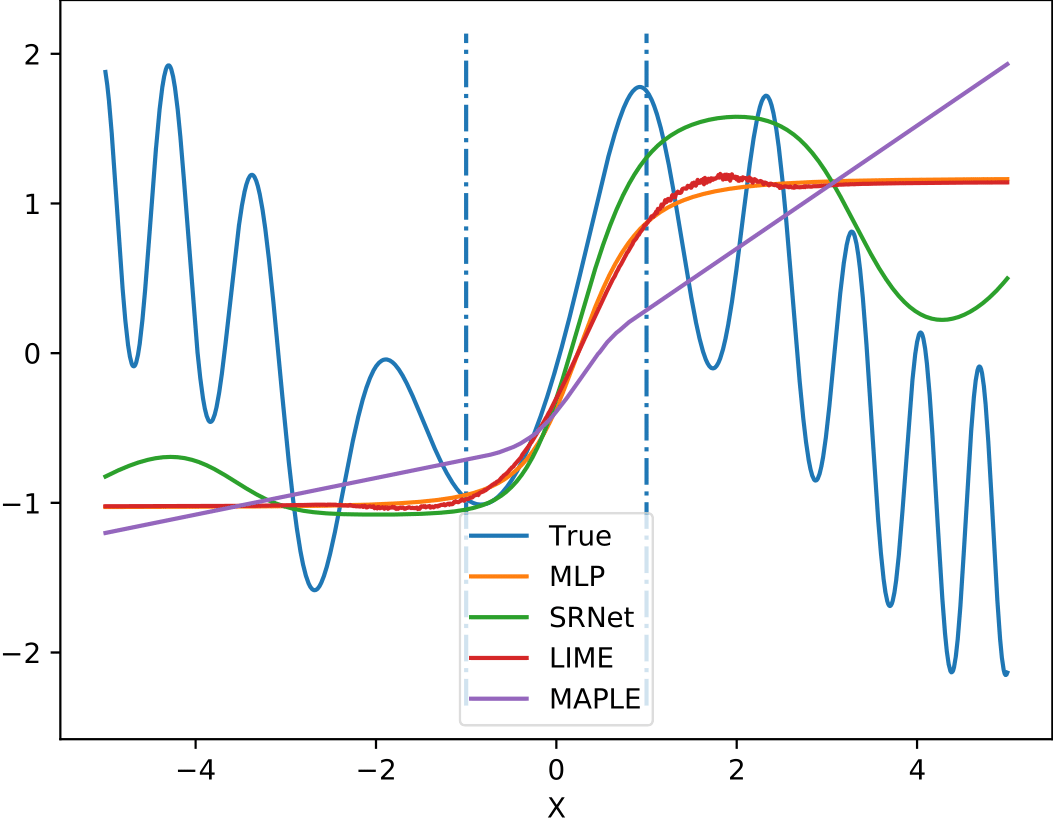}
		\caption{K0}
	\end{subfigure}
	\begin{subfigure}{0.24\textwidth}
		\includegraphics[width=\textwidth]{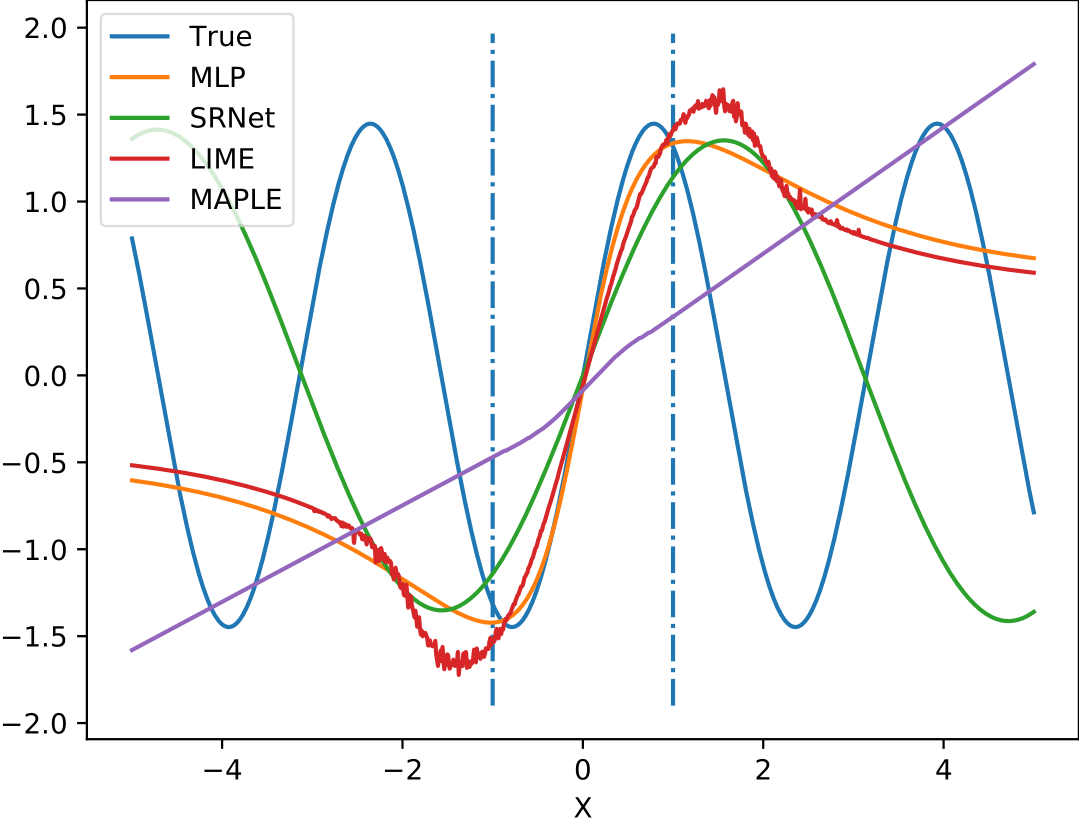}
		\caption{K1}
	\end{subfigure}
	\begin{subfigure}{0.24\textwidth}
		\includegraphics[width=\textwidth]{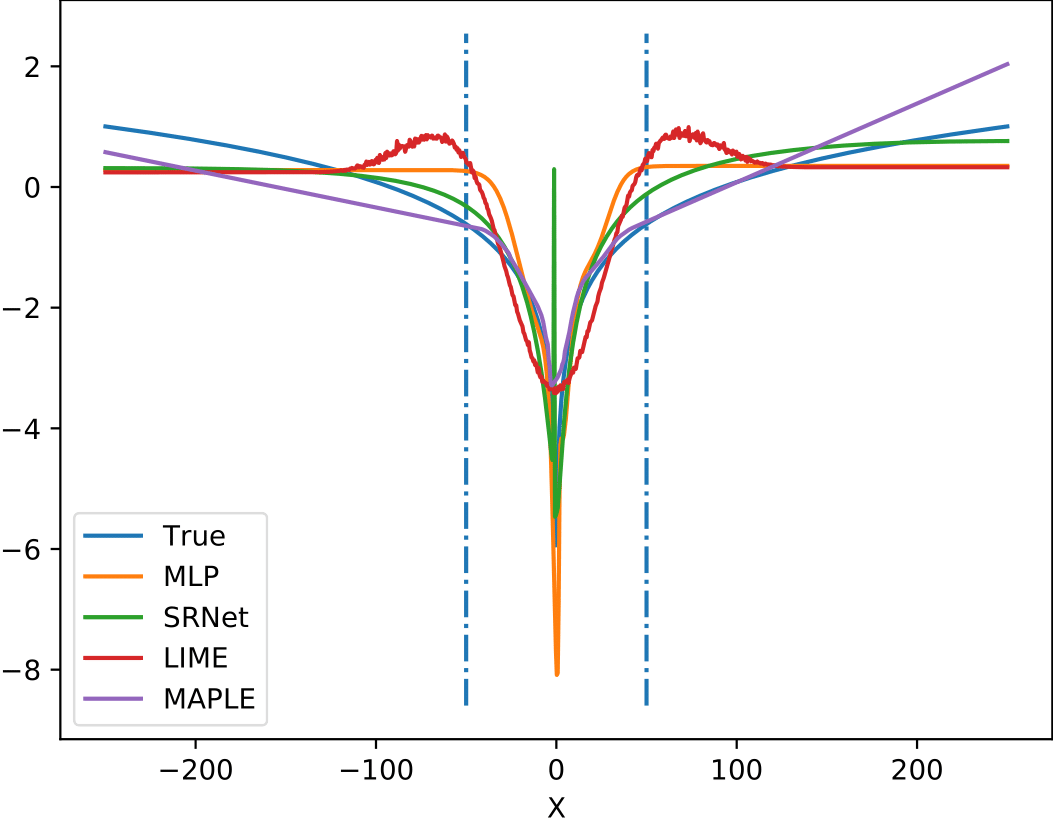}
		\caption{K2}
	\end{subfigure}
	\begin{subfigure}{0.24\textwidth}
		\includegraphics[width=\textwidth]{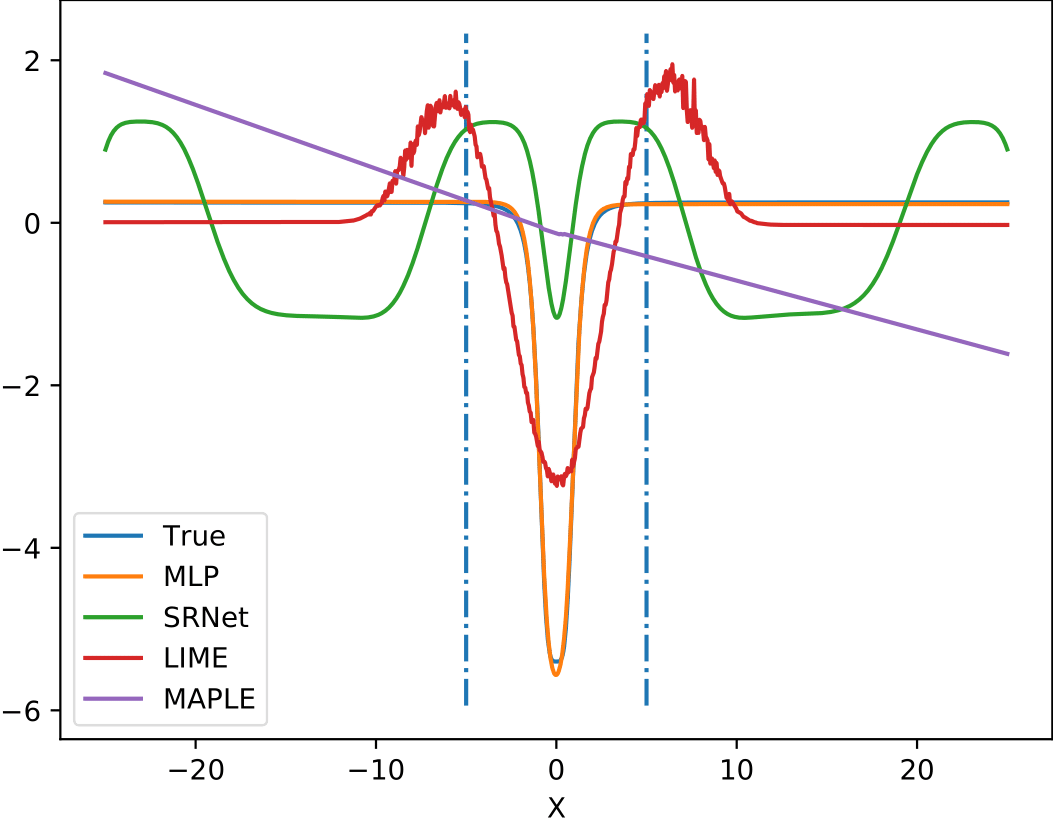}
		\caption{K3}
	\end{subfigure}
\\
	\begin{subfigure}{0.24\textwidth}
		\includegraphics[width=\textwidth]{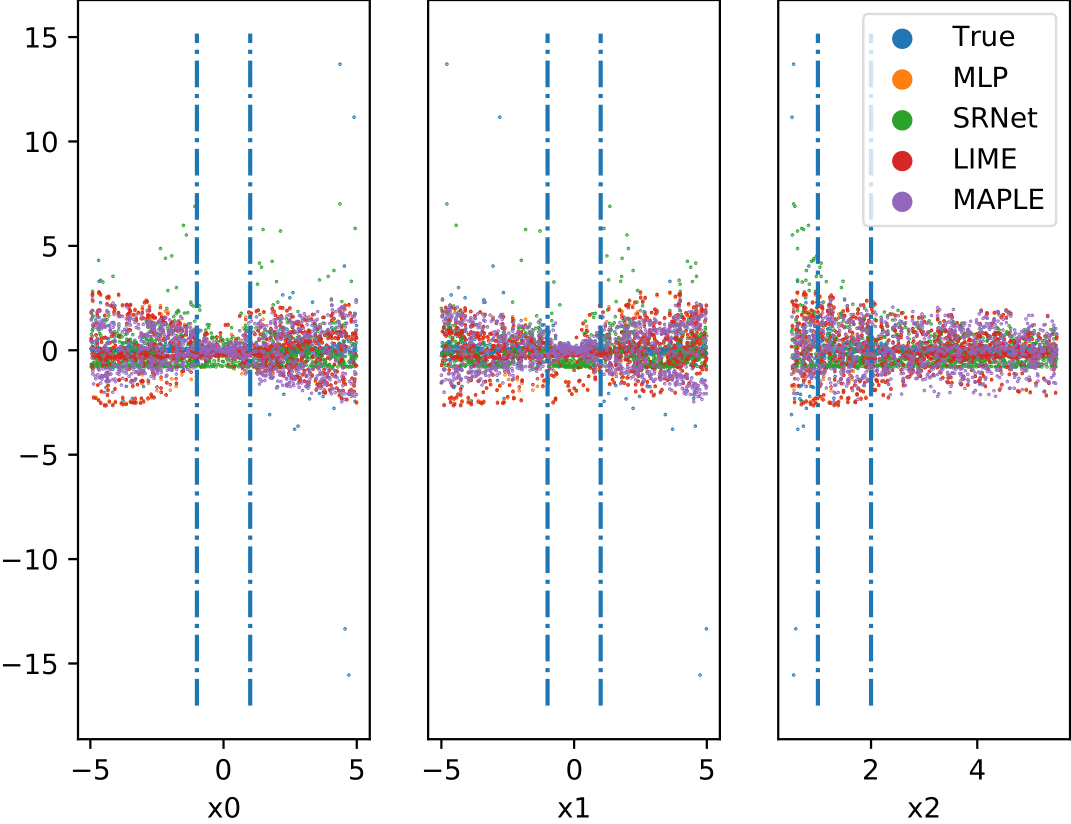}
		\caption{K4}
	\end{subfigure}
	\begin{subfigure}{0.24\textwidth}
		\includegraphics[width=\textwidth]{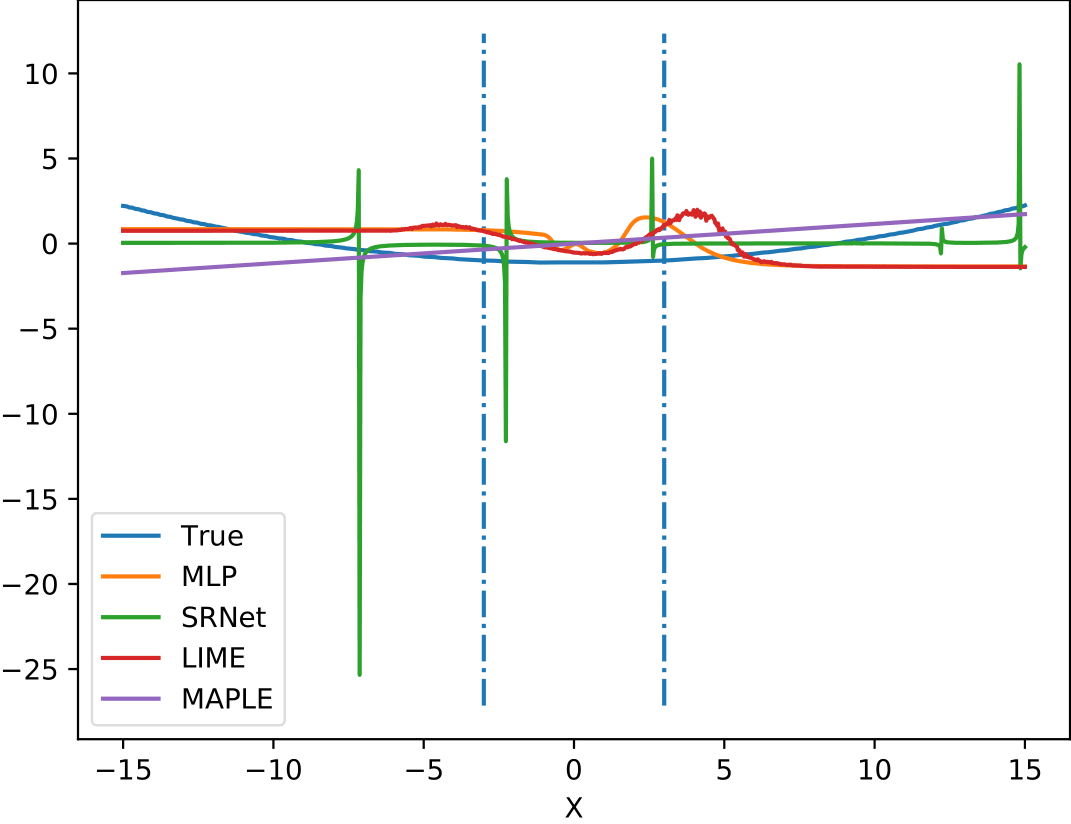}
		\caption{K5}
	\end{subfigure}
	\begin{subfigure}{0.24\textwidth}
		\includegraphics[width=\textwidth]{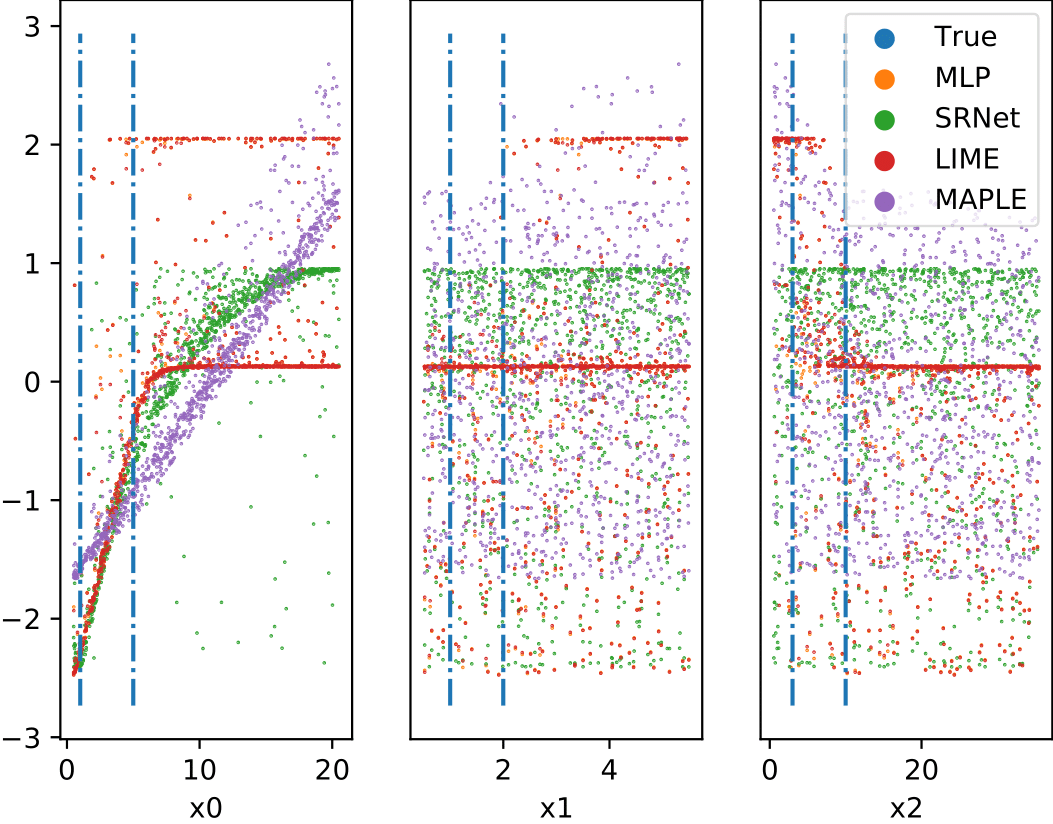}
		\caption{F0}
	\end{subfigure}
	\begin{subfigure}{0.24\textwidth}
		\includegraphics[width=\textwidth]{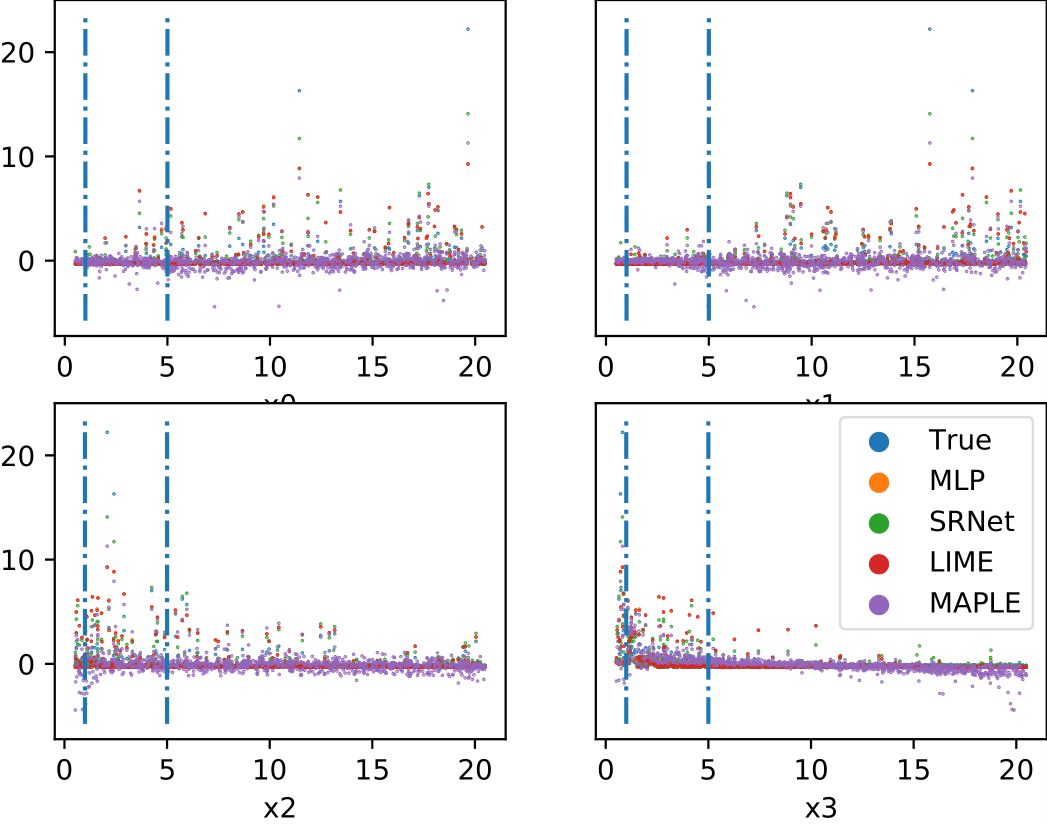}
		\caption{F1}
	\end{subfigure}
\\
	\begin{subfigure}{0.24\textwidth}
		\includegraphics[width=\textwidth]{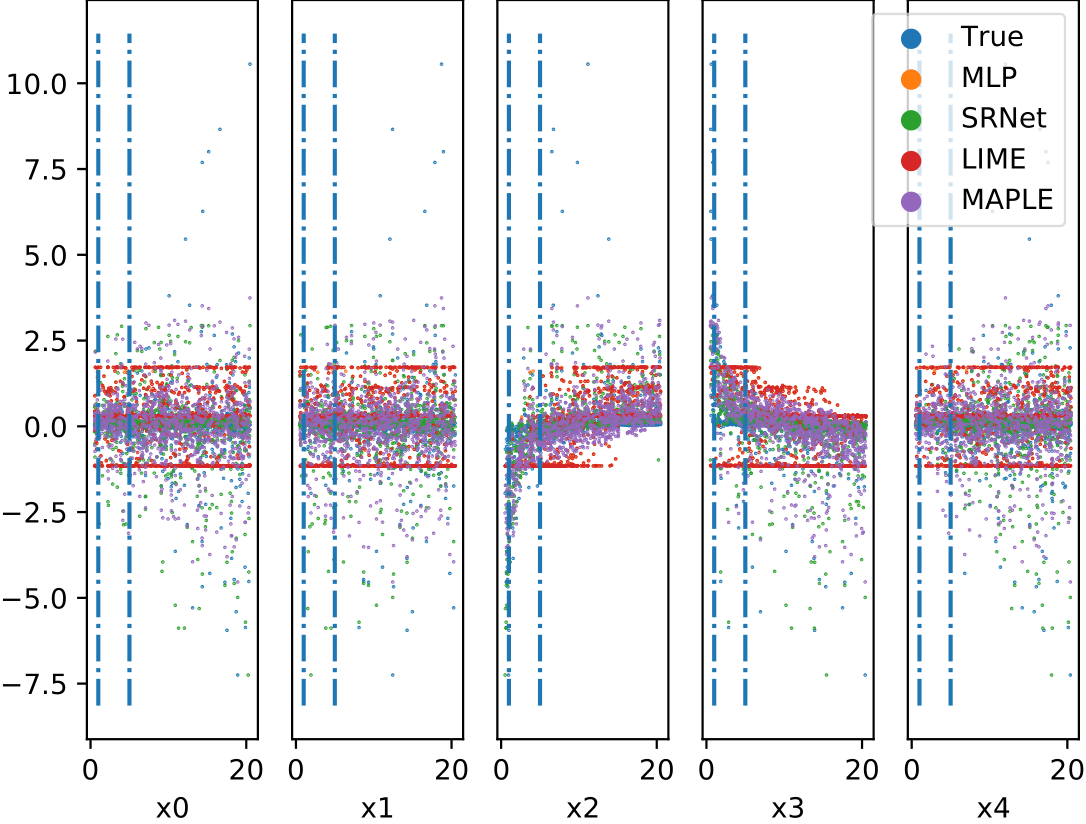}
		\caption{F2}
	\end{subfigure}
	\begin{subfigure}{0.24\textwidth}
		\includegraphics[width=\textwidth]{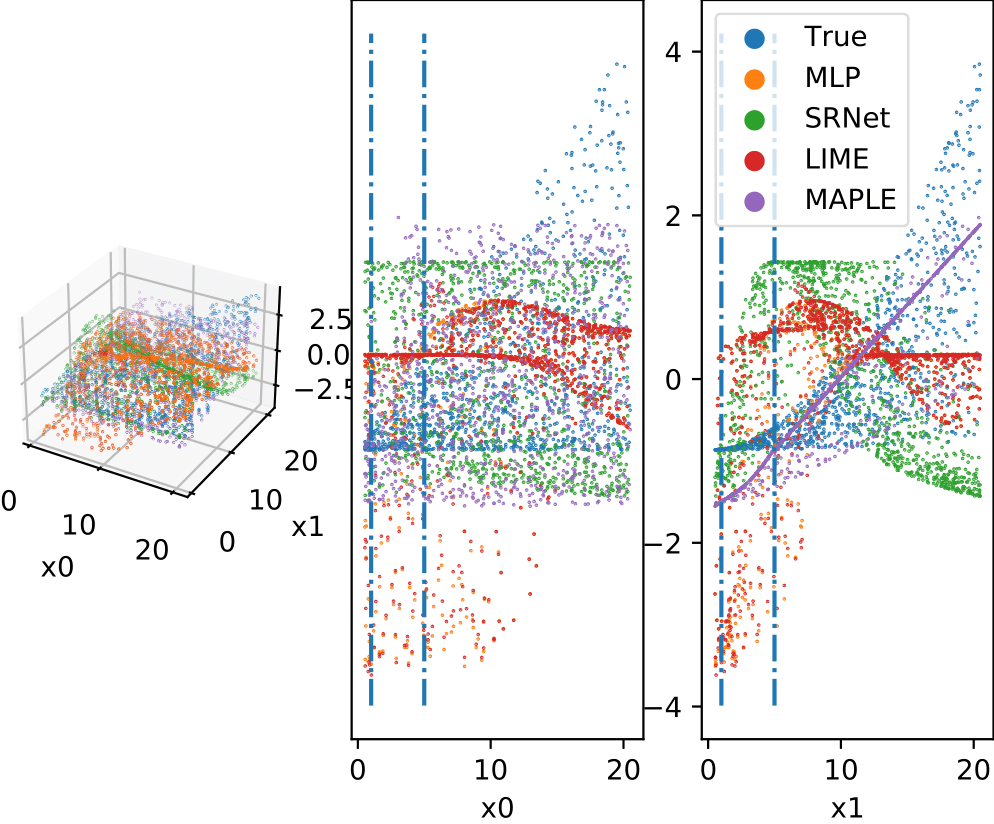}
		\caption{F3}
	\end{subfigure}
	\begin{subfigure}{0.24\textwidth}
		\includegraphics[width=\textwidth]{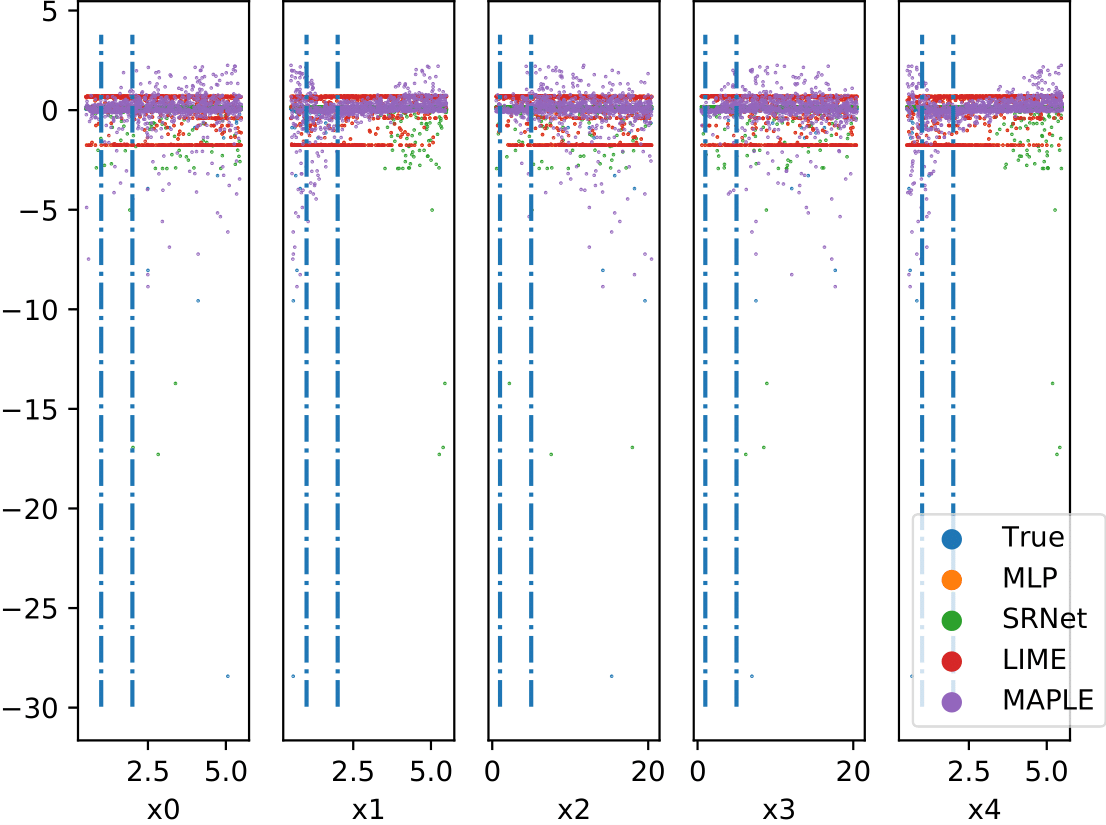}
		\caption{F4}
	\end{subfigure}
	\begin{subfigure}{0.24\textwidth}
		\includegraphics[width=\textwidth]{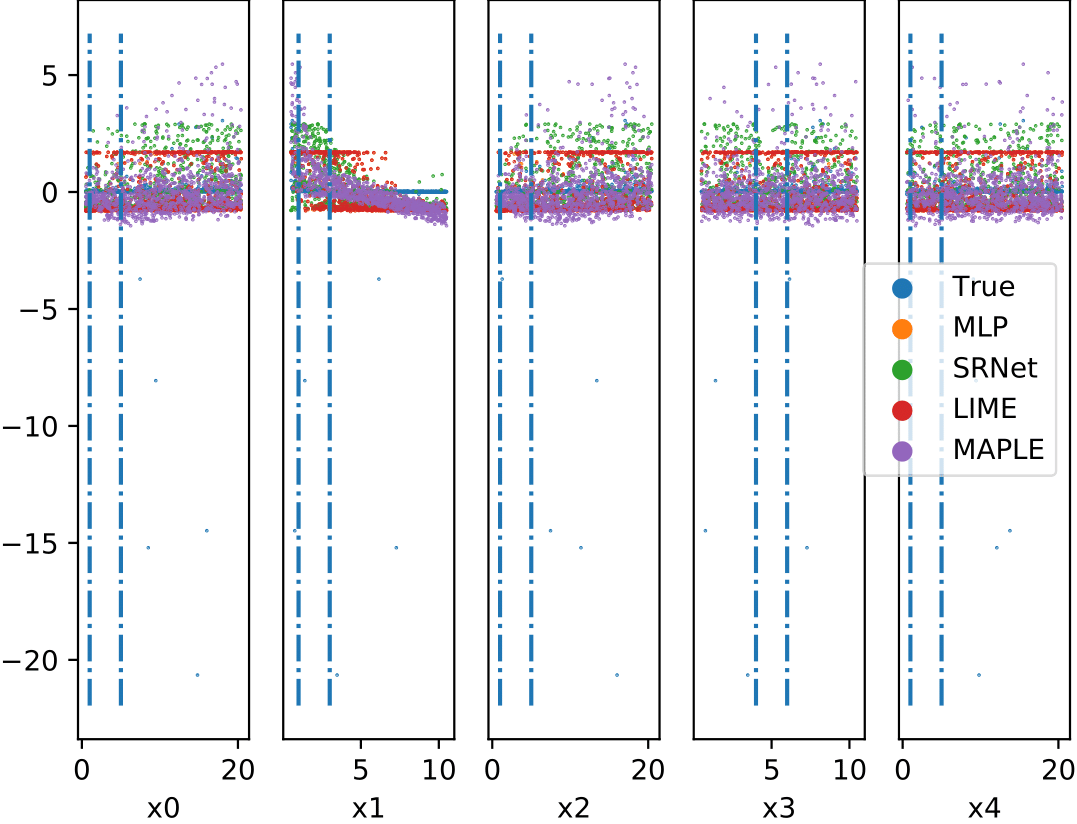}
		\caption{F5}
	\end{subfigure}
	
	\caption{SRNet vs LIME vs MAPLE on the interpolation and extrapolation domain of 12 SR benchmarks. The area between two blue vertical lines is the interpolation domain. The other area is the extrapolation domain}
	\label{append_fig:extrapolation}
\end{figure*}
\newpage
\subsection{Fitting of Hidden Neurons}
\label{appendix_sec:hidden_fitting}
% Figure \ref{append_fig:hidden_heat_map} shows the comparison of outputs of the SRNet layer vs the NN layer with 9 random input values on 12 SR benchmarks.
\begin{figure*}[ht]
	\centering
	\begin{subfigure}{0.23\linewidth}
	    \caption{K0-h0}
		\includegraphics[width=\linewidth]{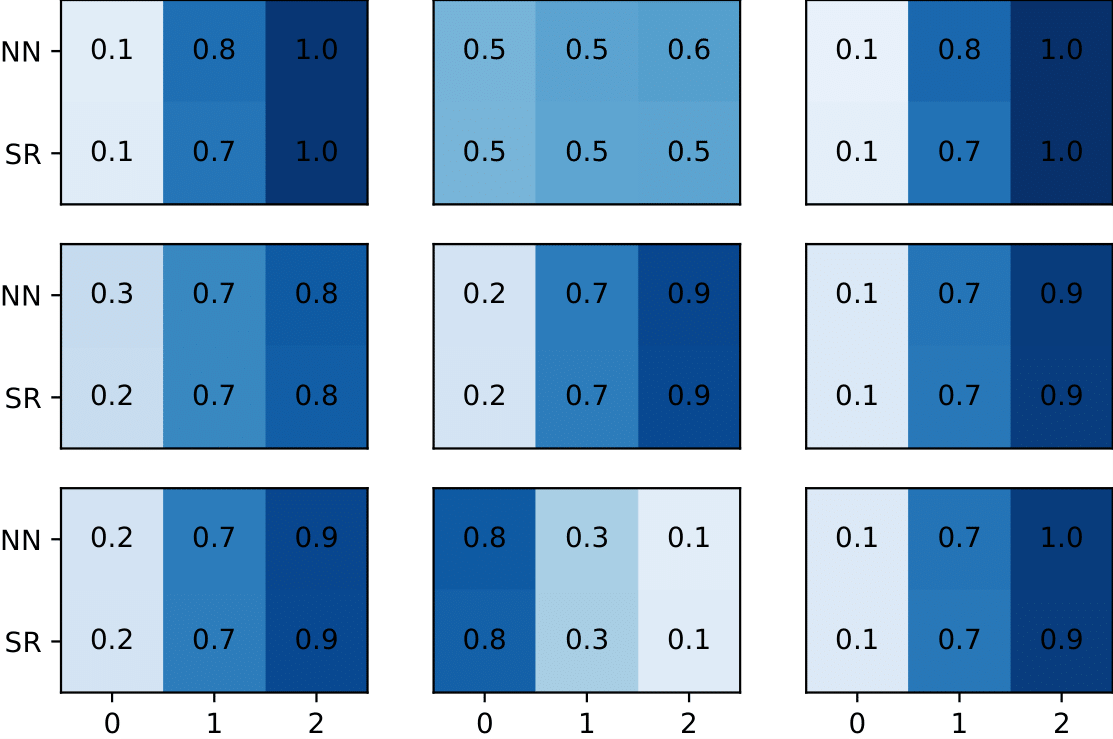}
	\end{subfigure}
	\begin{subfigure}{0.23\linewidth}
	    \caption{K0-h1}
		\includegraphics[width=\linewidth]{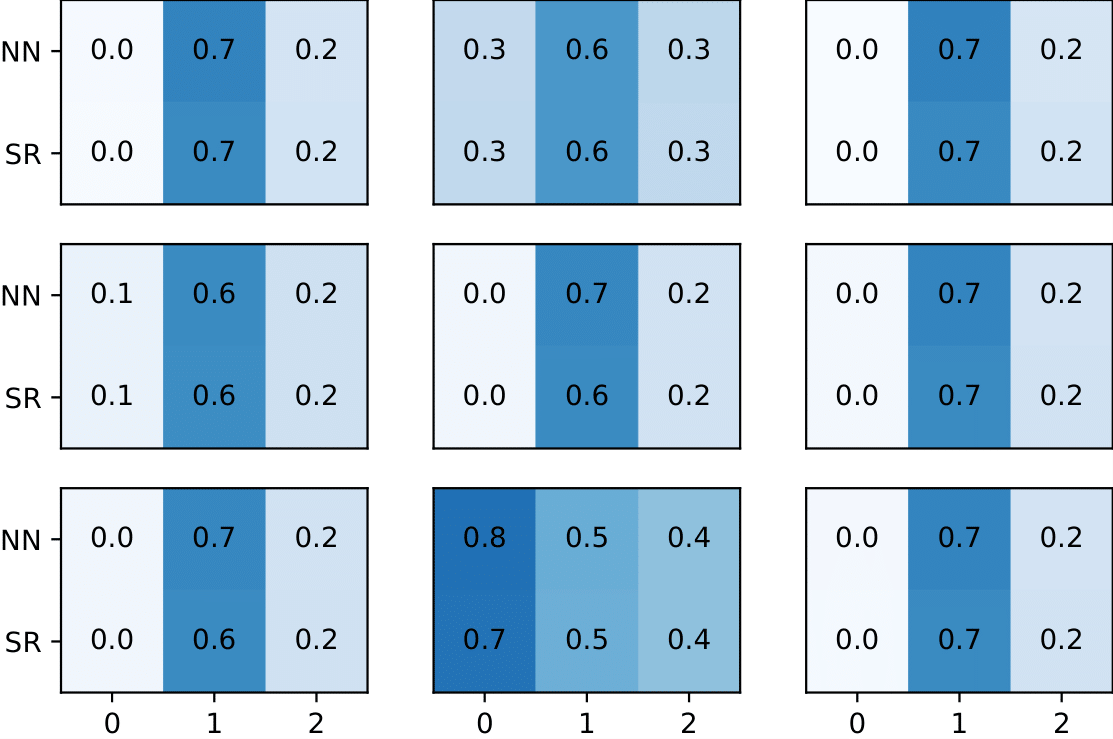}
	\end{subfigure}
	\begin{subfigure}{0.23\linewidth}
	    \caption{K1-h0}
		\includegraphics[width=\linewidth]{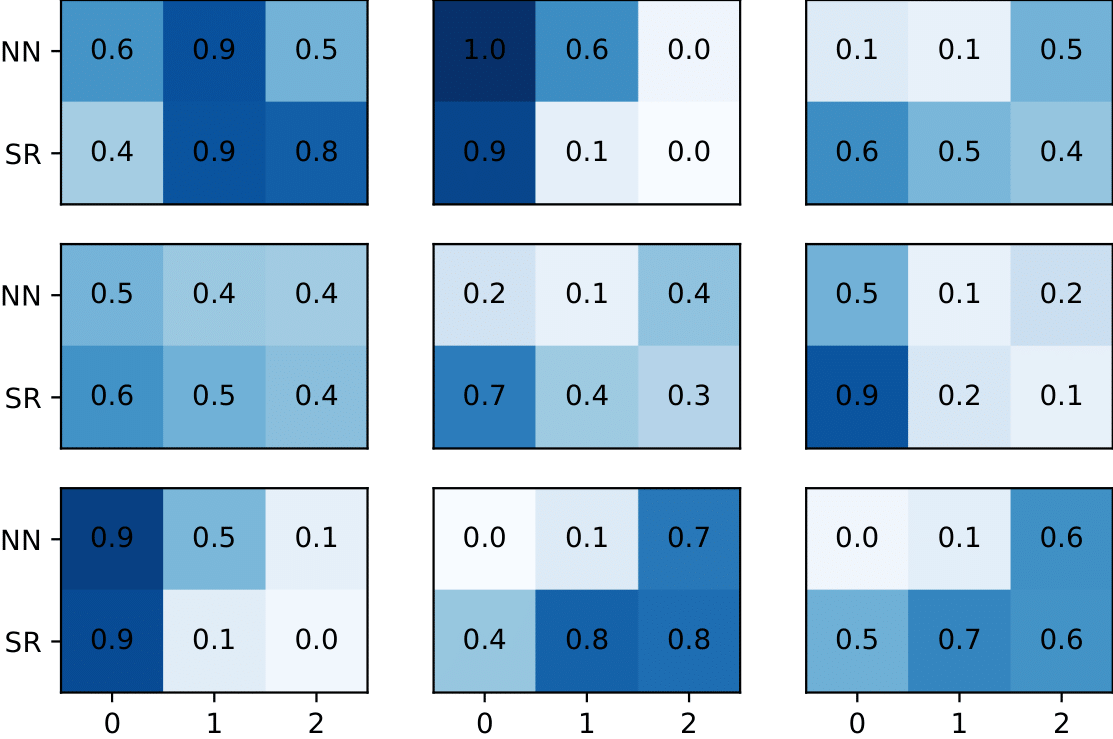}
	\end{subfigure}
	\begin{subfigure}{0.23\linewidth}
	    \caption{K1-h1}
		\includegraphics[width=\linewidth]{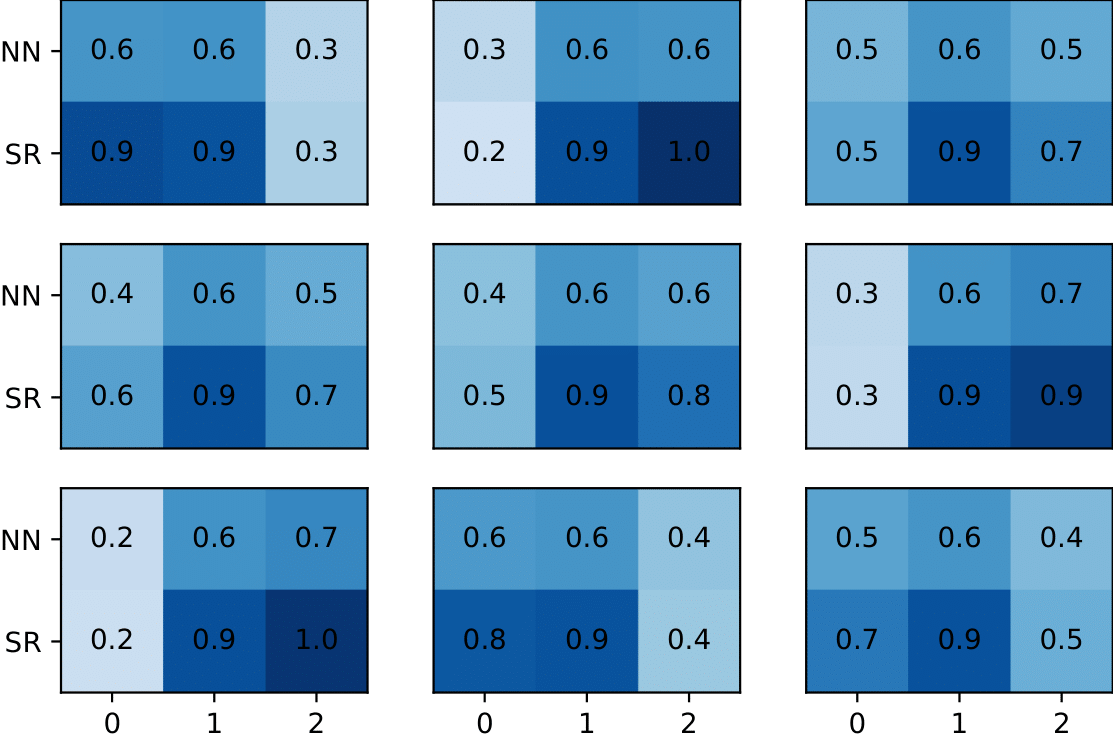}
	\end{subfigure}
\\
	\begin{subfigure}{0.23\linewidth}
	    \caption{K2-h0}
		\includegraphics[width=\linewidth]{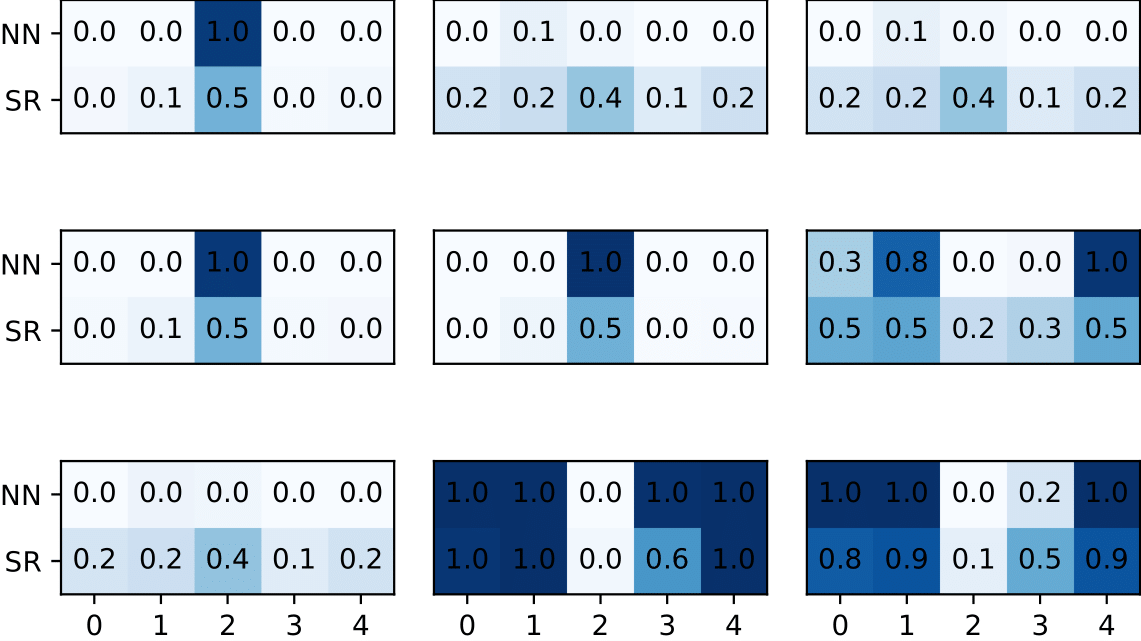}
	\end{subfigure}
	\begin{subfigure}{0.23\linewidth}
	    \caption{K2-h1}
		\includegraphics[width=\linewidth]{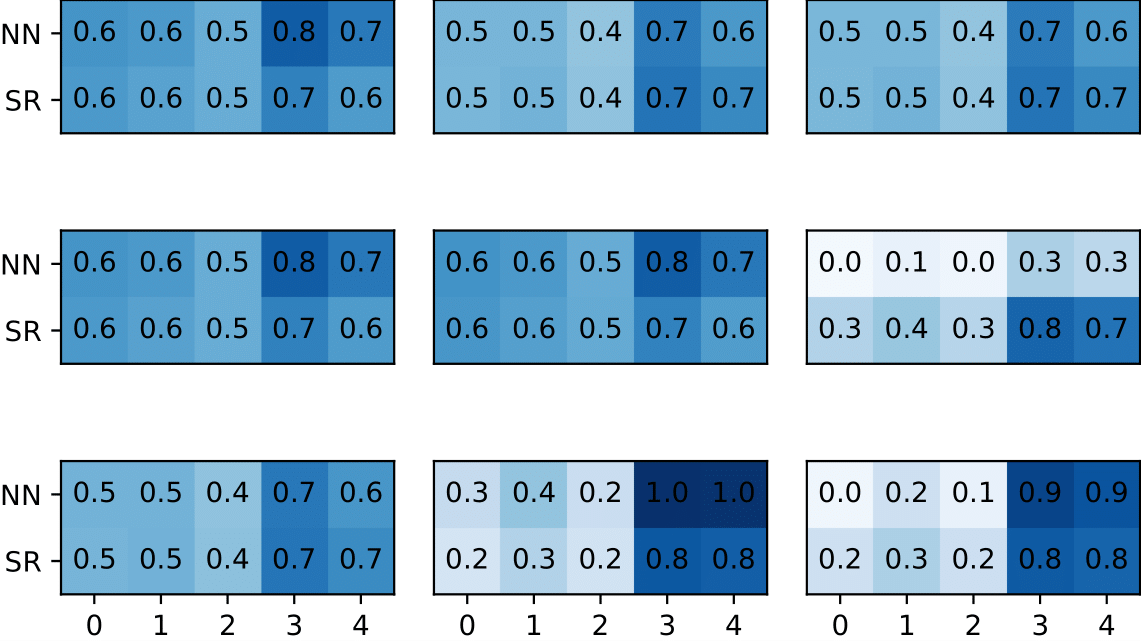}
	\end{subfigure}
	\begin{subfigure}{0.16\linewidth}
	    \caption{K3-h0}
		\includegraphics[width=\linewidth]{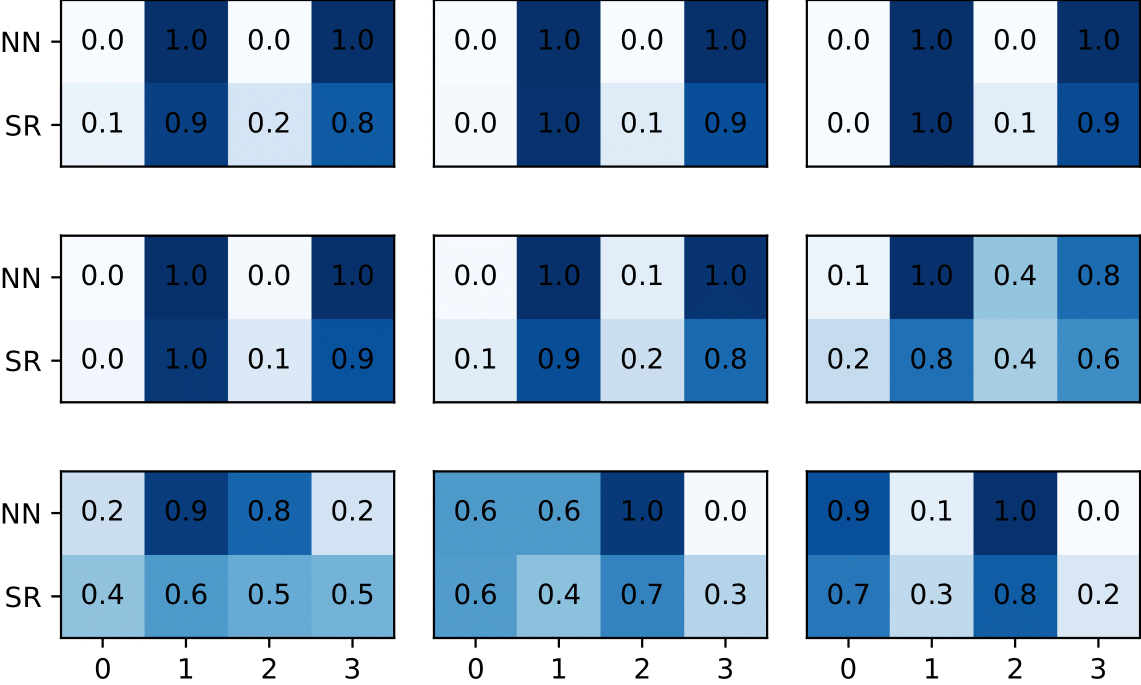}
	\end{subfigure}
	\begin{subfigure}{0.16\linewidth}
	    \caption{K3-h1}
		\includegraphics[width=\linewidth]{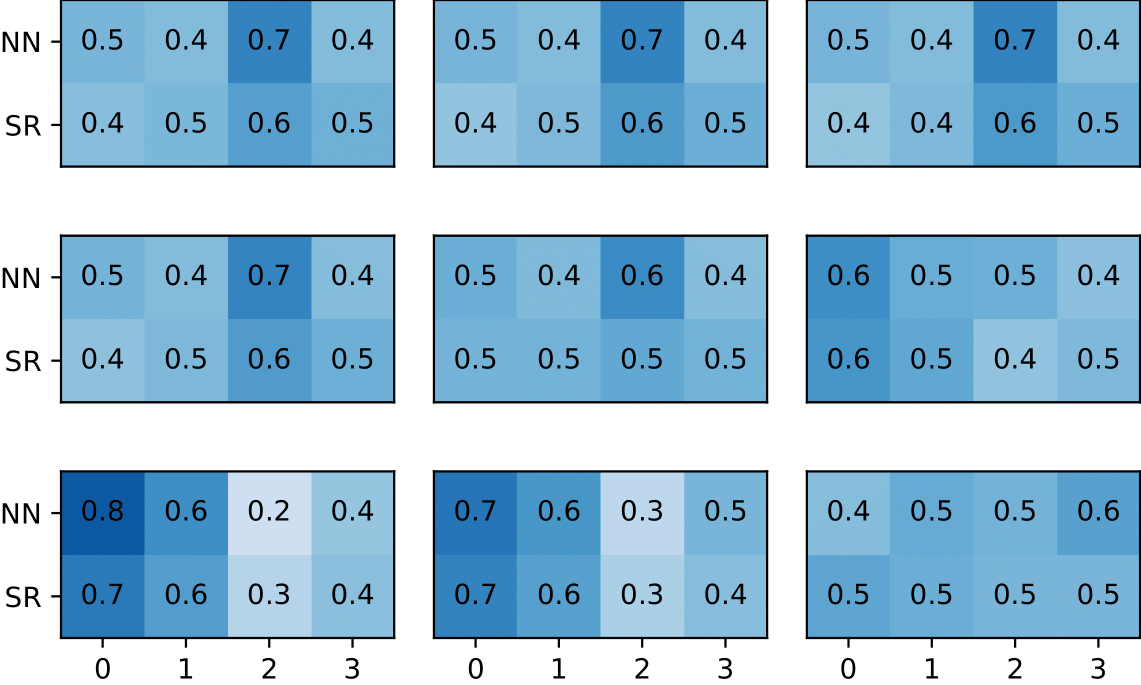}
	\end{subfigure}
	\begin{subfigure}{0.16\linewidth}
	    \caption{K3-h2}
		\includegraphics[width=\linewidth]{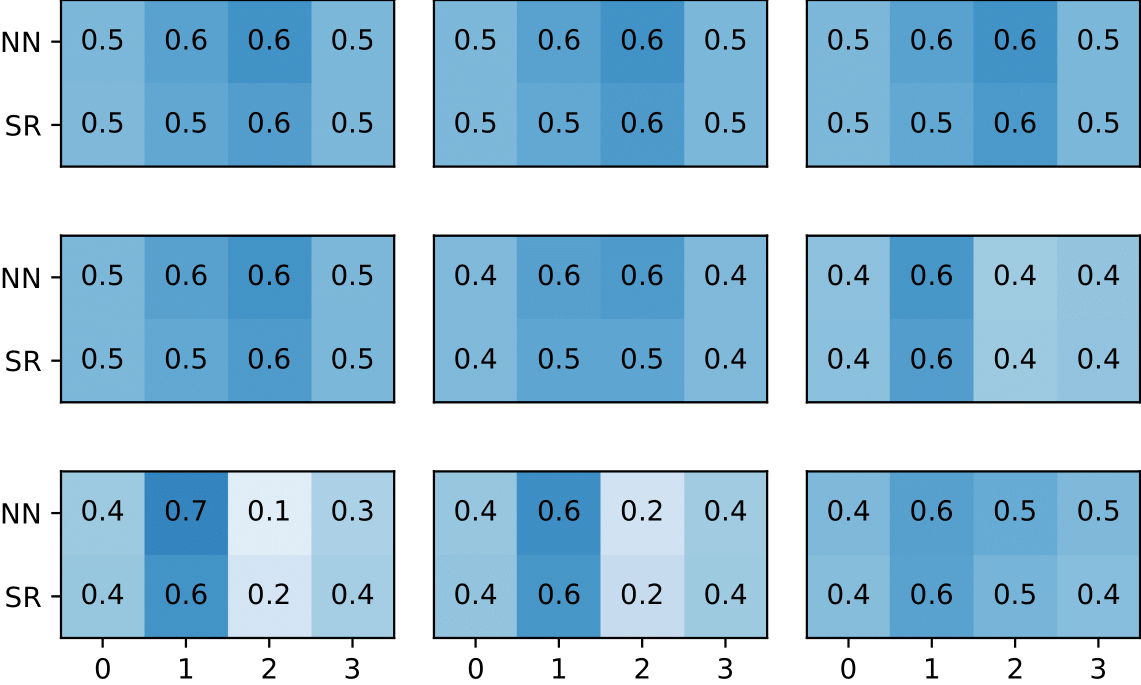}
	\end{subfigure}
\\
	\begin{subfigure}{0.23\linewidth}
	    \caption{K4-h0}
		\includegraphics[width=\linewidth]{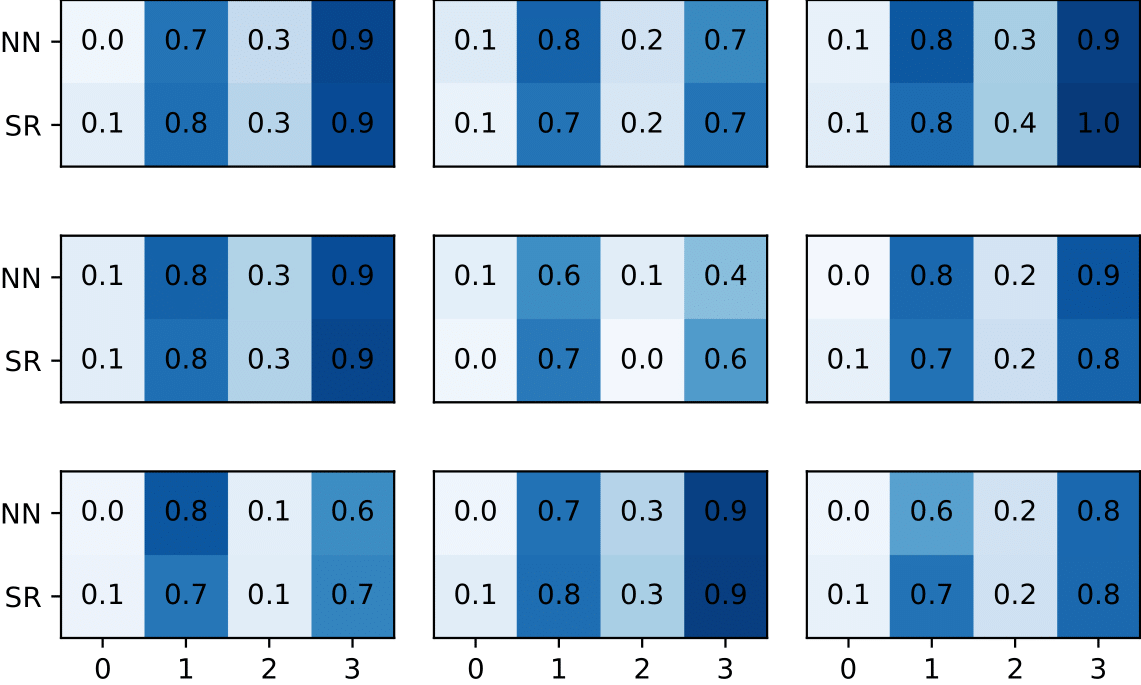}
	\end{subfigure}
	\begin{subfigure}{0.23\linewidth}
	    \caption{K4-h1}
		\includegraphics[width=\linewidth]{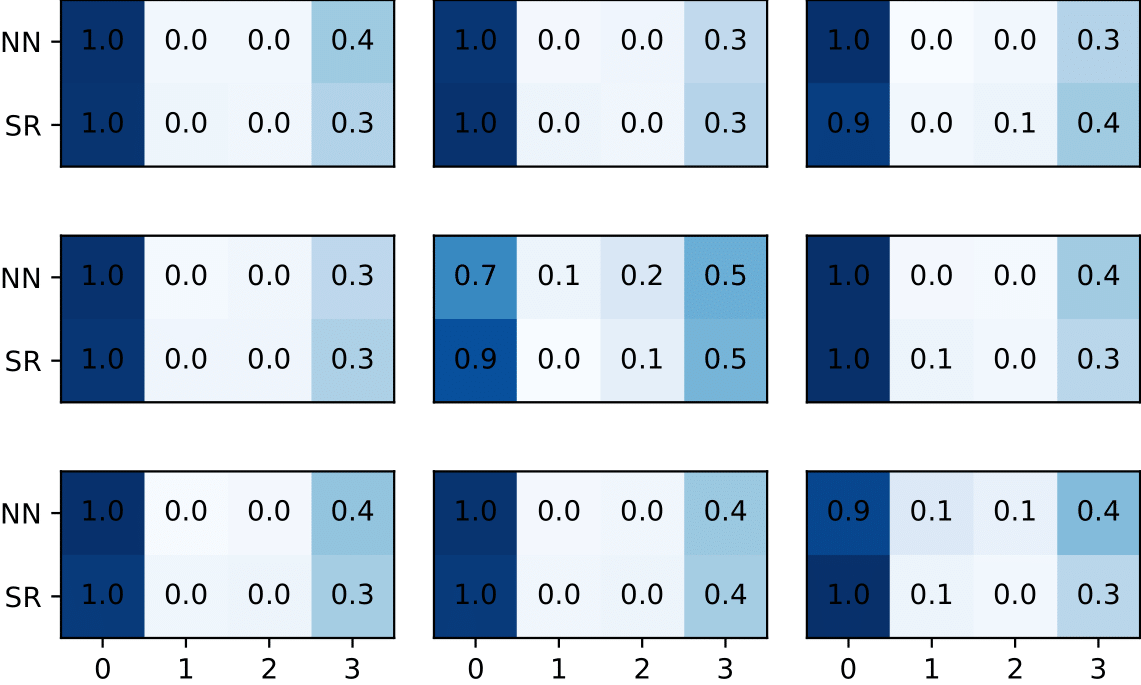}
	\end{subfigure}
	\begin{subfigure}{0.23\linewidth}
	    \caption{K5-h0}
		\includegraphics[width=\linewidth]{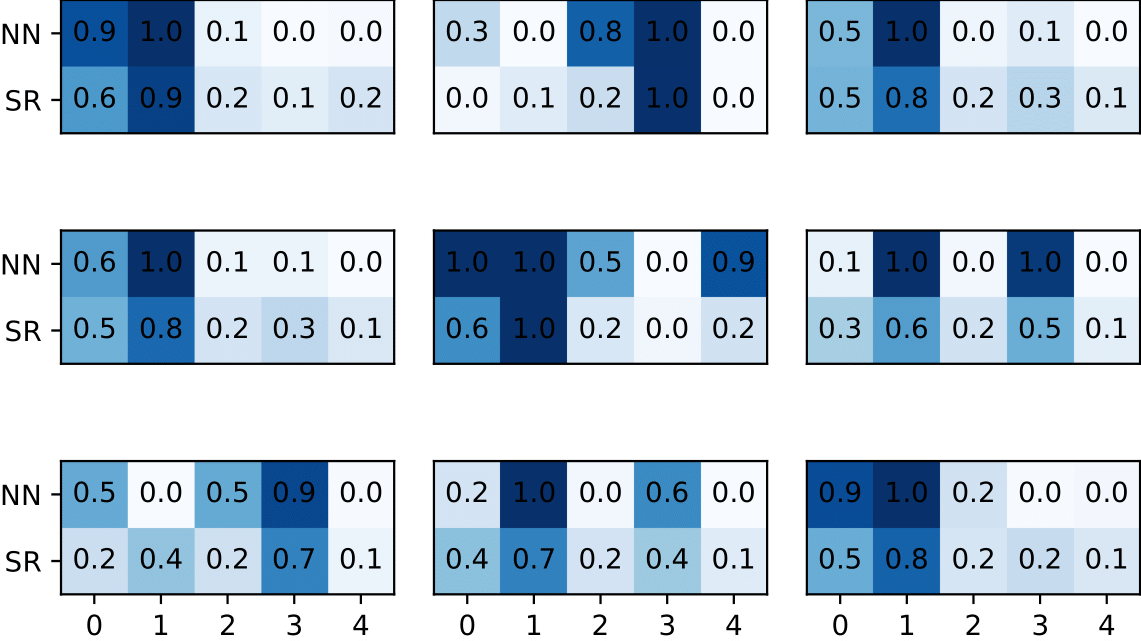}
	\end{subfigure}
	\begin{subfigure}{0.23\linewidth}
	    \caption{K5-h1}
		\includegraphics[width=\linewidth]{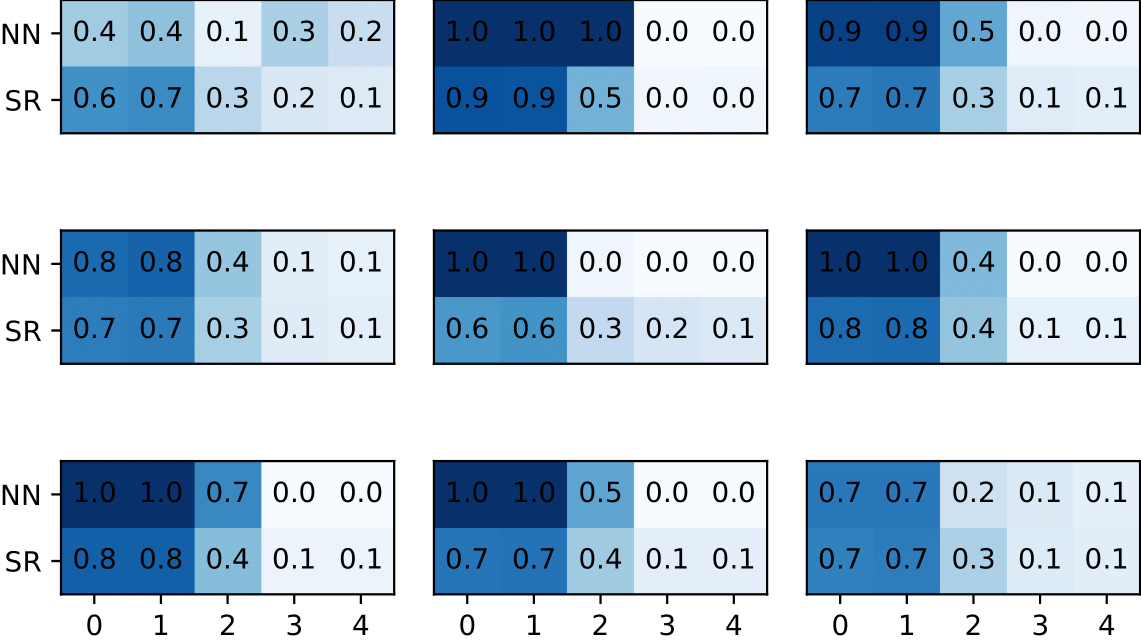}
	\end{subfigure}
\\
	\begin{subfigure}{0.23\linewidth}
	    \caption{F0-h0}
		\includegraphics[width=\linewidth]{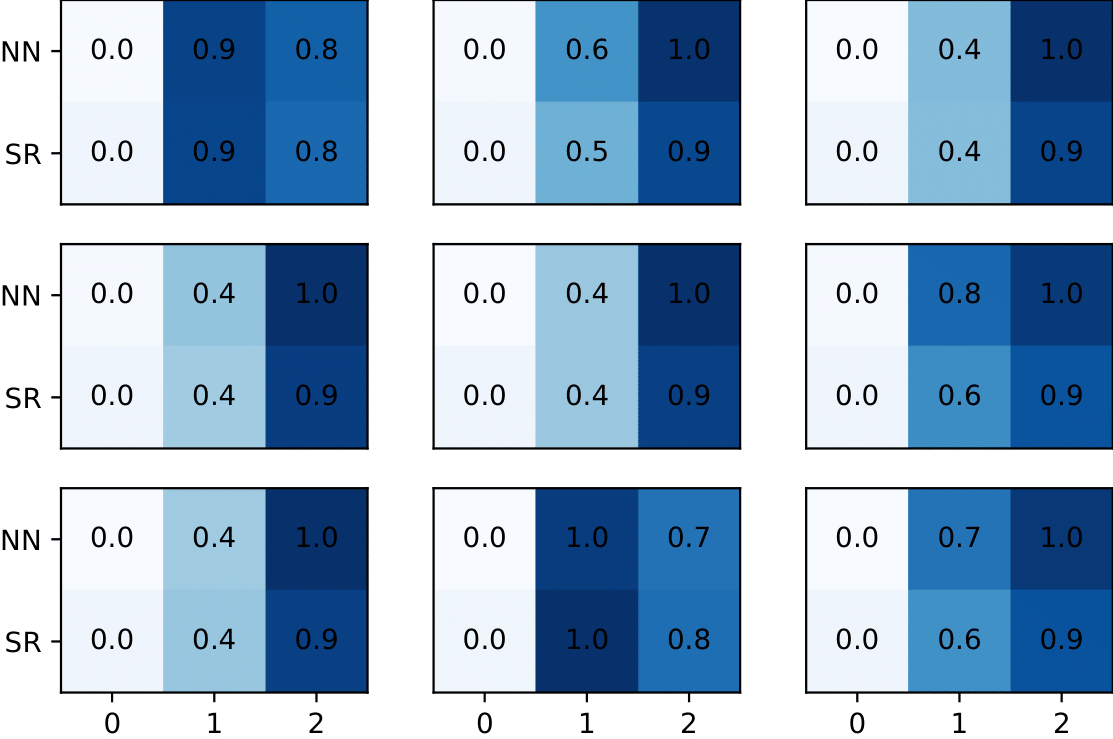}
	\end{subfigure}
	\begin{subfigure}{0.23\linewidth}
	    \caption{F0-h1}
		\includegraphics[width=\linewidth]{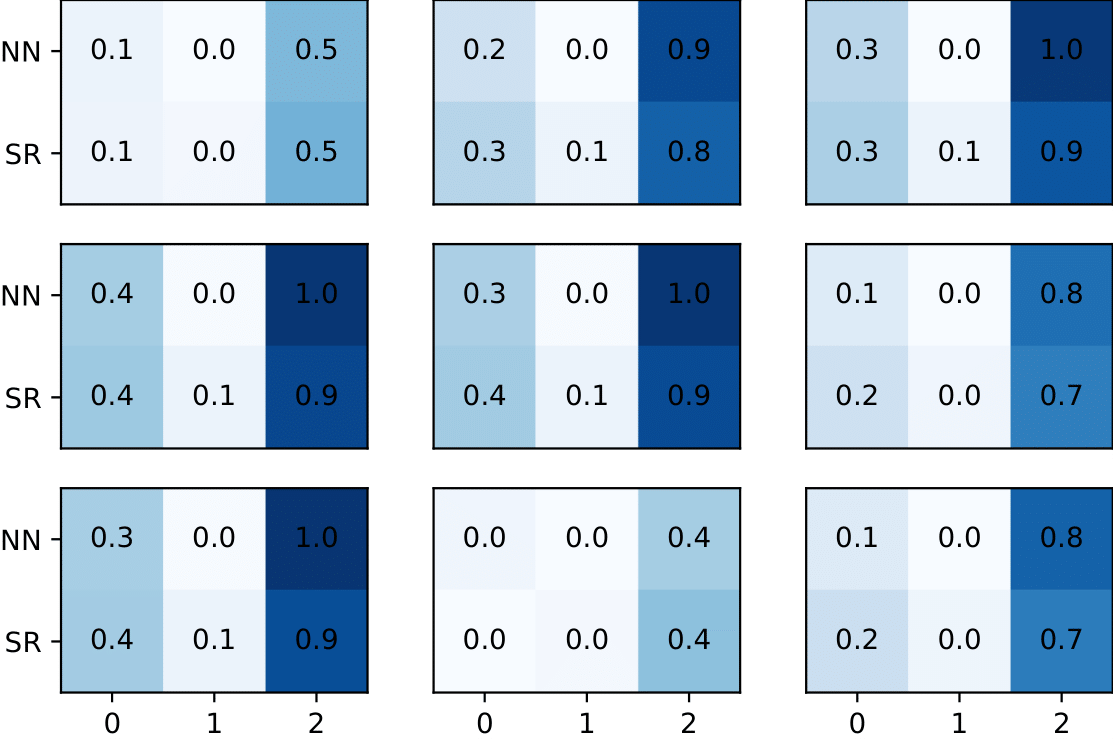}
	\end{subfigure}
	\begin{subfigure}{0.23\linewidth}
	    \caption{F1-h0}
		\includegraphics[width=\linewidth]{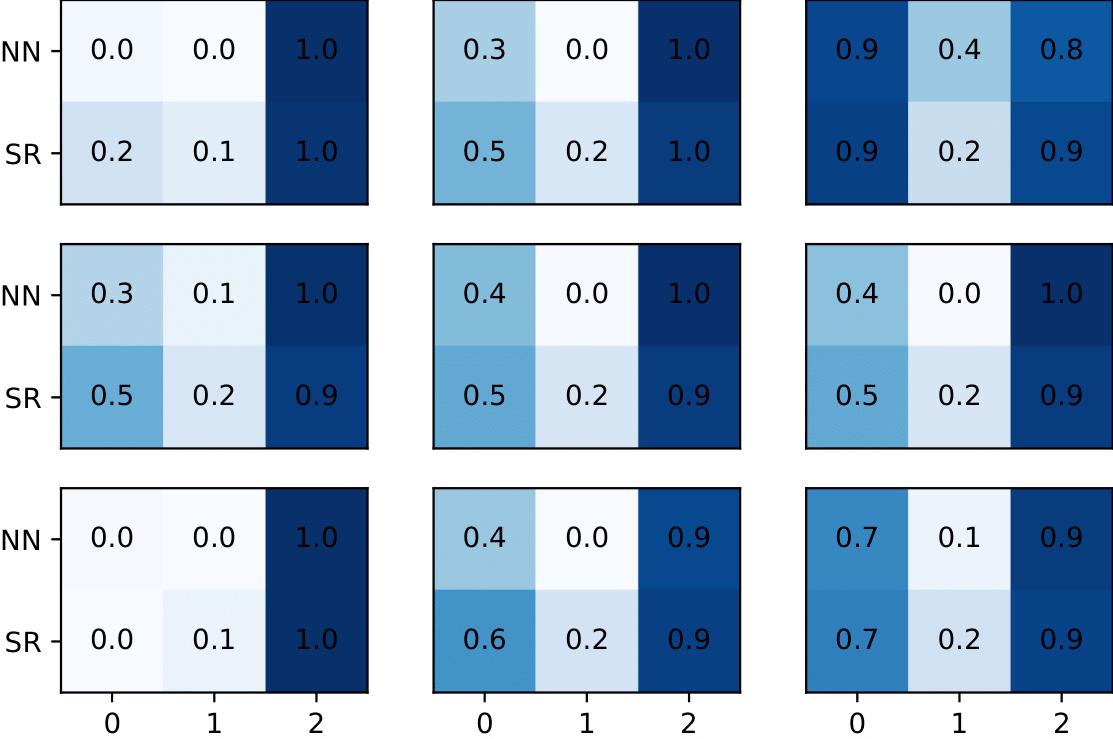}
	\end{subfigure}
	\begin{subfigure}{0.23\linewidth}
	    \caption{F1-h1}
		\includegraphics[width=\linewidth]{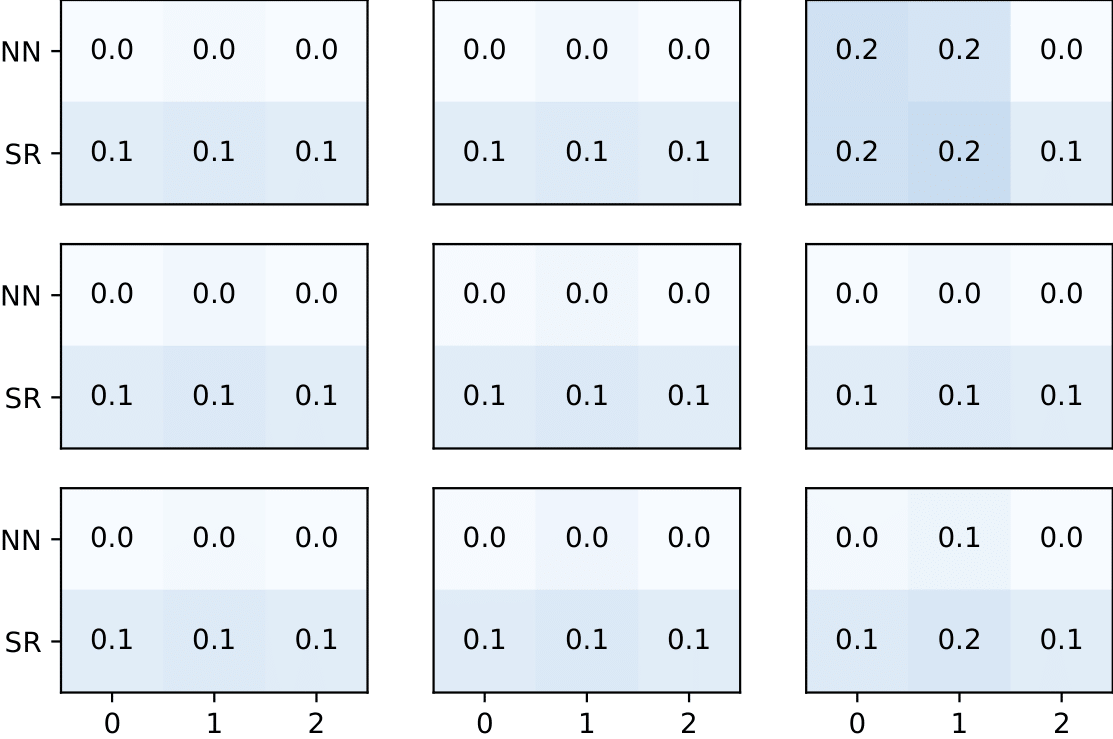}
	\end{subfigure}
\\
	\begin{subfigure}{0.23\linewidth}
	    \caption{F2-h0}
		\includegraphics[width=\linewidth]{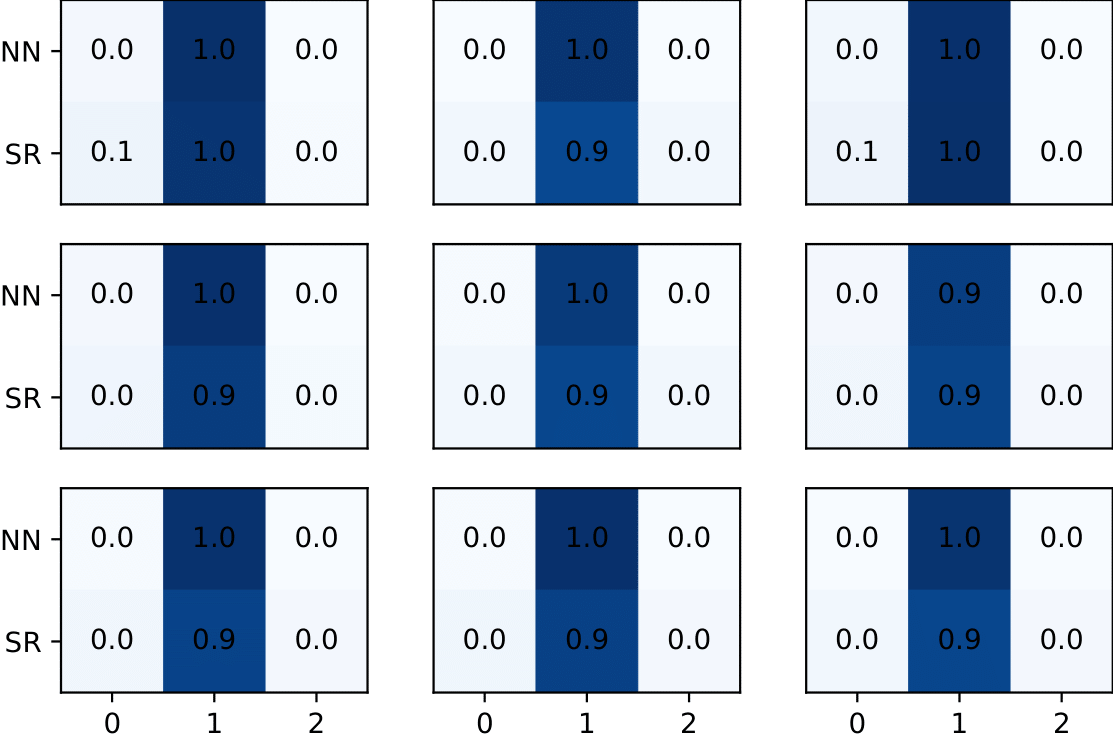}
	\end{subfigure}
	\begin{subfigure}{0.23\linewidth}
	    \caption{F2-h1}
		\includegraphics[width=\linewidth]{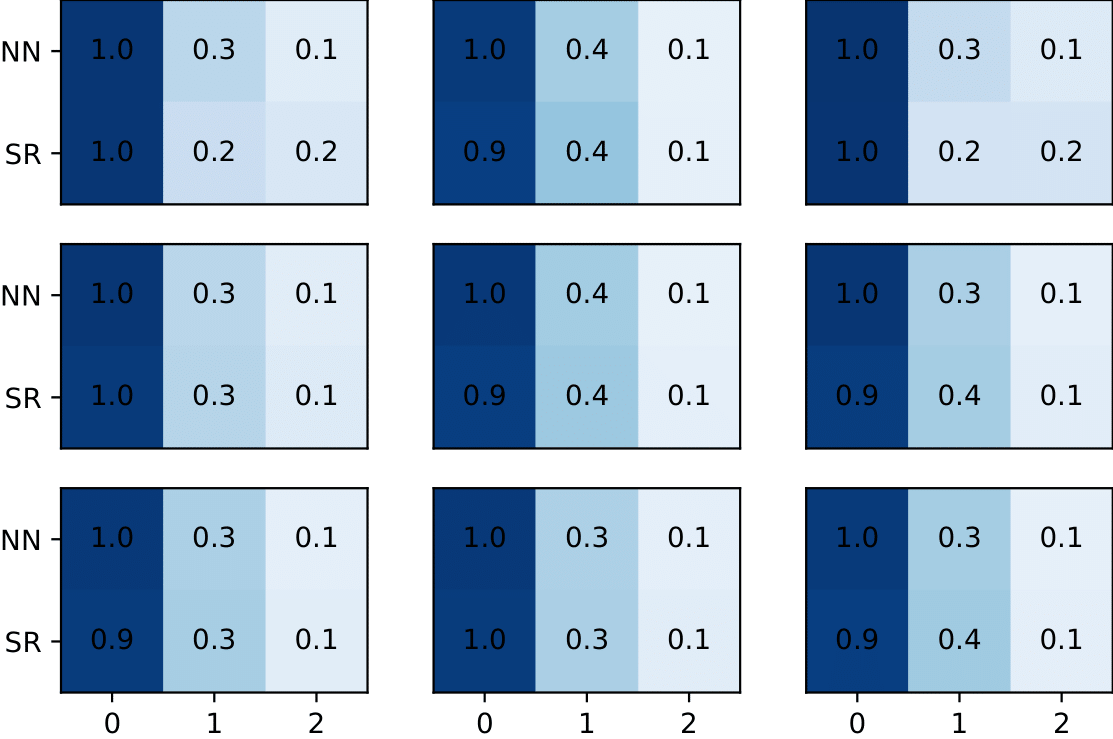}
	\end{subfigure}
	\begin{subfigure}{0.23\linewidth}
	    \caption{F3-h0}
		\includegraphics[width=\linewidth]{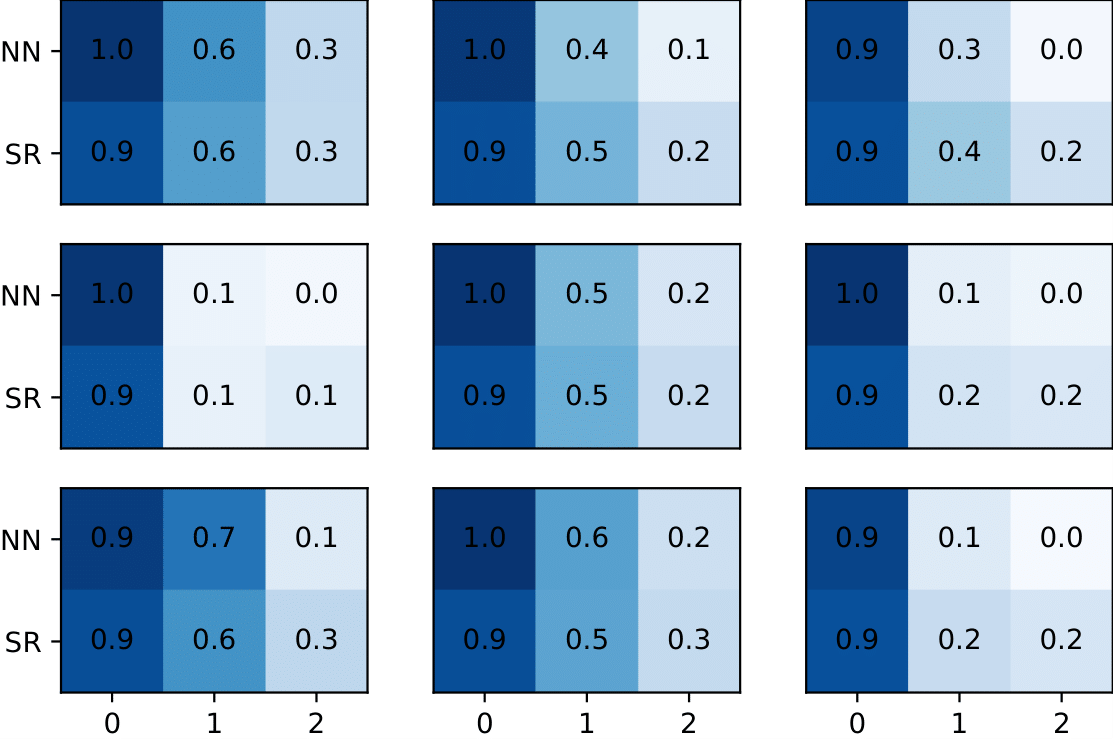}
	\end{subfigure}
	\begin{subfigure}{0.23\linewidth}
	    \caption{F3-h1}
		\includegraphics[width=\linewidth]{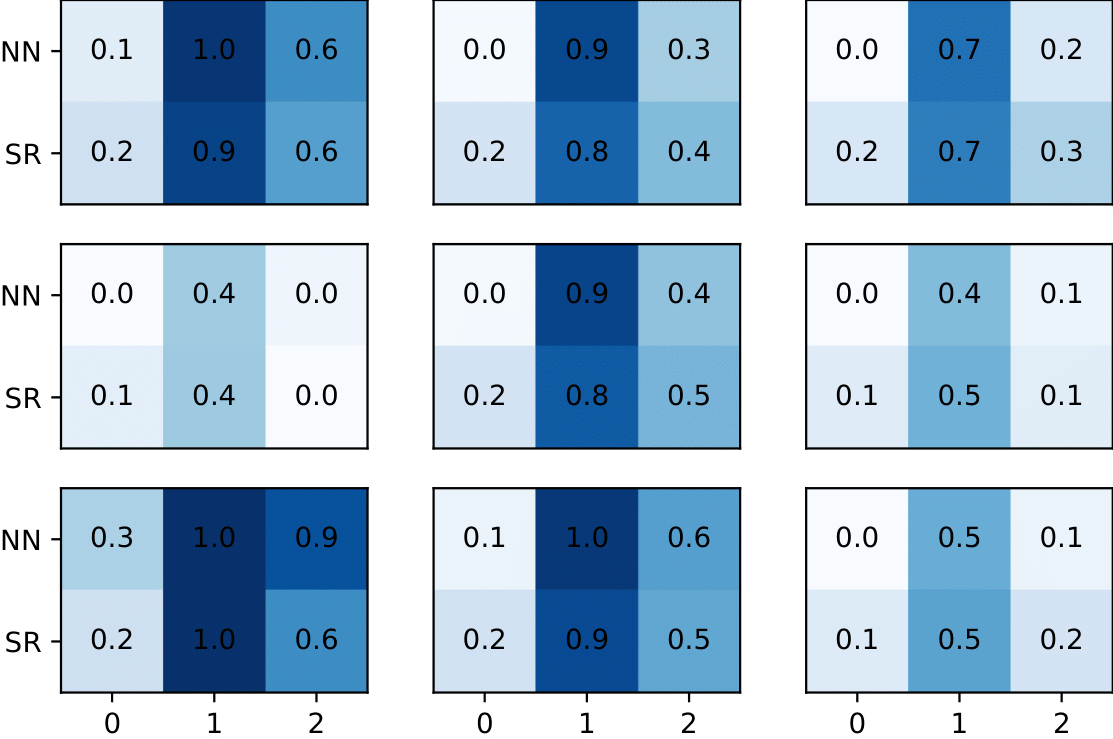}
	\end{subfigure}
\\
	\begin{subfigure}{0.23\linewidth}
	    \caption{F4-h0}
		\includegraphics[width=\linewidth]{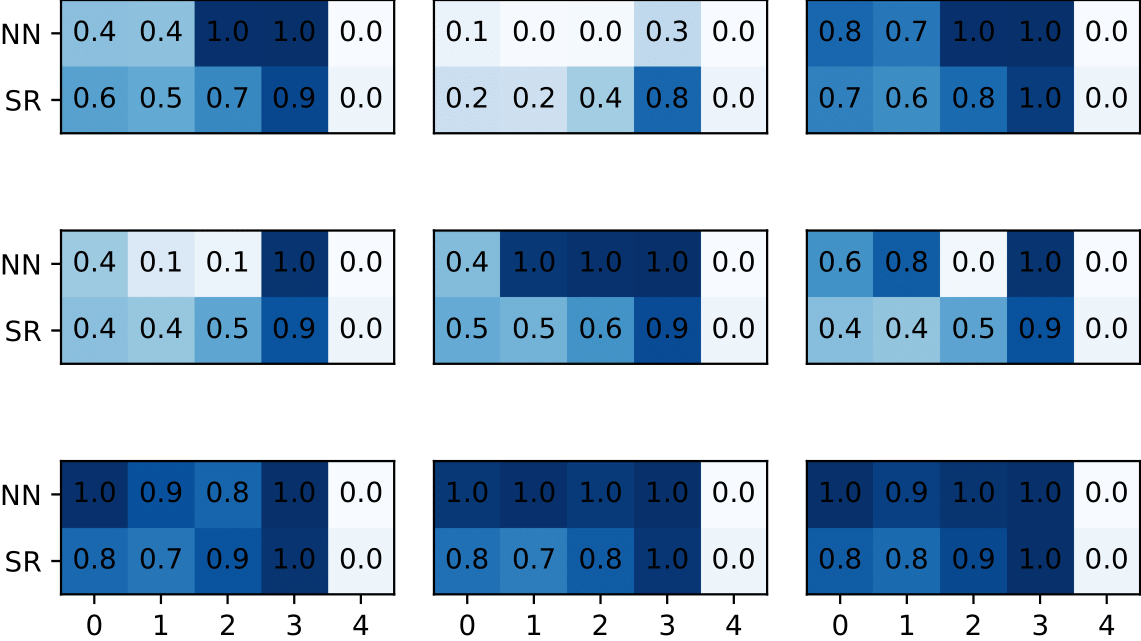}
	\end{subfigure}
	\begin{subfigure}{0.23\linewidth}
	    \caption{F4-h1}
		\includegraphics[width=\linewidth]{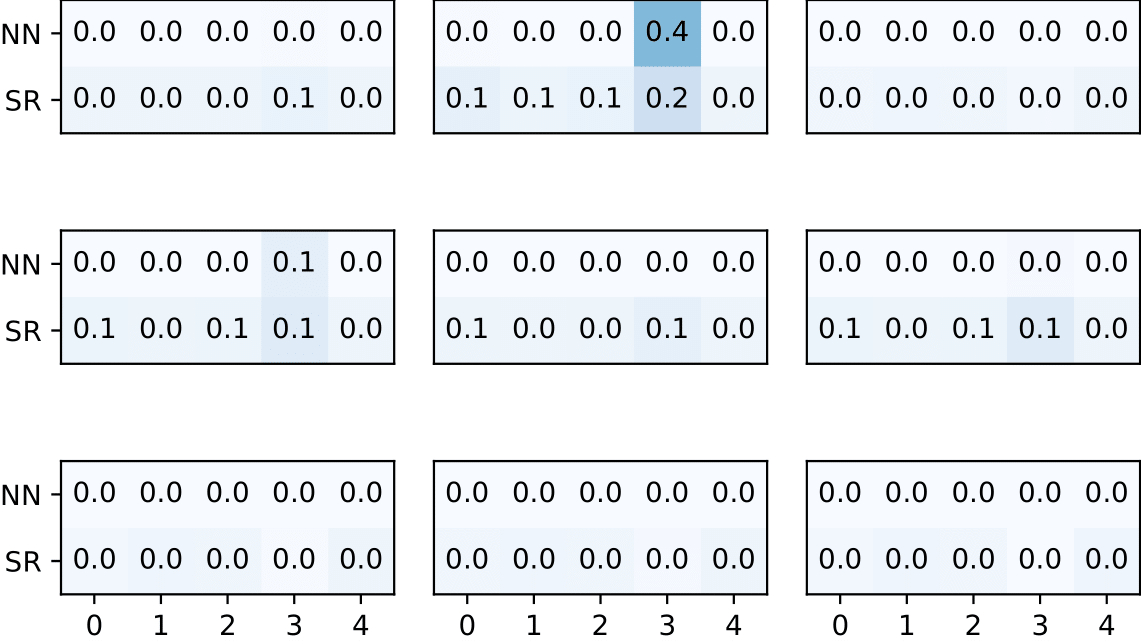}
	\end{subfigure}
	\begin{subfigure}{0.16\linewidth}
	    \caption{F5-h0}
		\includegraphics[width=\linewidth]{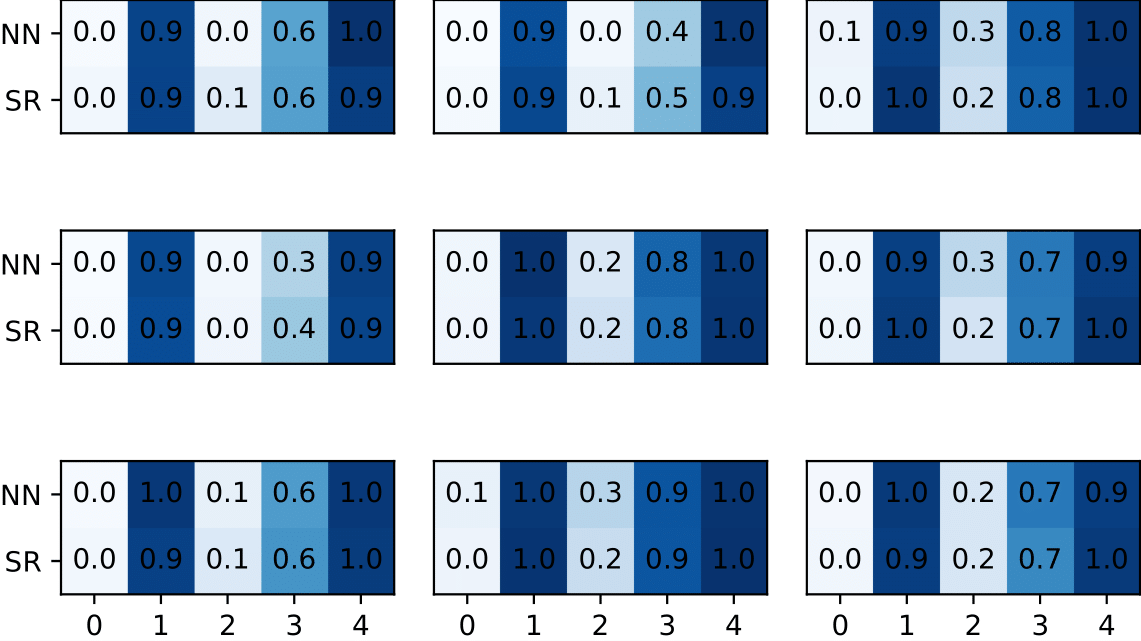}
	\end{subfigure}
	\begin{subfigure}{0.16\linewidth}
	    \caption{F5-h1}
		\includegraphics[width=\linewidth]{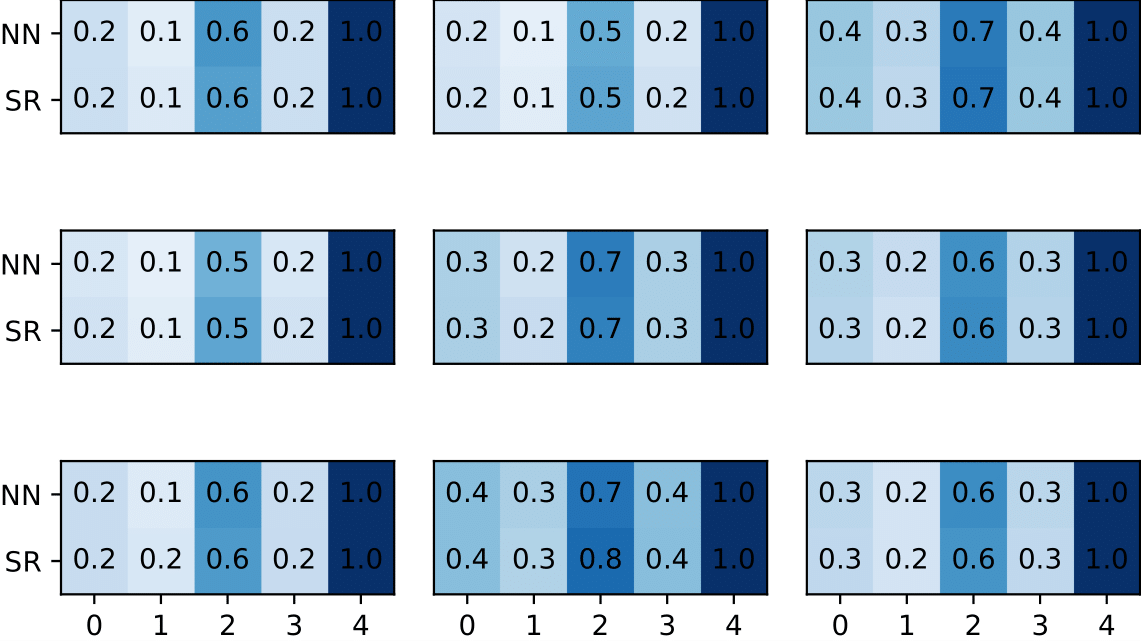}
	\end{subfigure}
	\begin{subfigure}{0.16\linewidth}
	    \caption{F5-h2}
		\includegraphics[width=\linewidth]{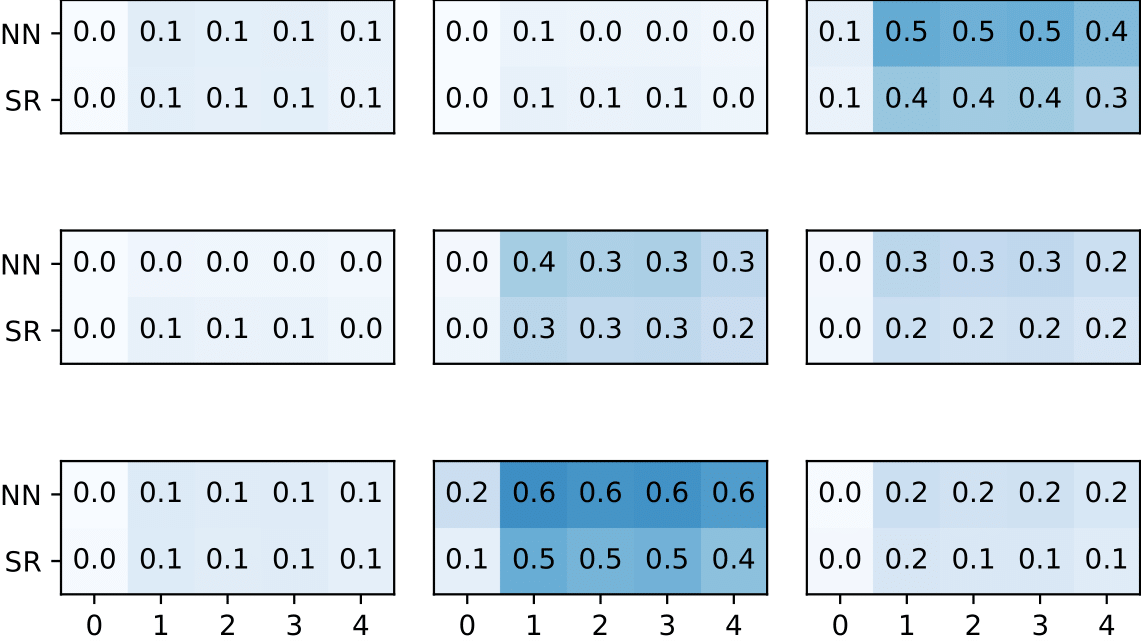}
	\end{subfigure}

	\caption{Each group of 9 heat maps represents the comparison of outputs of the SRNet layer vs the NN layer with 9 random input values 12 SR benchmarks.}
	\label{append_fig:hidden_heat_map}
\end{figure*}